\documentclass[sigconf]{acmart}

\AtBeginDocument{%
  \providecommand\BibTeX{{%
    \normalfont B\kern-0.5em{\scshape i\kern-0.25em b}\kern-0.8em\TeX}}}

\copyrightyear{2021}
\acmYear{2021}
\setcopyright{acmcopyright}
\acmConference[SenSys'21]{The 19th ACM Conference on Embedded Networked Sensor Systems}{November 15--17, 2021}{Coimbra, Portugal}
\acmBooktitle{The 19th ACM Conference on Embedded Networked Sensor Systems (SenSys'21), November 15--17, 2021, Coimbra, Portugal}
\acmPrice{15.00}
\acmDOI{10.1145/3485730.3485932}
\acmISBN{978-1-4503-9097-2/21/11}

\usepackage{subfig}
\usepackage{lipsum}

\ifodd 2
\newcommand{\rev}[1]{{\color{blue}#1}} %
\else
\newcommand{\rev}[1]{#1}
\fi

\begin{document}

\newcommand{\name}{MoRe-Fi}

\title{MoRe-Fi: Motion-robust and Fine-grained Respiration Monitoring via Deep-Learning UWB Radar}

\author{Tianyue Zheng$^{1,2}$,\quad Zhe Chen$^{1, 3}$, \quad Shujie Zhang$^{1}$,\quad Chao Cai$^{1}$,\quad Jun Luo$^1$}

\affiliation{
   \institution{$^1$ School of Computer Science and Engineering, Nanyang Technological University, Singapore \\
   $^2$ Energy Research Institute, Interdisciplinary Graduate Programme, Nanyang Technological University, Singapore \\
   $^3$ China-Singapore International Joint Research Institute, Guangzhou, China \\
}
    \country{Email: \{tianyue002, shujie002, junluo\}@ntu.edu.sg, chenz@ssijri.com, chriscai@hust.edu.cn}
}
\renewcommand{\authors}{T. Zheng, Z. Chen, S. Zhang, C. Cai, and J. Luo}
\renewcommand{\shortauthors}{T. Zheng, Z. Chen, S. Zhang, C. Cai, and J. Luo}

\begin{abstract}
Crucial for healthcare and biomedical applications, \textit{respiration monitoring} often employs wearable sensors in practice, causing inconvenience due to their direct contact with human bodies. Therefore, researchers have been constantly searching for \textit{contact-free} alternatives. Nonetheless, existing contact-free designs mostly require human subjects to remain static, largely confining their adoptions in everyday environments where body movements are inevitable. Fortunately, \textit{radio-frequency} (RF) enabled contact-free sensing, though suffering motion interference inseparable by conventional filtering, may offer a potential to distill respiratory waveform with the help of deep learning. To realize this potential, we introduce \name\ to conduct fine-grained respiration monitoring under body movements. \name\ leverages an IR-UWB radar to achieve contact-free sensing, and it fully exploits the complex radar signal for data augmentation. The core of \name\ is a novel \textit{variational encoder-decoder} network; it aims to single out the respiratory waveforms that are modulated by body movements in a non-linear manner. Our experiments with 12 subjects and 66-hour data demonstrate that \name\ accurately recovers respiratory waveform despite the interference caused by body movements. We also discuss potential applications of \name\ for pulmonary disease diagnoses. 
\end{abstract}

	


\maketitle

\section{Introduction}
Respiratory diseases~\cite{corren2003upper, williams1993chronic, soriano2020prevalence} are so common that the deadly health conditions caused by them influence people worldwide: they affect 2.4\% of the global population~\cite{vos2016global} and cause 7.6 million deaths per year globally~\cite{who}. Fortunately, most of them can be detected in their early stages with proper monitoring, as symptoms of these diseases, such as airflow obstruction~\cite{fletcher1977natural} and shortness of breath~\cite{holland2012breathing}, are usually reflected on different vital signs including respiratory rate~\cite{pollock1993estimation, gift1992relaxation} and fine-grained patterns in the respiratory waveform~\cite{bhatt2014fev1, tsuyuki2020diagnostic}. 
Traditionally, to obtain such vital signs 
for disease diagnosis, wearable devices ranging from smartwatches to medical sensors have been
used~\cite{hao2017mindfulwatch, min2014simplified, guder2016paper, neulog, jia2017monitoring, fang2016bodyscan}. Unfortunately, the inconvenience caused by the contact (even intrusive) nature of these sensors has prevented them from being widely adopted under daily environments. To overcome the drawbacks of contact sensing for
achieving ubiquitous respiration monitoring, \textit{contact-free} sensing has attracted increasing attention from both academia and industry~\cite{kaltiokallio2014non, adib2015smart, wang2018c, xu2019breathlisener, nguyen2016continuous, zheng2020v2ifi, yang2017vital, wang2019contactless, song2020spirosonic}. Among these developments. \textit{radio-frequency} (RF) sensing leveraging various commercial-grade radars has demonstrated a promising future~\cite{adib2015smart,yang2017vital,zheng2020v2ifi,movifi}, thanks to its noise resistance at a reasonable cost.

\begin{figure}[b]
    \setlength\abovecaptionskip{8pt}
    \vspace{-2ex}
    \centering
	\includegraphics[width=.96\linewidth]{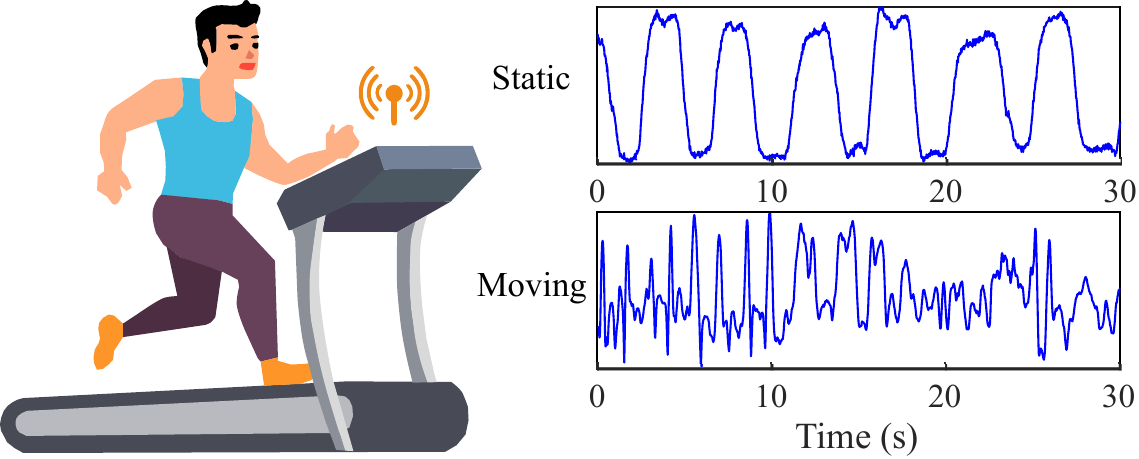}
	\caption{\rev{If the human subject is in motion, previous RF-sensing solutions for respiration monitoring fail.}}
	\label{fig:teaser}
\end{figure}

Ideally, monitoring should be performed continuously so that respiration patterns can be used as markers for intervention or evidence for diagnoses~\cite{brochard2012clinical, folke2003critical}. However, existing RF-sensing systems fail to deliver continuous monitoring as they often assume static human subjects.
Facing strong body movements, 
these systems have to suspend respiration monitoring~\cite{adib2015smart,yue2018extract}; otherwise, they would obtain noise-like readings as shown in Figure~\ref{fig:teaser}. 
In reality, we cannot expect subjects to remain static during monitoring for two major reasons. On one hand, human subjects may undergo a constant motion, e.g., typewriting or exercising. On the other hand, unintentional posture drifts~\cite{rasouli2017unintentional} and unconscious movements (e.g., turning-over during sleep) may always occur. Therefore, the static assumption contradicts the intention of continuous monitoring, rendering existing systems less applicable to real-life scenarios. Now the key question becomes: \textit{can we design a fine-grained respiration monitoring system working on subjects with body movements?}

Answering this question faces four major challenges. First of all, recovering respiratory waveform with RF-sensing is far from trivial even for static subjects, as existing systems often rely on linear filtering (potentially susceptible to nonlinear interference) to obtain coarse-grained waveform or only respiratory rate~\cite{adib2015smart,yue2018extract,zeng2019farsense}.
Secondly, the effect of body movements on the complex RF signal 
has %
never been put under scrutiny; prior art resorts to either phase or amplitude of RF signals~\cite{yue2018extract,zheng2020v2ifi,movifi}.
Thirdly, the reflected signals caused by body movements and respiration are composed in a nonlinear manner due to varying body positions, making it extremely hard to separate respiration from motion interference. 
Last but not least, reflected signals by body movements exhibit various statistical properties that cannot be readily processed by a single model-based method. So far, very few proposals have touched motion-robust
respiration monitoring: the RF-sensing methods either incur a high complexity~\cite{movifi} or handle very small-scale movements~\cite{zheng2020v2ifi}, while the acoustic sensing methods can be susceptible to real-life acoustic interference~\cite{xu2019breathlisener, song2020spirosonic}.

To tackle these challenges, we propose \name\ for motion-robust and fine-grained respiration monitoring. We construct \name\ based on a commercial-grade IR-UWB radar platform~\cite{octopus}, leveraging its large bandwidth to achieve high-resolution motion sensing.
Given the raw motion-induced IR-UWB signal embedded with fine-grained spatial information, we first analyze how respiration is modulated in the complex in-phase and quadrature (I/Q) components, as well as the limitations of model-based methods in extracting respiratory waveform. Based on the characteristics of I/Q-domain signal representation, we design a corresponding data augmentation process for enriching our dataset, which further drives an IQ Variational Encoder-Decoder (IQ-VED) for robust recovery of respiratory waveform. As the core of \name, IQ-VED captures the complementary I/Q information from the radar signal and encodes it to an interpretable latent representation for facilitating fine-grained respiratory waveform extraction. Our major contributions in designing and implementing \name\ are as follows:
\begin{itemize}
    \item %
    To the best of our knowledge, \name\ is the first fine-grained respiration monitoring system operating in a low-complexity and \textit{full-scale motion-robust} manner.
    \item We analyze the necessity to process radar signal in its complex I/Q domain, instead of leveraging incomplete information, such as only phase or amplitude of the signal.
    \item We propose IQ-VED for recovering and refining respiratory waveform. This novel encoder-decoder architecture fully utilizes the I/Q components together and achieves fine-grained respiratory waveform recovery leveraging the generalizability brought by the variational inference.
    \item We conduct extensive evaluations on \name\ with a 66-hour dataset; the results strongly confirm its excellent waveform recovery ability under body movements.
\end{itemize}
The rest of the paper is organized as follows. Section~\ref{sec:background} introduces the background for respiration monitoring with IR-UWB radar. Section~\ref{sec:design} presents the system design of \name, with detailed implementation discussed in Section~\ref{sec:implementation}. Section~\ref{sec:evaluation} reports the evaluation results. Potential applications of \name\ are discussed in Section~\ref{sec:discussion}. Related works are presented in Section~\ref{sec:related}. Finally, Section~\ref{sec:conclusion} concludes this paper and points out future directions.

\section{Background and preliminaries}\label{sec:background}
In this section, we carefully study the principles of respiration monitoring with an IR-UWB radar. We focus on modeling the radar signal to reflect how it represents respiration in the I/Q domain, hence laying a theoretical foundation for the robust respiration monitoring discussed in Section~\ref{sec:design}.

\subsection{Capturing Breath with IR-UWB Radar}\label{ssec:capturing}
\begin{figure}[h]
    \vspace{-1ex}
    \setlength\abovecaptionskip{8pt}
    \centering
	\includegraphics[width=.88\linewidth]{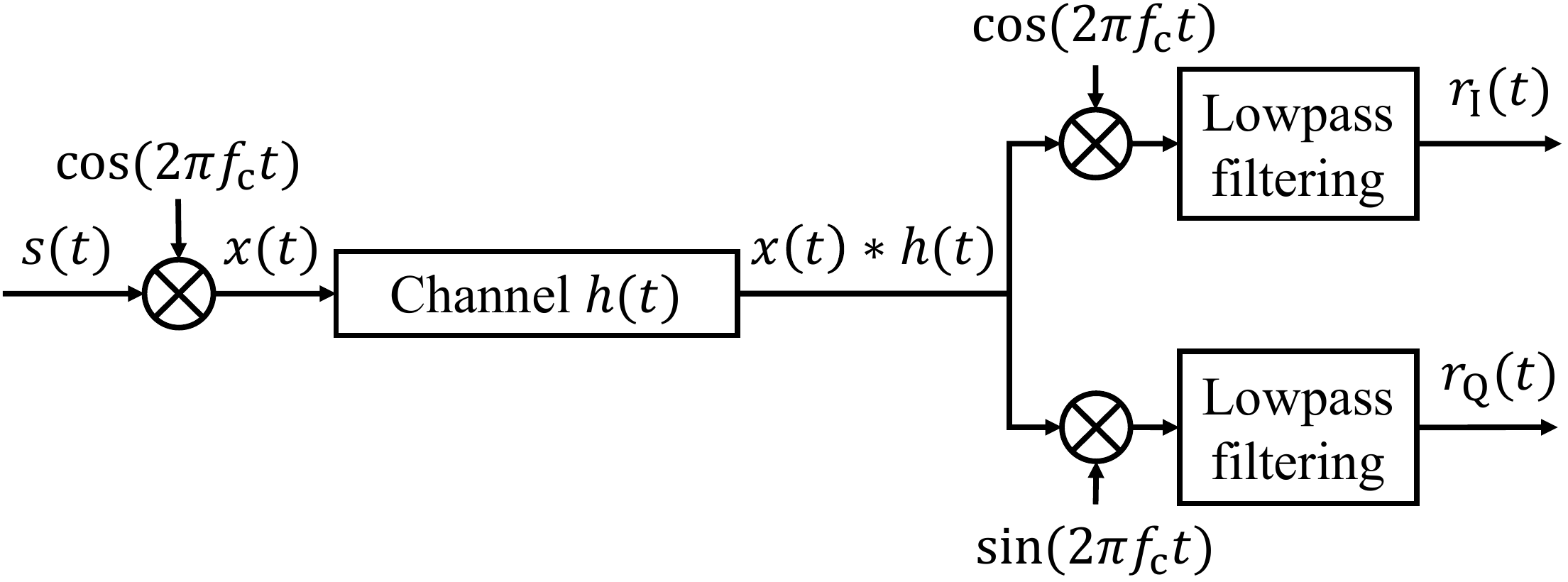}
	\caption{System diagram of the IR-UWB radar.}
	\label{fig:sysdiag}
	\vspace{-1ex}
\end{figure}

\begin{figure}[b]
    \setlength\abovecaptionskip{8pt}
    \vspace{-2ex}
	   \captionsetup[subfigure]{justification=centering}
		\centering
		\subfloat[Received signal $|r(t)|$.]{
		    \begin{minipage}[b]{0.47\linewidth}
		        \centering
			    \includegraphics[width = 0.96\textwidth]{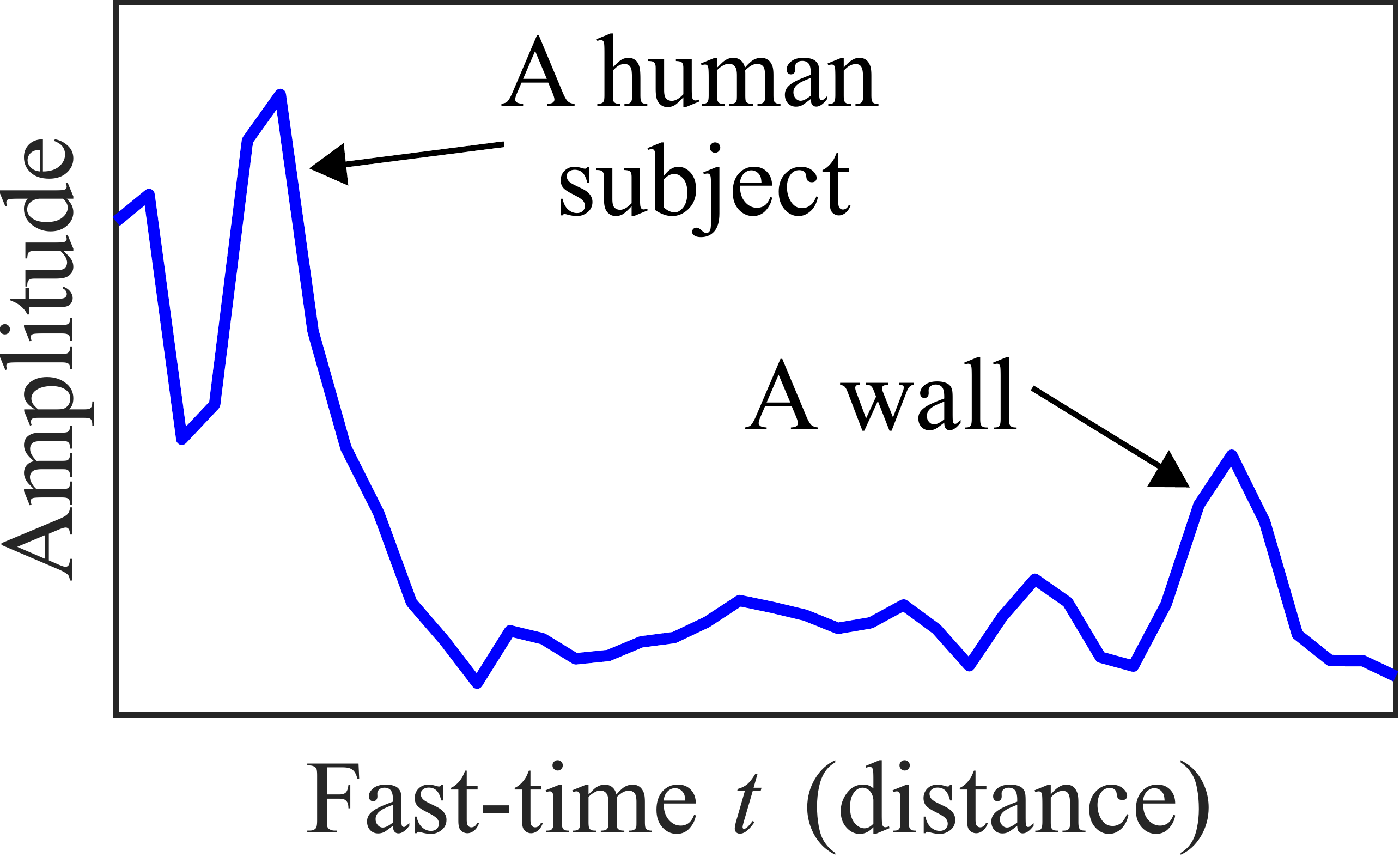}
			    \label{subfig:signal}
			\end{minipage}
		}
    		\subfloat[Signal matrix $\boldsymbol{r}(t, n)$.]{
		  \begin{minipage}[b]{0.47\linewidth}
		        \centering
			    \includegraphics[width = 0.96\textwidth]{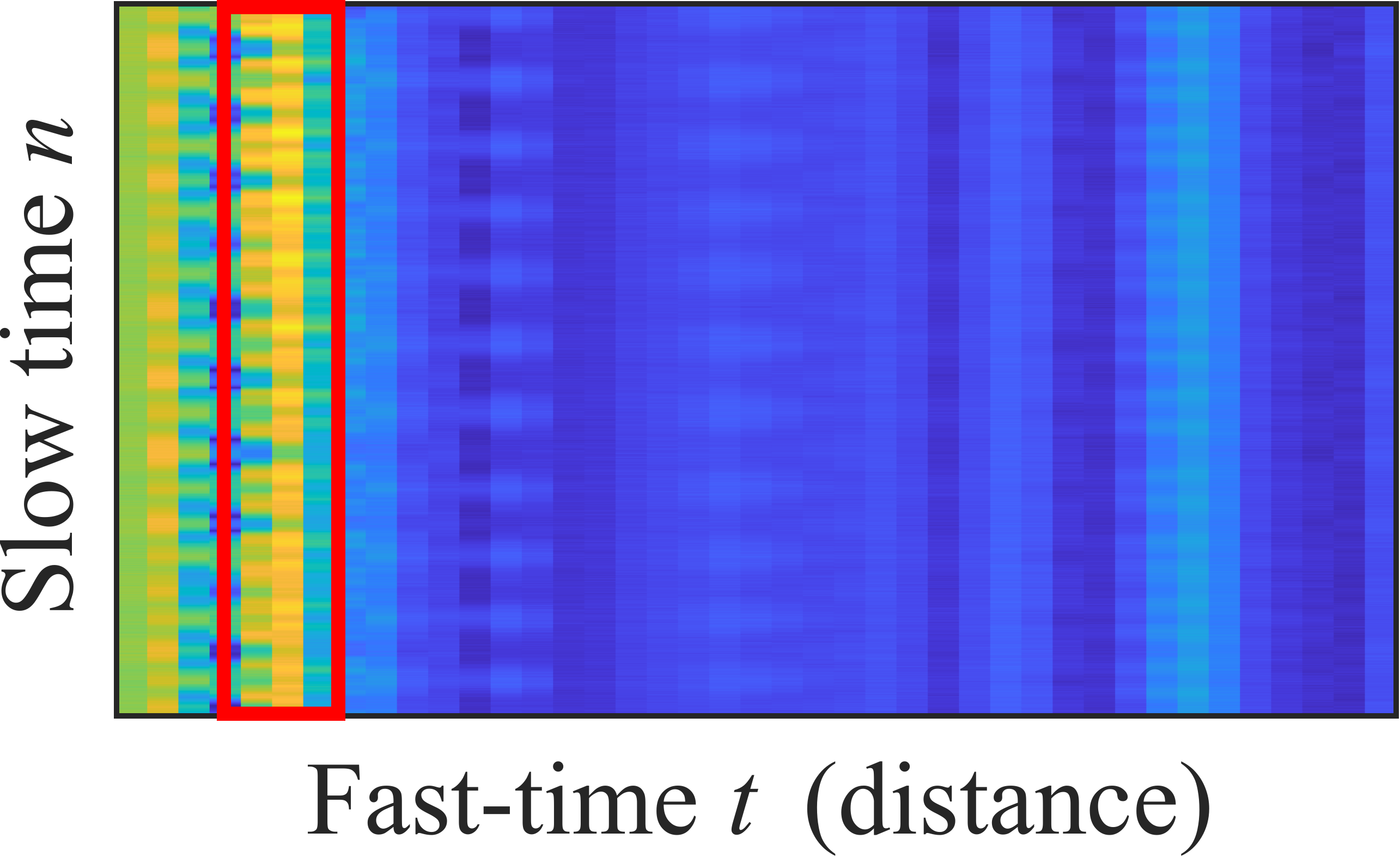}
			    \label{subfig:matrix}
			\end{minipage}
		}
		\caption{Amplitude of single-frame radar signal $r(t)$ and signal matrix $\boldsymbol{r}(t, n)$ composed of multiple frames.}
		\label{fig:signalmatrix}
\end{figure}

We first explain the working principles of IR-UWB radar on how it captures human respiration. As shown by the system diagram of IR-UWB radar in Figure~\ref{fig:sysdiag}, 
each frame $x(t)$ is formed by a baseband Gaussian pulse $s(t)$ modulated by a cosine carrier at frequency $f_\mathbf{c}$. This frame is then transmitted and reflected by a moving human chest and other irrelevant objects, so as to produce the received signal $x(t)*h(t)$, where $h(t)$ denotes the channel impulse response. After I/Q downconversion, the demodulated complex signal becomes $r(t) = r_{\mathrm{I}}(t) + j r_{\mathrm{Q}}(t)$. In Figure~\ref{subfig:signal}, we plot the amplitude of the complex signal; objects at different distances can be clearly differentiated thanks to the large bandwidth. 

\begin{figure}[t]
    \setlength\abovecaptionskip{6pt}
    \vspace{-1ex}
	   \captionsetup[subfigure]{justification=centering}
		\centering
		\subfloat[Signal segment 1.]{
		    \begin{minipage}[b]{0.47\linewidth}
		        \centering
			    \includegraphics[width = 0.96\textwidth]{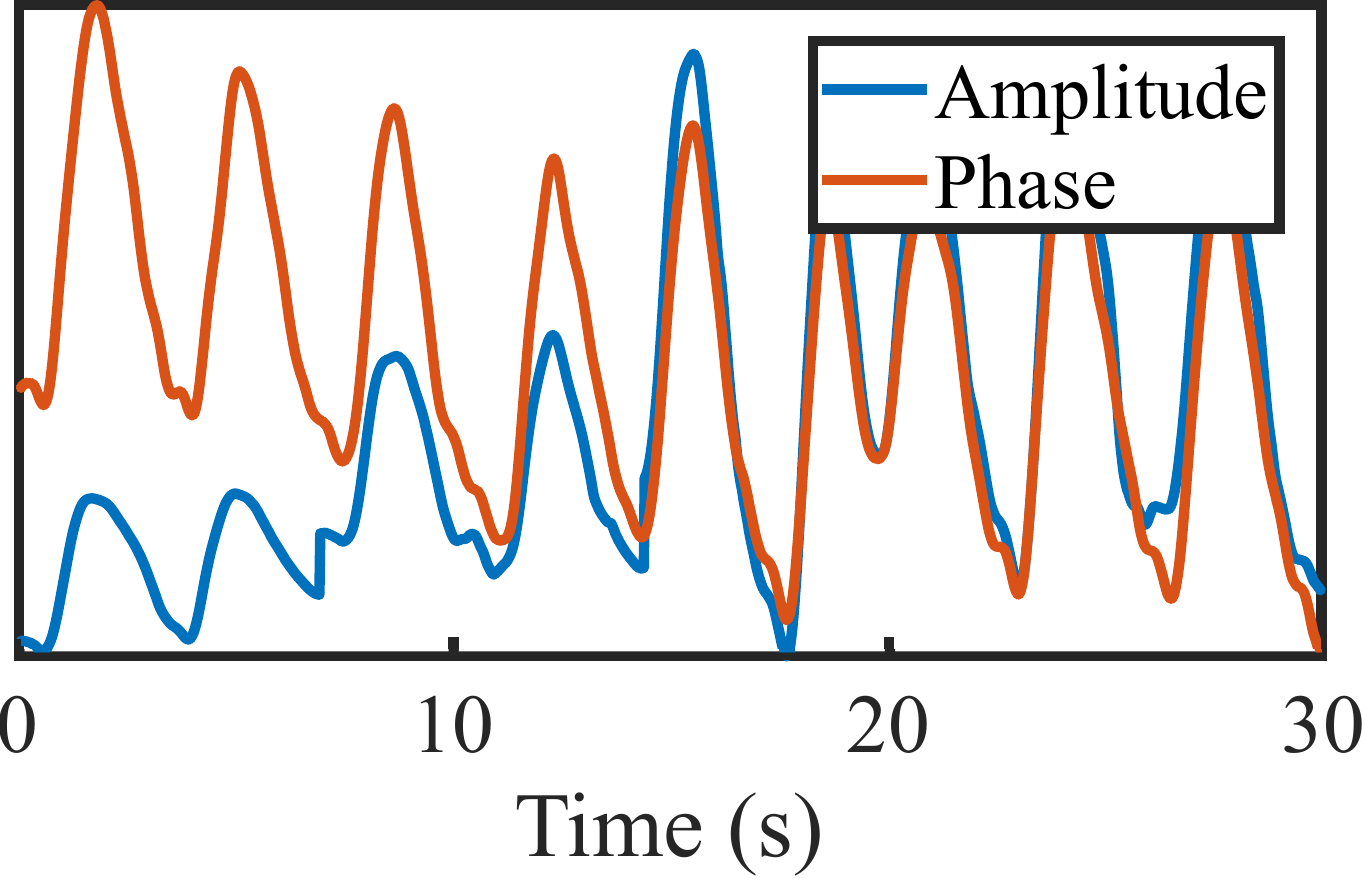}
			    \label{subfig:segment1}
			\end{minipage}
		}
		\subfloat[Signal segment 2.]{
		  \begin{minipage}[b]{0.47\linewidth}
		        \centering
			    \includegraphics[width = 0.96\textwidth]{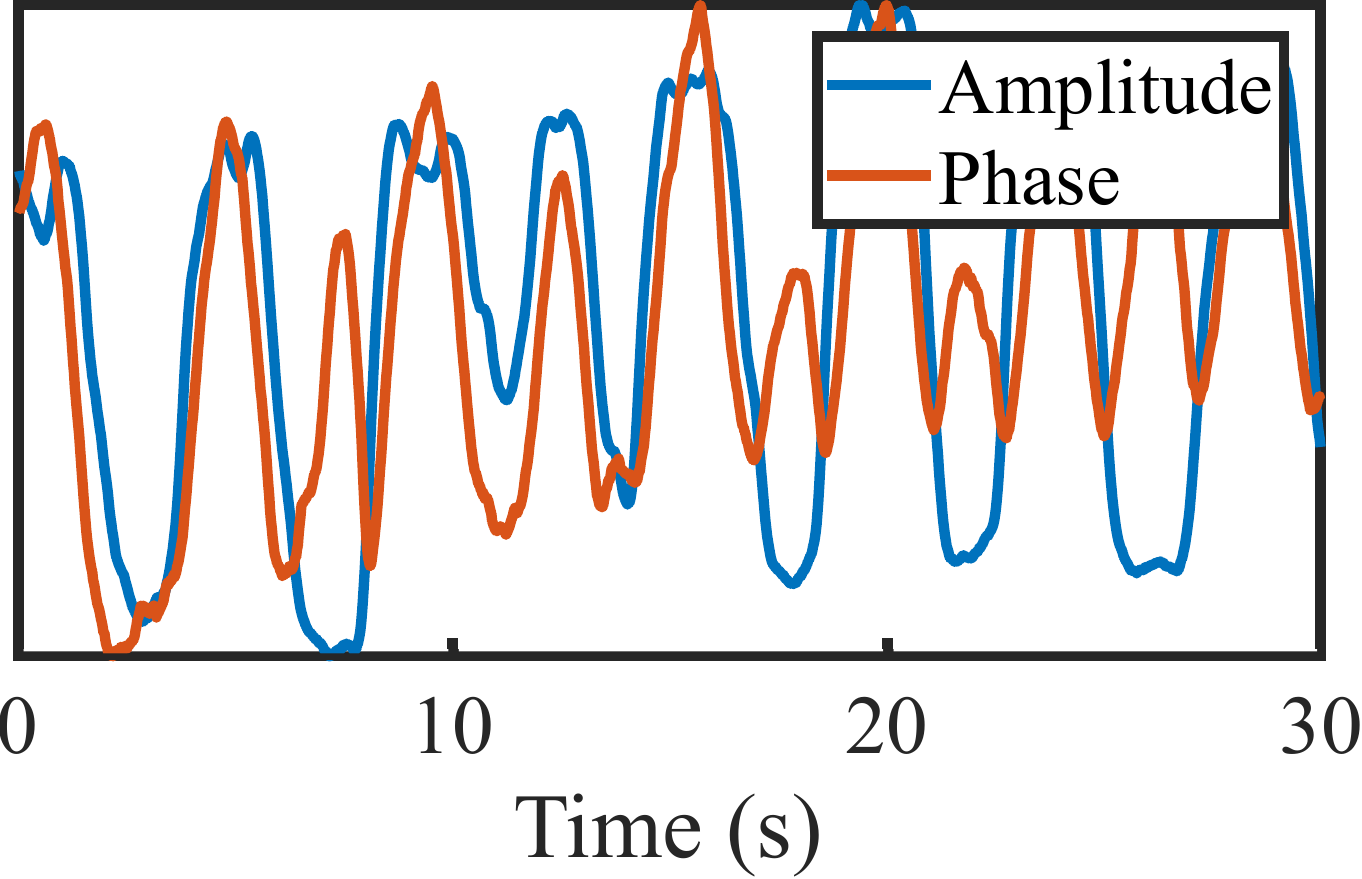}
			    \label{subfig:segment2}
			\end{minipage}
		}
		\caption{\rev{Neither amplitude nor phase of $r_t(n)$ alone is sufficient to correctly recover respiratory waveforms.}}
		\label{fig:IQ_bad}
	    \vspace{-2ex}
\end{figure}

However, a single frame of the received signal is not enough for respiration monitoring. To observe periodic movements, the radar transmits frames at a regular interval, and stacks the received frames to form a signal matrix $\boldsymbol{r}(t) = [r_1(t), \cdots r_n(t), \cdots r_N(t)]^T$, where $t$ and $n$ are respectively the \textit{fast-time} and \textit{slow-time} indices, and $N$ is the number of slow-time frames~\cite{RFMag,ding2020rfnet}. We hereafter slightly abuse the terminology by writing $\boldsymbol{r}(t)$ as $\boldsymbol{r}(t,n)$ to clearly indicate its matrix nature, as illustrated in Figure~\ref{subfig:matrix}. One may easily recognize a breathing person in the matrix, as circled in the red box. Considering a particular $t$ corresponding to the respiration, conventional methods individually adopt either amplitude or phase of the slow-time signal $r_t(n)$ to characterize respiration~\cite{adib2015smart, zeng2019farsense, yue2018extract,zheng2020v2ifi}; we plot two such results in Figure~\ref{fig:IQ_bad}. 

\begin{figure}[b]
    \setlength\abovecaptionskip{6pt}
    \vspace{-3ex}
	   \captionsetup[subfigure]{justification=centering}
		\centering
		\subfloat[Human subject is static.]{
		    \begin{minipage}[b]{0.47\linewidth}
		        \centering
			    \includegraphics[width = 0.96\textwidth]{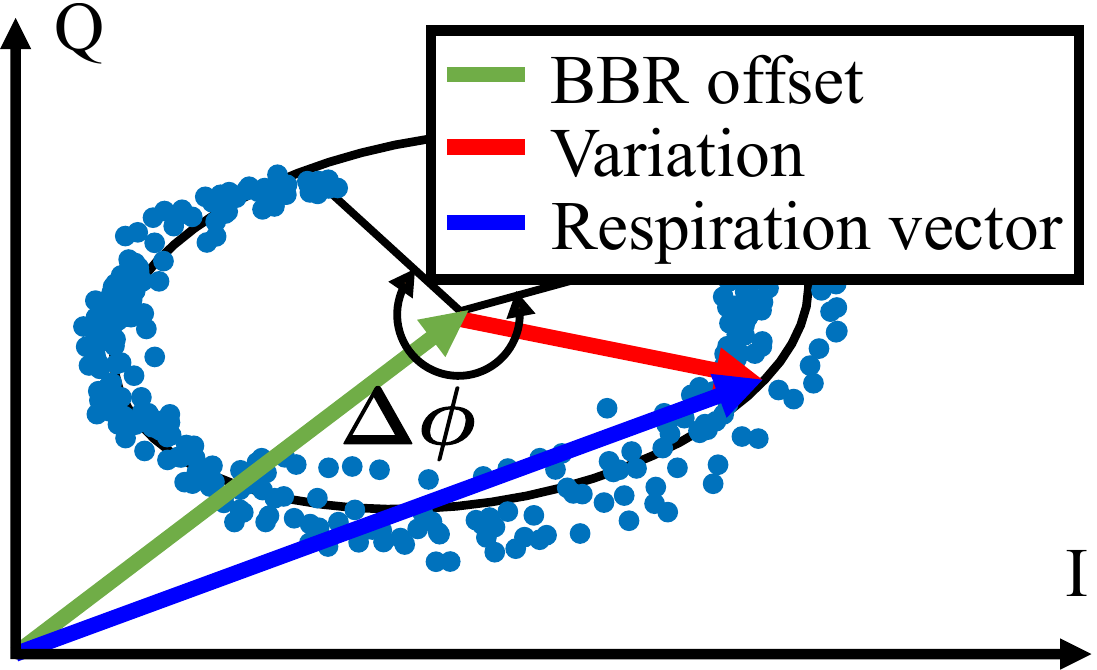}
			    \label{subfig:ellipse_static}
			\end{minipage}
		}
		\subfloat[Human subject is moving.]{
		  \begin{minipage}[b]{0.47\linewidth}
		        \centering
			    \includegraphics[width = 0.96\textwidth]{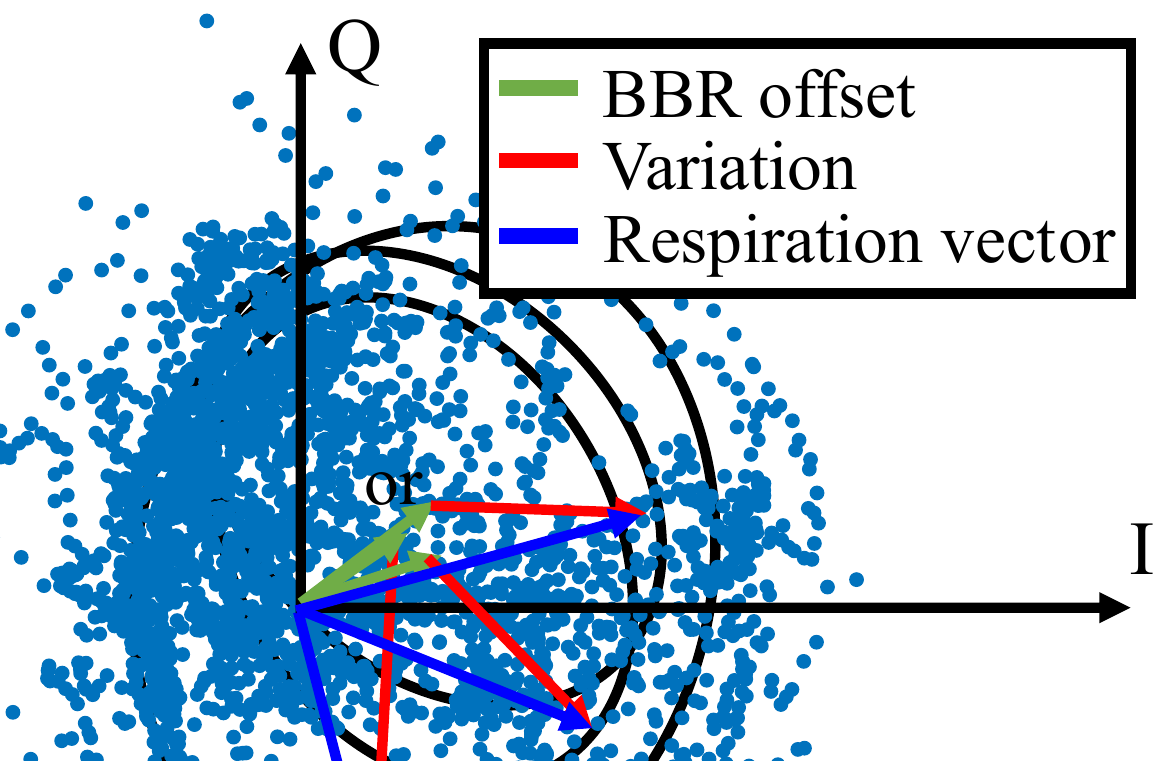}
			    \label{subfig:ellipse_dynamic}
			\end{minipage}
		}
		\caption{Constellation diagrams of $r(n)$.}
		\label{fig:ellipse}
\end{figure}

In fact, neither of these two real sequences can fully depict respiratory waveform accurately, despite their periodic structures resembling the ``baseband'' of respiration. As illustrated in Figure~\ref{fig:IQ_bad}, both amplitude and phase waveforms exhibit distortions to various extents, including strength variations (amplitude in Figure~\ref{subfig:segment1}) and missing cycles (phase in Figure~\ref{subfig:segment2}).
To understand why such distortions take place, we analyze $r(n)$ (subscript neglected for brevity) on both I/Q components
with Equations~\eqref{eq:IQ1} and~\eqref{eq:IQ2}: 
\begin{align}
r_{\mathrm{I}}(n)=\alpha(n) \cos \left(\frac{4 \pi d_{0}}{\lambda}+\frac{4 \pi z(n)}{\lambda}\right)+o_{\mathrm{I}}^{\mathrm{BBR}}, \label{eq:IQ1} \\
r_{\mathrm{Q}}(n)=\alpha(n) \sin \left(\frac{4 \pi d_{0}}{\lambda}+\frac{4 \pi z(n)}{\lambda}\right)+o_{\mathrm{Q}}^{\mathrm{BBR}}, \label{eq:IQ2}
\end{align}
where $\alpha(n)$ is the strength of the reflected signal from the human chest, $d_0$ is the distance from the radar to the chest, $\lambda$ is carrier wavelength, and $z(n)$ denotes human chest movement. In both equations, the first terms are caused by respiration, and the second terms $o_{\mathrm{I}}^{\mathrm{BBR}}$ and $o_{\mathrm{Q}}^{\mathrm{BBR}}$ are the offsets caused by \textit{body background reflection} (BBR). 
To visualize $r_{\mathrm{I}}(n)$ and $r_{\mathrm{Q}}(n)$, we take the signal of the breathing person (as bounded by the red box in Figure~\ref{subfig:matrix}) as an example, and display it as a constellation diagram in Figure~\ref{subfig:ellipse_static}. The blue \textit{respiration vector} %
in the graph corresponds to the complex signal $r_{\mathrm{I}}(n)+j r_{\mathrm{Q}}(n)$; they are the sum of BBR offset (green vector) and respiration-induced variation (red vector). One may observe that, as the human subject breathes, the red vector rotates and the trace of the blue vector forms an elliptic arc. The arc may not be circular because $\alpha(n)$, the radius, is time-varying due to a varying radar cross-section~\cite{lee2017effects}. 
Now Figure~\ref{subfig:ellipse_static} clearly explains why neither amplitude nor phase waveforms can correctly characterize respiratory waveforms in Figure~\ref{fig:IQ_bad}: although they oscillate with a similar frequency to a human breath, they are only projections of the respiration vector trace onto a lower dimension.

\subsection{Interference Caused by Body Movements} 
\label{ssec:interference}
\begin{figure}[h]
    \setlength\abovecaptionskip{8pt}
    \vspace{-1ex}
    \centering
	\includegraphics[width=.68\linewidth]{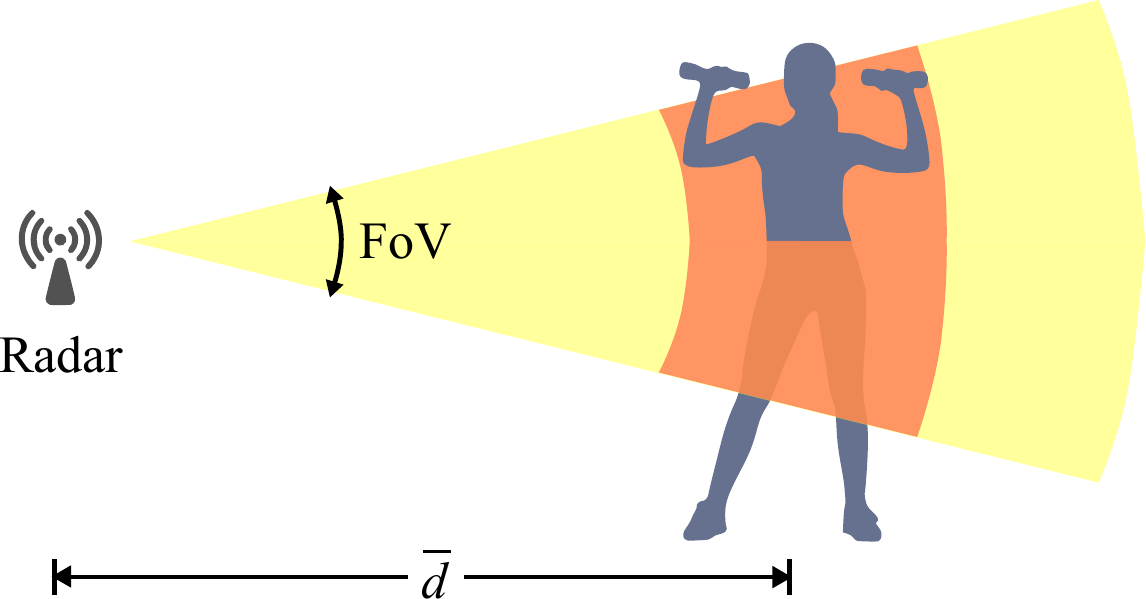}
	\caption{A subject of \name\ should stay within the radar FoV, with its distance from the radar remaining within a reasonable range around a constant mean value $\bar{d}$.}
	\label{fig:scope}
	\vspace{-1ex}
\end{figure}

To better understand the effects of body movements, we begin by defining their scope, as shown in Figure~\ref{fig:scope}. Basically, our concerned movements should maintain the position of the human subject i) to stay within the Field of View (FoV) of the radar and ii) to have the variable distance $d$ between the radar and the subject lying within a reasonable range (e.g., 50~\!cm) around its mean $\bar{d}$.
Such a scope encompasses body movements not affecting the chest (e.g., typewriting or limb position drift), as well as even larger-scale ones involving chest motion (e.g., exercising on treadmill or at a fixed spot). It, therefore, forbids the subject to i) drastically change its posture (e.g., from standing to lying) and ii) significantly alter its position (so as to avoid the need for tracking). This movement scope has barely been studied in the literature, as previous ``motion-robust'' systems mostly tackle small-scale movements, e.g., hand drifts when holding a device~\cite{song2020spirosonic} and turning steering wheels during driving~\cite{xu2019breathlisener, zheng2020v2ifi}. It should be noted that acoustic sensing~\cite{song2020spirosonic, xu2019breathlisener} are not applicable to our problem scope, because they are easily interfered with by acoustic noise accompanying large-scale movements (e.g., typewriting or exercising on treadmill) and real-life environments (e.g., music in vehicles).

With the scope defined, we hereby analyze the impacts of body movements on the I/Q signal with a simple experiment. %
We let a human subject move his body (lean forward and backward, sway left and right) while sitting in a chair. The corresponding $r(n)$ shown in Figure~\ref{subfig:ellipse_dynamic} demonstrates that body movements prevent the trace of $r(n)$ from falling onto a single elliptic arc; the trace is scattered across the I/Q plane in a rather arbitrary manner.
The reason is that the BBR offset (caused by reflections from limbs and torso other than the chest) previously presumed to be static is no longer so, resulting in a random shift of the elliptic center $\left(o_{\mathrm{I}}^{\mathrm{BBR}}, o_{\mathrm{Q}}^{\mathrm{BBR}}\right)$. In addition, the distance from the radar to the human subject $d$ also varies and further changes the signal phase.\footnote{Being a constant $d_0$ in~\eqref{eq:IQ1} and~\eqref{eq:IQ2}, the distance becomes a variable now.} Furthermore, the radar cross-section changes with body movements as well, causing a varying reflected signal strength and hence unpredictable changes in the long/short axes of the ellipse. Last but not least, Figure~\ref{subfig:ellipse_dynamic} only shows $r(n)$ for a single fast-time index, yet large-scale movements can potentially affect multiple fast-time indices.

\subsection{Conventional Respiration Recovery}
\label{ssec:conventional}
In order to recover respiratory waveform, conventional approaches measure the human chest displacement $\Delta d$; it is related to the phase change $\Delta \phi$ of the respiration vector in Figure~\ref{subfig:ellipse_static} by 
$
    \Delta d = \lambda \frac{\Delta \phi}{2 \pi}
$.
Because the center of the ellipse may not coincide with the origin due to the BBR offset, $\Delta \phi$ is not always obtainable from the raw I/Q signal. To recover the respiratory waveform, a recent proposal~\cite{song2020spirosonic} suggests fitting elliptic arcs to (acoustic) I/Q signal and unifies their centers to the origin. Although this method theoretically allows the phase of the respiration vector to be calculated by taking the inverse tangent of the shifted I/Q signal, it is intrinsically designed for quasi-static human subjects, although it can tolerate very small limb position drifts (e.g., hand motions when holding a phone~\cite{song2020spirosonic}) by selectively fitting to relatively ``clean'' data segments free of strong motion interference.

\begin{figure}[h]
    \vspace{-.5ex}
    \setlength\abovecaptionskip{8pt}
    \centering
	\includegraphics[width=.95\linewidth]{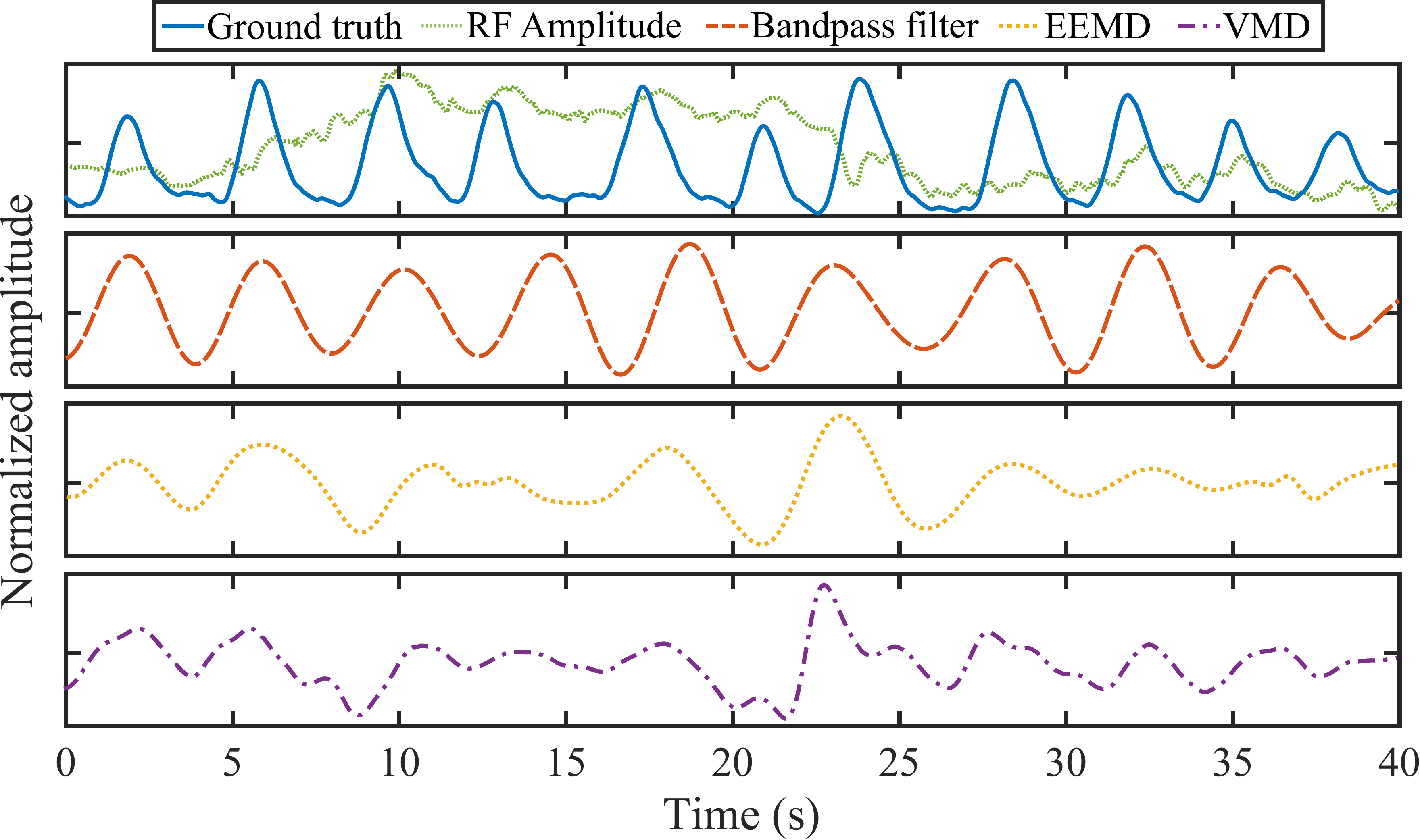}
	\caption{Respiratory waveforms obtained by 1-D signal processing methods under strong body movements; the event details on the waveforms are all lost (compared with the ground truth), and the rates can be wrongly represented.}
	\label{fig:conventional}
	\vspace{-1.5ex}
\end{figure}

Under stronger movements, the traces of respiration vector in the form of elliptic arcs are no longer identifiable and analyzable, because they are severely deformed and blended, as explained in Section~\ref{ssec:interference}. %
As a result, one has to fall back to the 1-D signal projection shown by Figure~\ref{fig:IQ_bad}, and hence apply linear processing methods taken by previous works to extract respiration, such as bandpass filter~\cite{ravichandran2015wibreathe}, Ensemble Empirical Mode Decomposition (EEMD)~\cite{xu2019breathlisener}, and Variational Mode Decomposition (VMD)~\cite{zheng2020v2ifi}. We briefly illustrate the recovery results achieved by these 1-D methods in Figure~\ref{fig:conventional}, when the subject was exercising on the spot. One may observe that the resulting waveforms are very coarse-grained and lack correct event details (e.g., rate, duration of inhalation and exhalation, as well as tidal volume). Essentially, as these methods fail to properly treat the I/Q scrambling and fast-time crossing issues of large-scale movements, they definitely cannot handle the body movement scope defined in Section~\ref{ssec:interference}.

\paragraph{Remark} To summarize, in order to recover fine-grained respiratory waveform under strong body movements, we have to fully utilize the I/Q signal and trace its variations across multiple fast-time indices. Nonetheless, since conventional model-based signal processing algorithms have been demonstrated as unable to cope with this situation, we resort to a data-driven approach.

\section{System Design}\label{sec:design}
\begin{figure}[t]
    \setlength\abovecaptionskip{8pt}
    \centering
	\includegraphics[width=.98\linewidth]{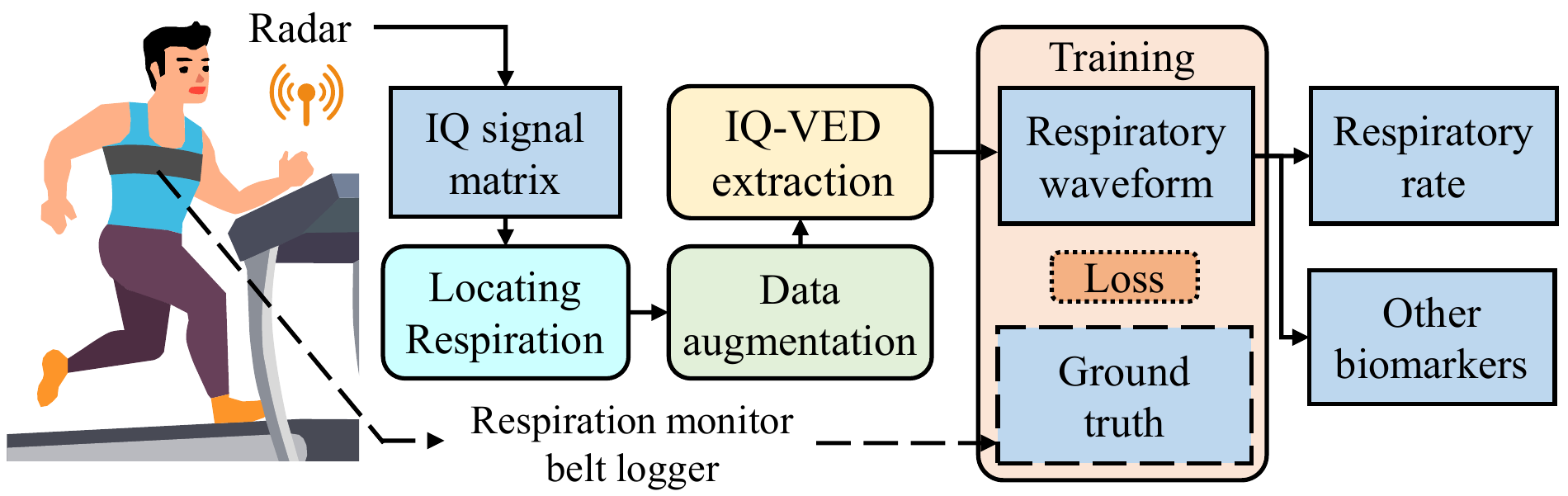}
	\caption{System diagram of \name.}
	\label{fig:diagram}
	\vspace{-2ex}
\end{figure}

This section introduces the design of \name, whose rough diagram is shown in Figure~\ref{fig:diagram}. Upon obtaining the IQ signal matrix $\boldsymbol{r}(t,n)$ from the IR-UWB radar, \name\ locates respiration by extracting a sub-matrix 
corresponding to the concerned human subject. It then leverages the rotation invariance in the I/Q domain to augment the sub-matrix. The core of \name\ is a novel IQ Variational Encoder-Decoder (IQ-VED) neural network to distill respiratory waveform from the sub-matrix. Trained by the ground truth waveform obtained from a wearable sensor~\cite{neulog}, our IQ-VED is capable of recovering fine-grained respiratory waveform under severe interference produced by body movements.

\subsection{Locating Respiration} 
\label{ssec:loc}
\begin{figure}[b]
    \setlength\abovecaptionskip{8pt}
    \vspace{-2ex}
    \centering
	\includegraphics[width=.9\linewidth]{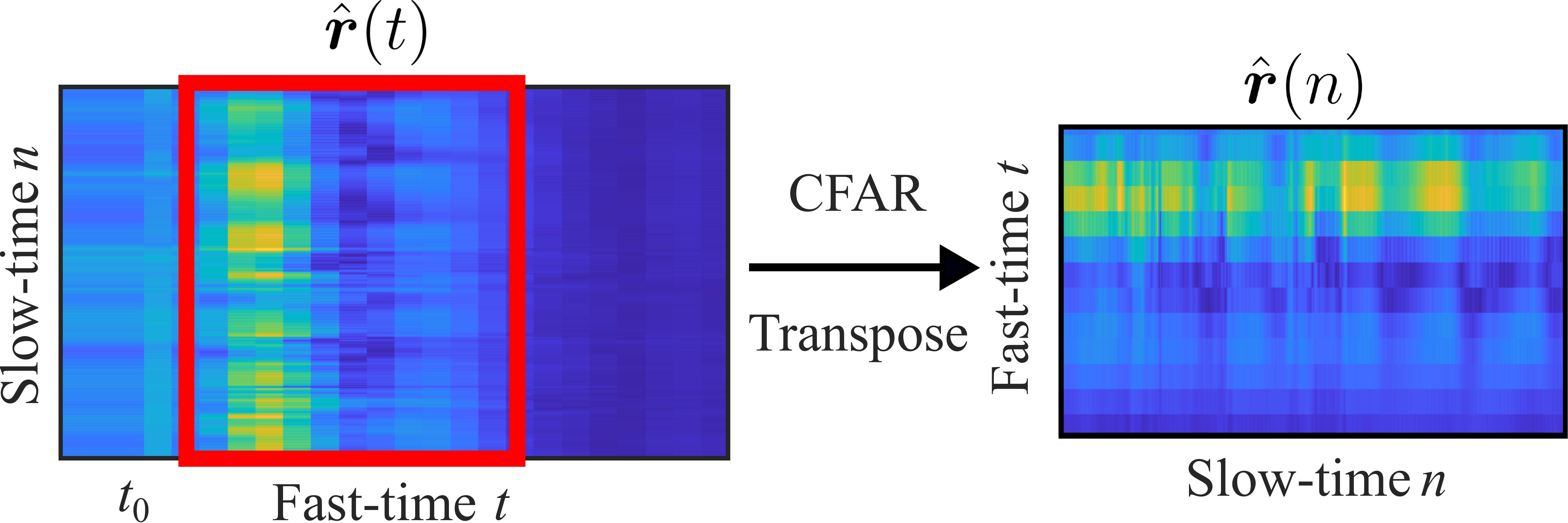}
	\caption{After locating respiration in a sub-matrix $\hat{\boldsymbol{r}}(t)$, \name\ transposes it to obtain $\hat{\boldsymbol{r}}(n)$, which contains multiple fast-time indices around the detected respiration.}
	\label{fig:localization}
\end{figure}

To locate respiration in the signal matrix $\boldsymbol{r}(t,n)$, \name\ first uses a loopback filter~\cite{3Dtracking-NSDI} to remove the influence of static background. The static clutter of the system can be described as: $c_{n}(t)=\beta c_{n-1}(t)+(1-\beta) r_{n}(t)$ and the background subtracted signal can be represented as $r^{-}_{n}(t)=r_{n}(t)-c_{n}(t)$, where $r_n(t)$ denotes the $n$-th frame and the weight $\beta$ is empirically set to 0.9. Then the Constant False Alarm Rate (CFAR) algorithm~\cite{levanon1988radar} kicks in to detect the peaks in $\boldsymbol{r}^{-}(t,n) = [r^{-}_1(t), \cdots r^{-}_n(t), \cdots r^{-}_N(t)]^T$ using an adaptive threshold; the threshold $\tau_{\mathrm{noise}}(t)$ is estimated
by averaging values at neighboring fast-time indices. \name\ selects multiple fast-time indices adjacent to the detected peaks to form a sub-matrix $\hat{\boldsymbol{r}}(t)$, and it finally transposes $\hat{\boldsymbol{r}}(t)$ to obtain a new matrix $\hat{\boldsymbol{r}}(n)$ with slow-time index $n$ as the argument, as shown in Figure~\ref{fig:localization}.

\subsection{Data Augmentation} \label{ssec:daug}
\begin{figure}[t]
    \setlength\abovecaptionskip{8pt}
    \vspace{-1ex}
	   \captionsetup[subfigure]{justification=centering}
		\centering
		\subfloat[Original signal.]{
		    \begin{minipage}[b]{0.32\linewidth}
		        \centering
			    \includegraphics[width = 0.96\textwidth]{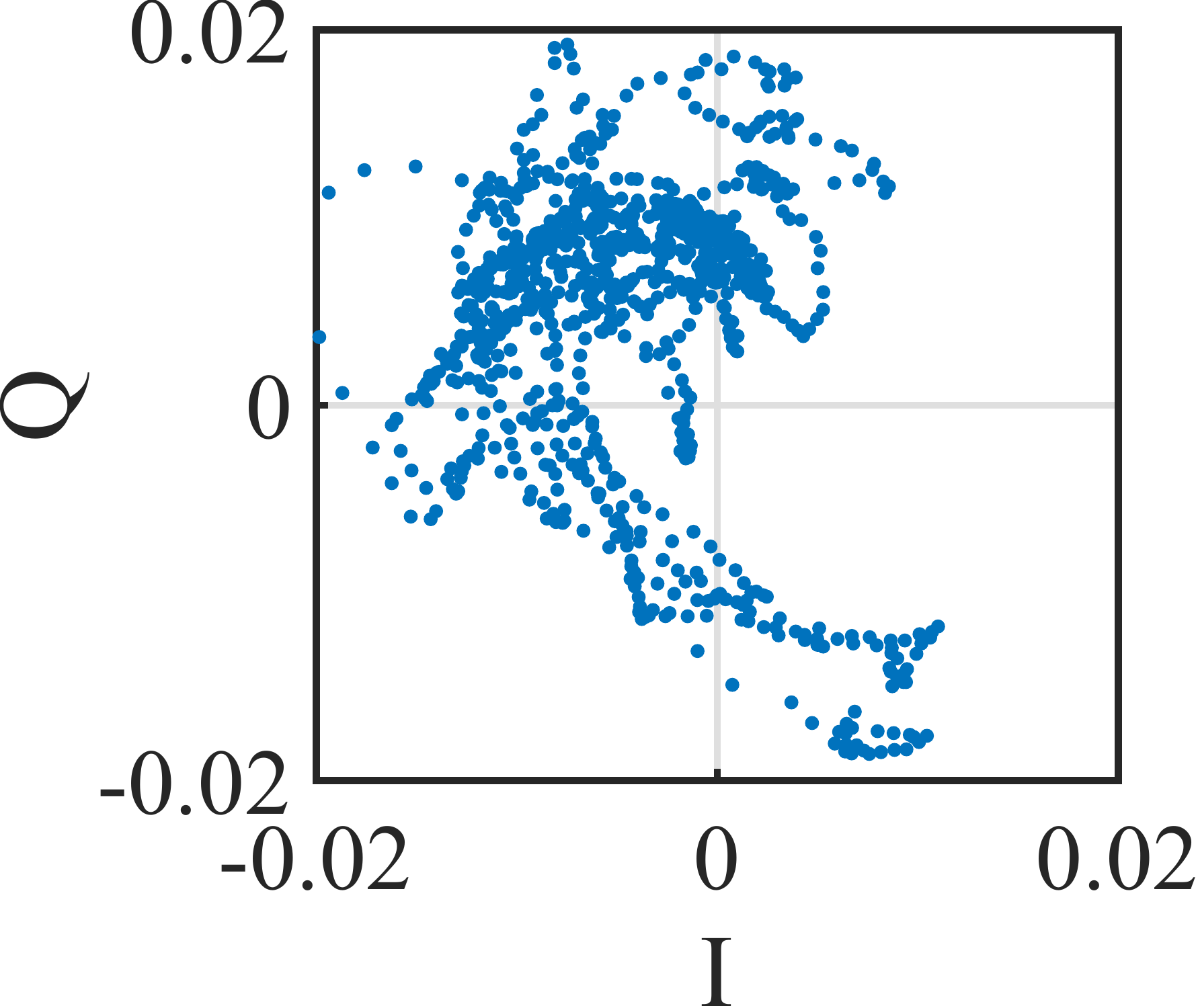}
			    \label{subfig:aug0}
			\end{minipage}
		}
		\subfloat[$\theta = \frac{\pi}{3}$.]{
		  \begin{minipage}[b]{0.32\linewidth}
		        \centering
			    \includegraphics[width = 0.96\textwidth]{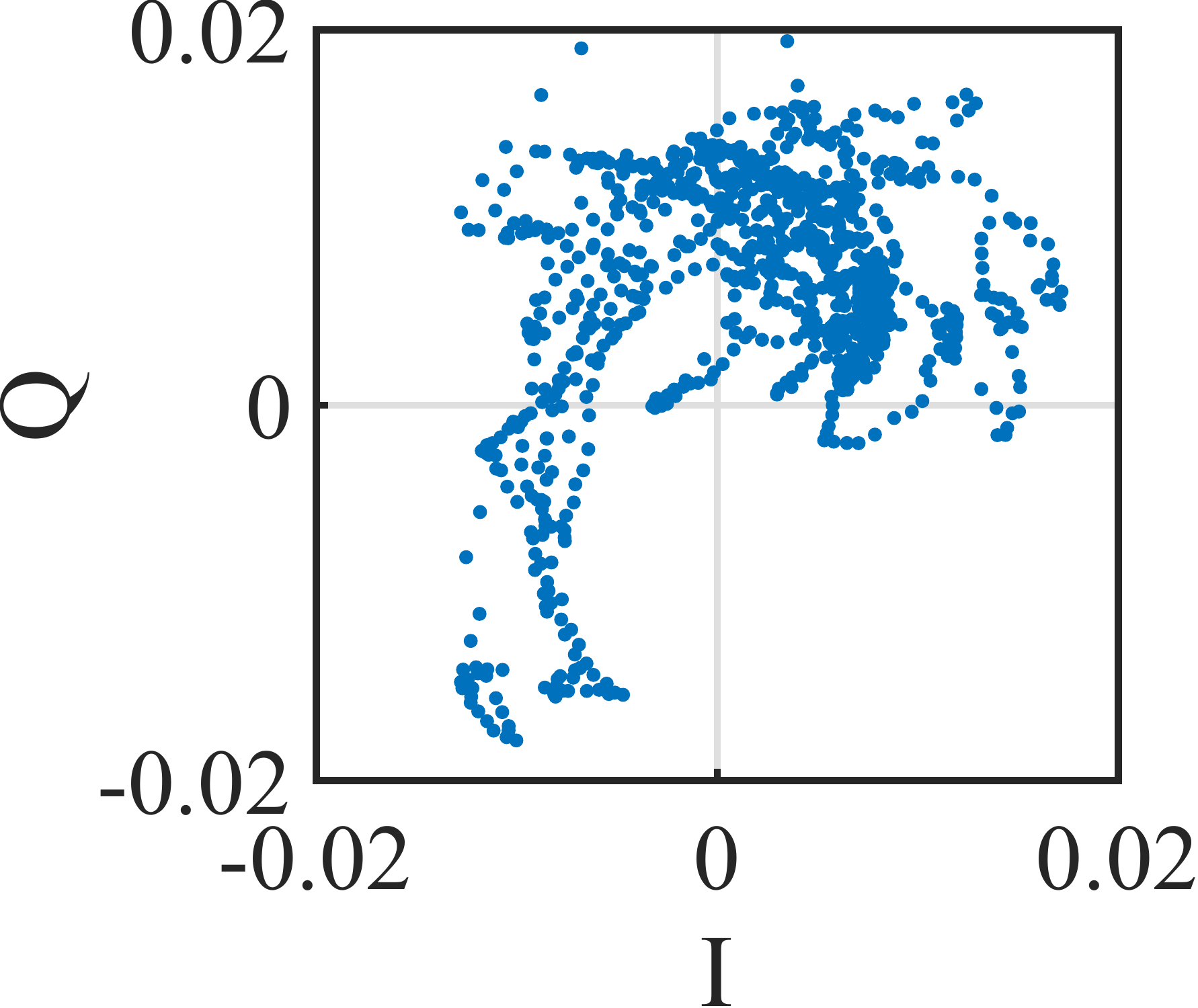}
			    \label{subfig:aug60}
			\end{minipage}
		}
		\subfloat[$\theta = \frac{2\pi}{3}$.]{
		  \begin{minipage}[b]{0.32\linewidth}
		        \centering
			    \includegraphics[width = 0.96\textwidth]{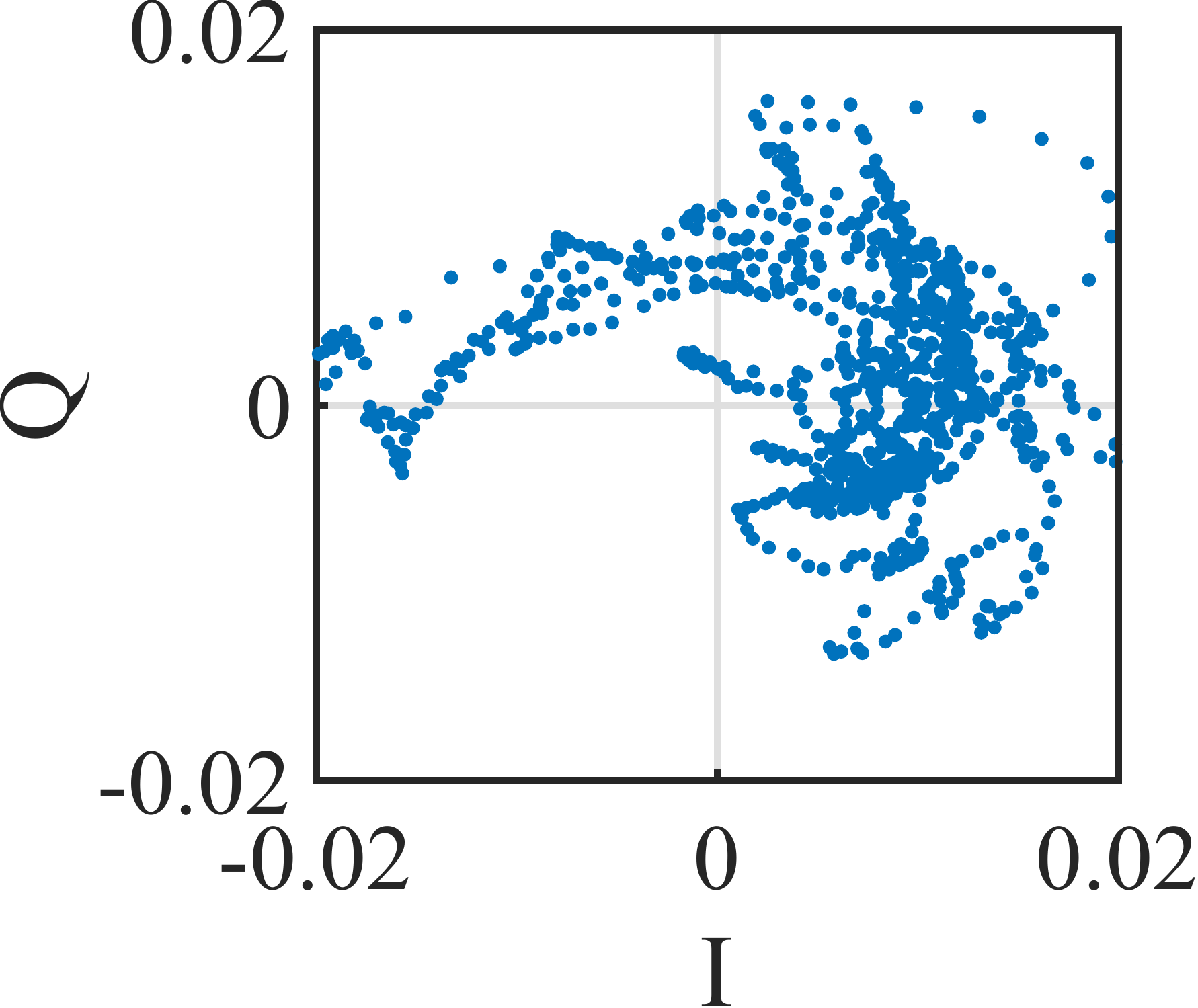}
			    \label{subfig:aug120}
			\end{minipage}
		}
		\\
		\subfloat[$\theta = \pi$.]{
		    \begin{minipage}[b]{0.32\linewidth}
		        \centering
			    \includegraphics[width = 0.96\textwidth]{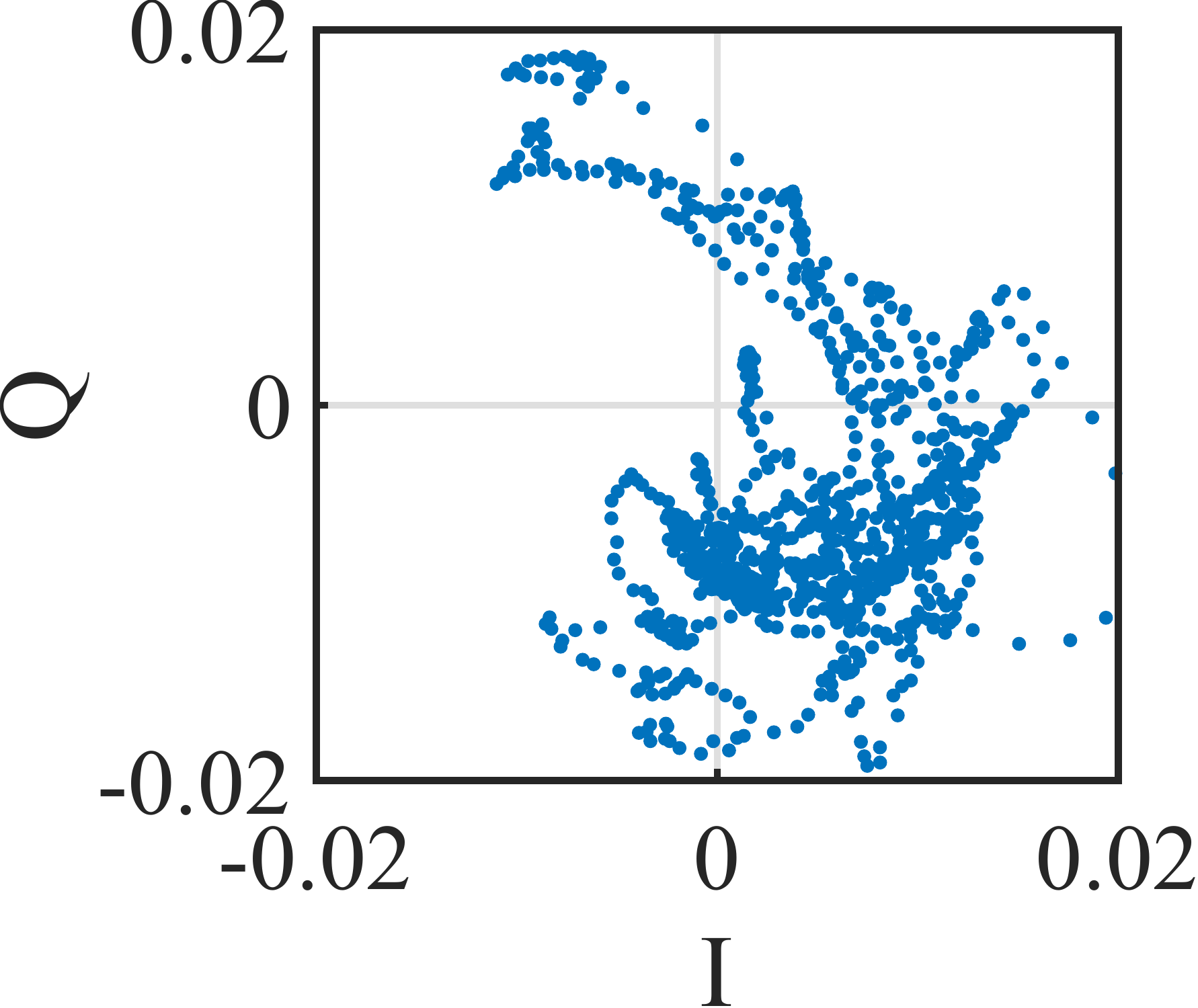}
			    \label{subfig:aug180}
			\end{minipage}
		}
		\subfloat[$\theta = \frac{4 \pi}{3}$.]{
		  \begin{minipage}[b]{0.32\linewidth}
		        \centering
			    \includegraphics[width = 0.96\textwidth]{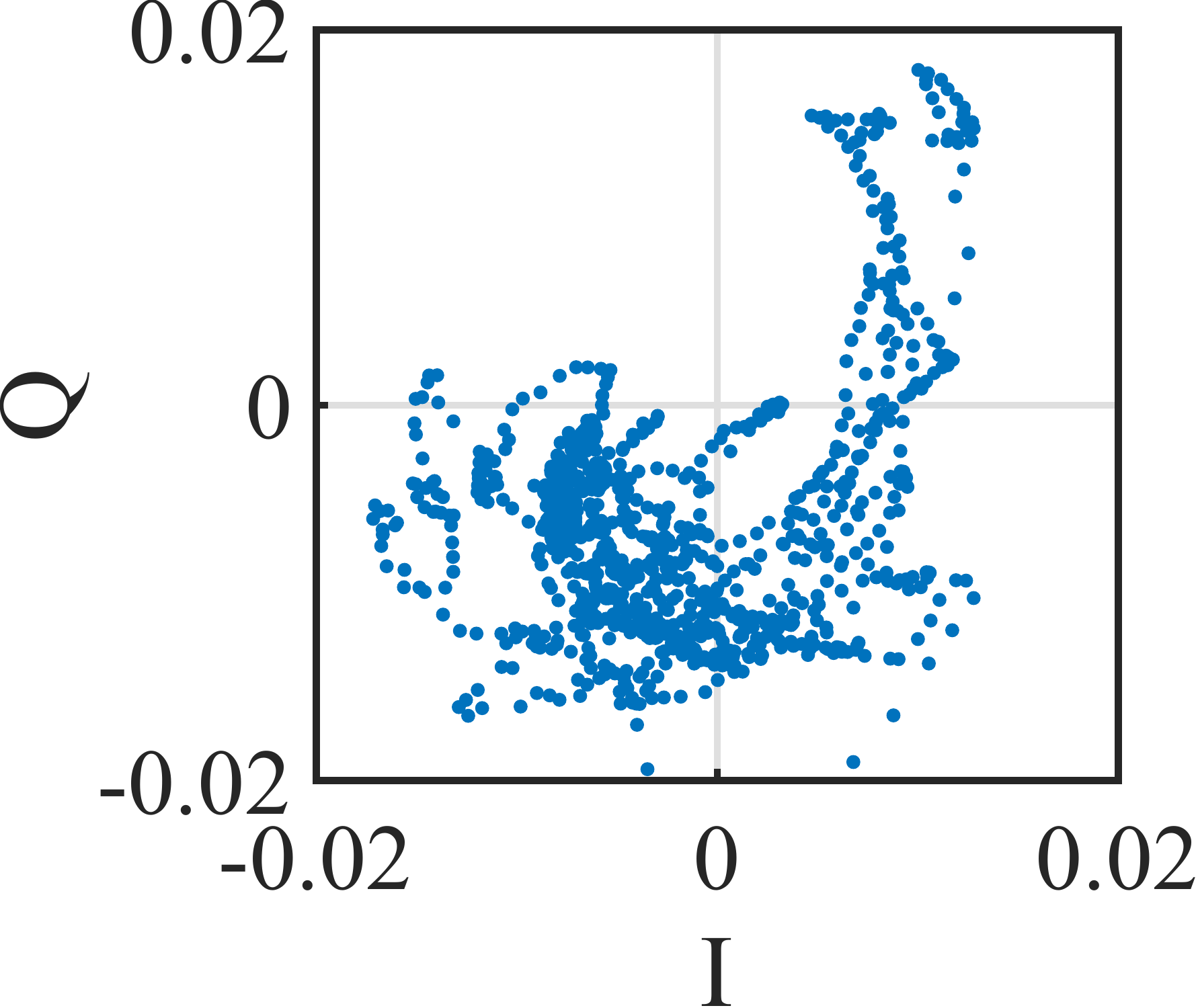}
			    \label{subfig:aug240}
			\end{minipage}
		}
		\subfloat[$\theta = \frac{5 \pi}{3}$.]{
		  \begin{minipage}[b]{0.32\linewidth}
		        \centering
			    \includegraphics[width = 0.96\textwidth]{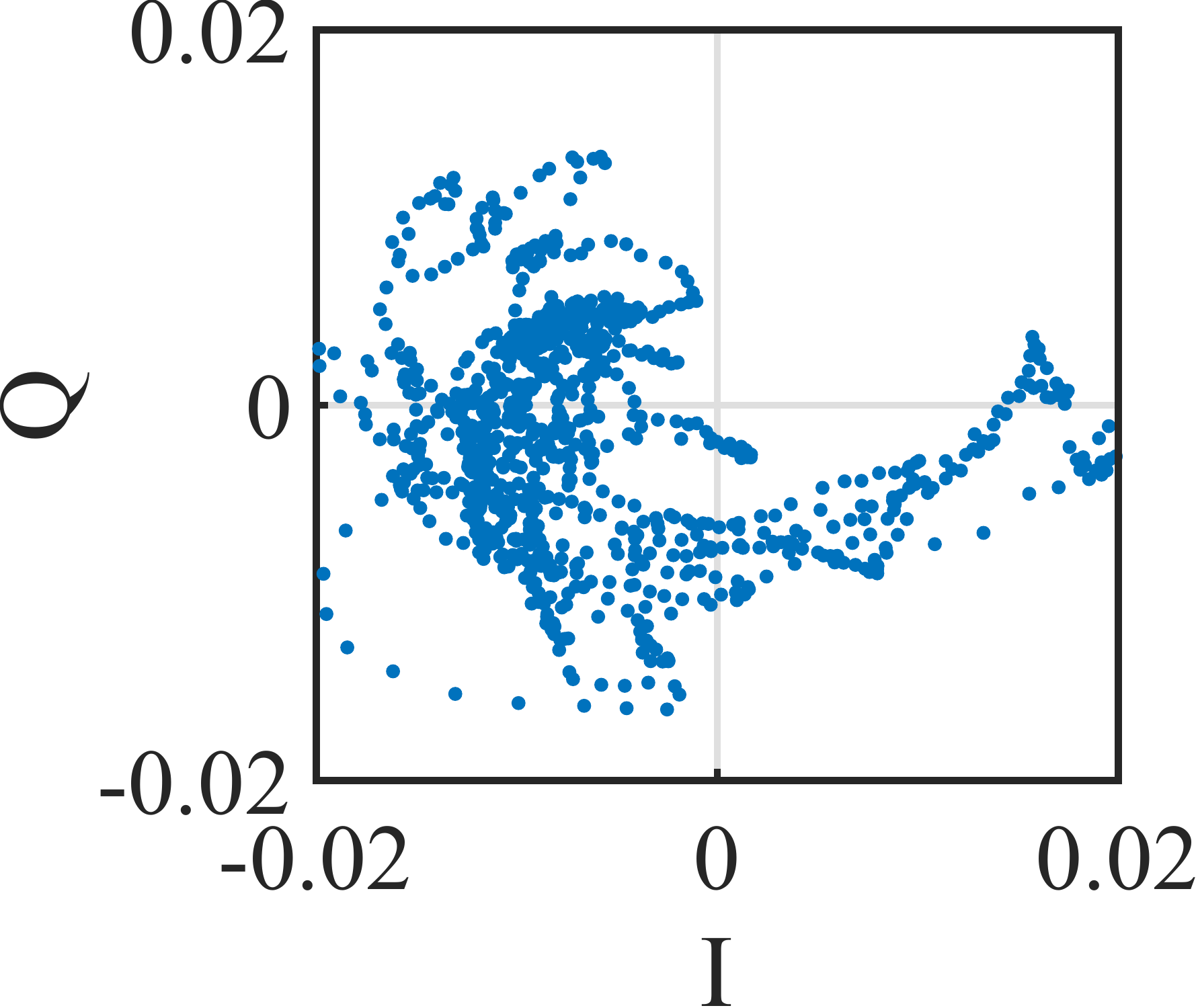}
			    \label{subfig:aug300}
			\end{minipage}
		}
		\caption{Augmenting data by rotating a slow-time row of $\hat{\boldsymbol{r}}(n)$ in I/Q domain. 
		}
		\label{fig:aug}
	    \vspace{-2ex}
\end{figure}

Before putting the I/Q signal contained in $\hat{\boldsymbol{r}}(n)$ 
into deep analytics, it is necessary to perform data augmentation for the following reasons. On one hand, data collection is highly non-trivial because one has to coordinate among human subjects, data recording of the IR-UWB radar, and the wearable ground truth sensor. Therefore, it is desirable to increase the diversity of a dataset by applying certain transformations.
On the other hand, data augmentation often helps a deep neural network comprehend intrinsic structures of the raw data. \rev{For \name, this \rev{I/Q-induced} intrinsic structure 
is non-trivially preserved only by rotation but not other transforms such as translation.} Therefore, we propose to augment $\hat{\boldsymbol{r}}(n)$ by rotating its every complex element in I/Q domain:
\begin{equation}
\left[\begin{array}{l}
r_\mathrm{I}^{\text{aug}}\\
r_\mathrm{Q}^{\text{aug}}
\end{array}\right]=\left[\begin{array}{cc}
\cos \theta & -\sin \theta \\
\sin \theta & \cos \theta
\end{array}\right]\left[\begin{array}{l}
r_\mathrm{I} \\ 
r_\mathrm{Q}
\end{array}\right],
\end{equation}
where $\theta$ specifies a rotation angle and it is varied to achieve data augmentation. Figure~\ref{fig:aug} illustrates five versions of augmented $\hat{\boldsymbol{r}}(n)$: the rotation preserves the respiration traces, because it affects only the distance $d$ (which is anyway varying drastically under body movements, according to Section~\ref{ssec:interference}) but not the respiration-induced periodic motions $\Delta d$ of the chest. 
As shown in Figure~\ref{fig:aug}, the overlapped respiration ellipses maintain the overall distribution despite varying rotations. In practice, \name\ may choose to 
employ more rotation angles for better enriching a dataset.

\subsection{Fine-Grained Waveform Recovery}
In this section, we first study the background of Variational Encoder-Decoder (VED), then discuss how to adapt VED architecture for I/Q complex signals, and finally provide details on respiratory waveform recovery using IQ-VED.
\subsubsection{Design Rationale} 
\label{sssec:background} 
Extracting certain signals from a nonlinear signal mixture is highly non-trivial~\cite{hyv1999Noica, jutten2003advances}; the deep learning community has been employing an Encoder-Decoder (ED) network for this task~\cite{luo2018tasnet, mimilakis2017recurrent}. Unfortunately, the latent space of a regular ED network is not continuous given limited training data, so it lacks sufficient generalization ability when dealing with unseen data. Inspired by the idea of \textit{variational inference}~\cite{blei2017variational, hoffman2013stochastic}, 
we tackle the problem of latent space irregularity by forcing the encoder to return probability distributions %
rather than discrete vectors, and we name the modified network Variational Encoder-Decoder (VED). It is worth noting that our VED is fundamentally different from Variational AutoEncoder (VAE)~\cite{kingma2013auto, rezende2014stochastic}: \rev{whereas VED aims to extract signal from a nonlinear mixture, VAE intends to learn an efficient representation of the input.}

\rev{To achieve the objective of \name\ in respiratory waveform recovery, a regular ED learns an \textit{encoder} $q_{\phi}(\boldsymbol{z} | \boldsymbol{r})$ mapping input data $\boldsymbol{r}$ to a latent representation $\boldsymbol{z}$, and generates output $\boldsymbol{r'}$ (i.e., respiratory waveform) by a \textit{decoder} $p_{\psi}(\boldsymbol{r'}|\boldsymbol{z})$. In other words, $\boldsymbol{z}$ represents the partial features extracted from $\boldsymbol{r}$ to characterize only $\boldsymbol{r'}$.} VED shares the pair of encoder $\phi$ and decoder $\psi$ with ED, but it maps $\boldsymbol{r}$ to a Gaussian distribution parameterized by a mean and variance. Essentially, the generative process of the VED is enabled
by maximizing the variational lower bound (VLB)~\cite{kingma2013auto}:
\begin{align}
\!\!\!\log p_{\psi}(\boldsymbol{r'}) \geq&~\mathrm{VLB}_{\mathrm{VED}}(\boldsymbol{r}, \boldsymbol{r'}; \psi, \phi) \nonumber \\
 =&~\mathbb{E}_{q_{\phi}(\boldsymbol{z} | \boldsymbol{r})}\left[\log p_{\psi}(\boldsymbol{r'} | \boldsymbol{z})\right] - \mathbf{D}_\mathrm{KL}\left(q_{\phi}(\boldsymbol{z} | \boldsymbol{r}) \| p_{\psi}(\boldsymbol{z})\right), \label{eq:elbo}
\end{align}
where $p_{\psi}(\boldsymbol{z}) = \mathcal{N}\left(\mathbf{0}, \mathbf{I}\right)$ is a Gaussian prior on the latent representation $\boldsymbol{z}$ and $\mathbf{D}_\mathrm{KL}(\cdot)$ donotes the Kullback-Leibler (KL) divergence~\cite{kullback1951information}; it works as a regularizer by minimizing the difference between $q_{\phi}(\boldsymbol{z} | \boldsymbol{r})$ and $p_{\psi}(\boldsymbol{z})$. In this way, VED gets around the hardness in estimating the (distribution) of $\boldsymbol{r'}$ directly from $\boldsymbol{r}$, by using the latent representation $\boldsymbol{z}$ as an intermediate relay. 
Moreover, representing $\boldsymbol{z}$ as a (continuous) probability distribution rather than a discrete vector set, VED is equipped with a continuous latent space. Upon unseen inputs, this latent space will be sampled in a more meaningful manner than that of a conventional ED. Essentially, the continuous property of the latent space enables VED to avoid overfitting and hence better handle out-of-range inputs.

There is yet one link missing before applying VED to separate respiration from I/Q-represented RF signal mixture: most building blocks for deep learning are based on real-valued operations and representations, how to reform VED to handle complex I/Q signals remains a problem. Previously, deep complex networks~\cite{deepcomplex,potok2018study} have been proposed to handle complex numbers, but they require redefining calculus operations including differentiation crucial to backpropagation~\cite{deepcomplex}, so they are in general hard to train (with super slow convergence) and hence not widely adopted. \rev{The same convergence and complexity issues also apply to neural networks for sequential processing, such as general RNNs that include LSTM~\cite{trinh2018learning}.} 
Consequently, our IQ-VED performs a bivariate analysis of the I/Q signal, as explained in Section~\ref{ssec:enc}.

\begin{figure*}[t]
    \setlength\abovecaptionskip{8pt}
    \centering
	\includegraphics[width=.88\linewidth]{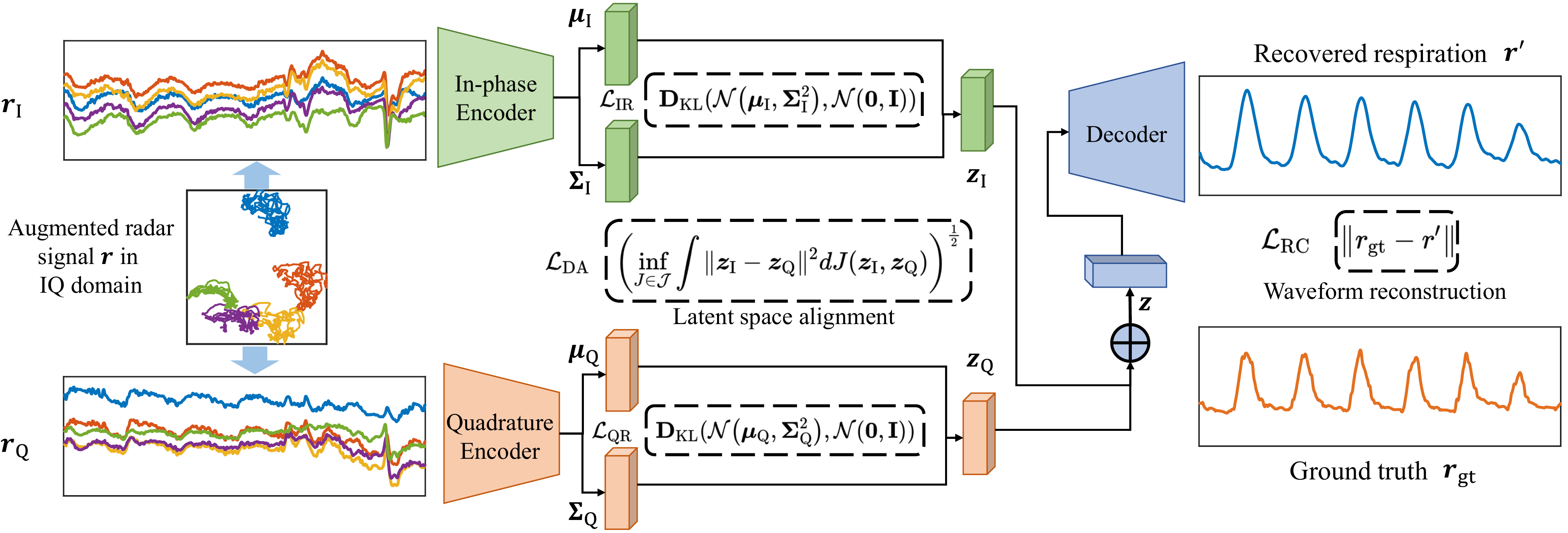}
	\caption{IQ-VED architecture: it takes in I/Q data as two streams and encodes them separately. The resulting latent representations are aligned and fed to the decoder to recover respiratory waveform by minimizing the reconstruction error. }
	\label{fig:iqved}
	\vspace{-1ex}
\end{figure*}
\subsubsection{IQ-VED Encoder} \label{ssec:enc}
\begin{figure}[t]
    \setlength\abovecaptionskip{8pt}
    \vspace{-1ex}
	   \captionsetup[subfigure]{justification=centering}
		\centering
		\subfloat[Encoder (both I and Q).]{
		    \begin{minipage}[b]{0.47\linewidth}
		        \centering
			    \includegraphics[width = 0.7\textwidth]{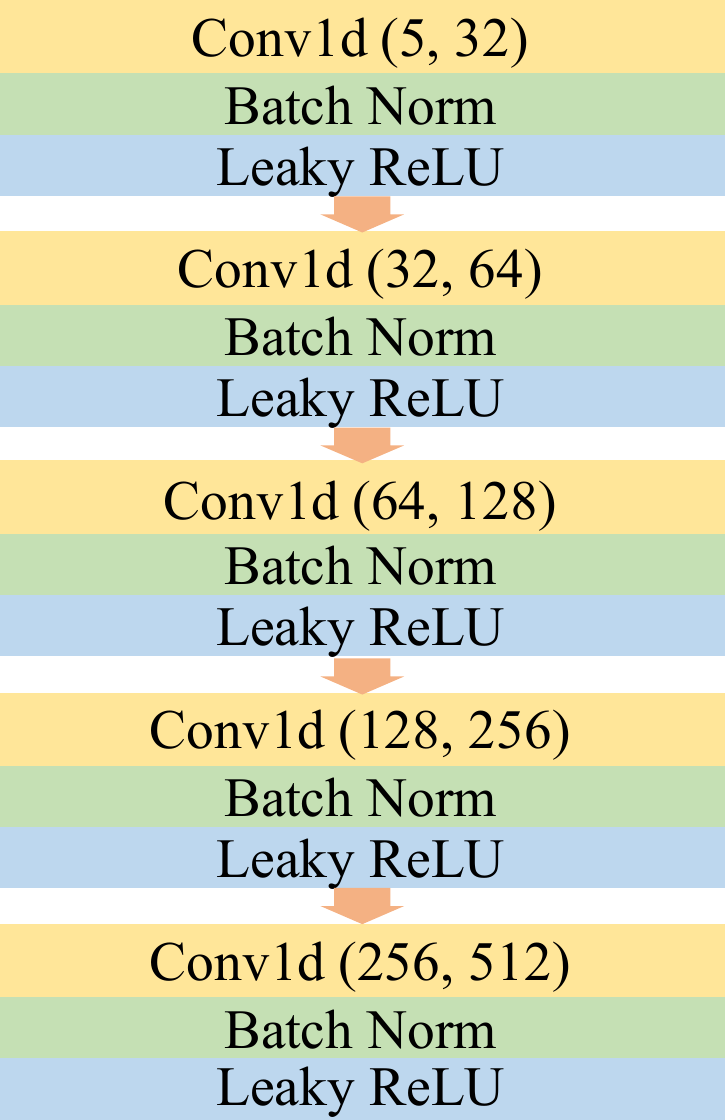}
			    \label{subfig:encoder}
			\end{minipage}
		}
		\subfloat[Decoder.]{
		  \begin{minipage}[b]{0.47\linewidth}
		        \centering
			    \includegraphics[width = 0.7\textwidth]{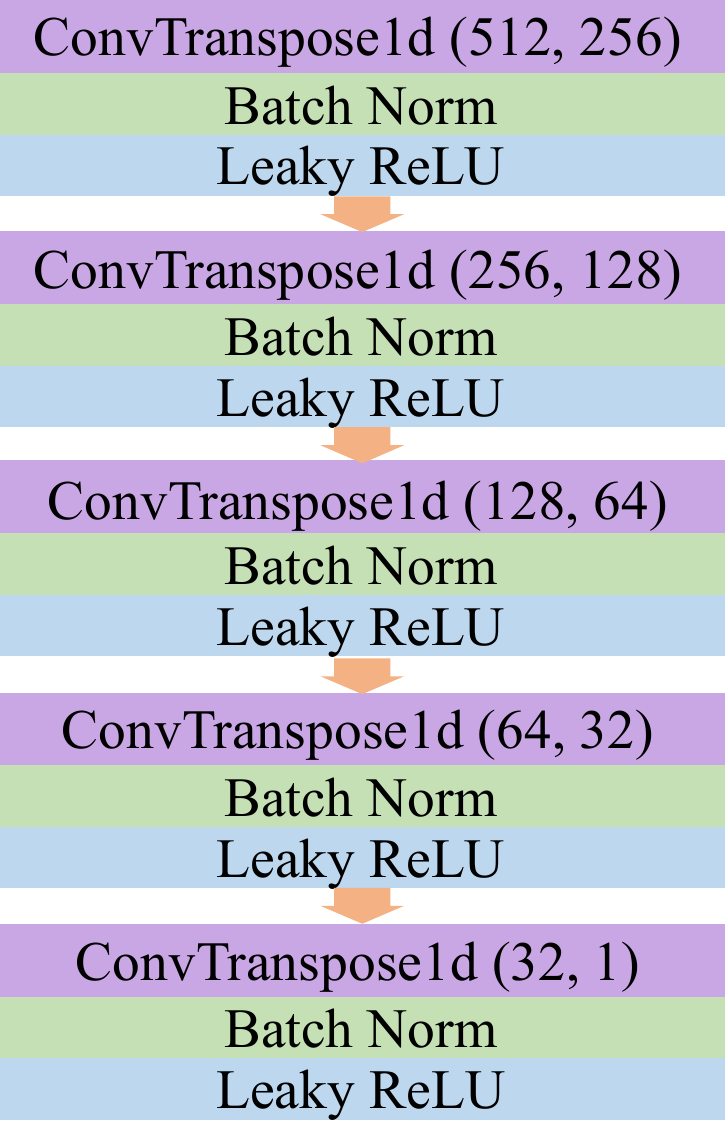}
			    \label{subfig:decoder}
			\end{minipage}
		}
		\caption{Encoder and decoder configurations. Two values in parentheses indicate the amounts of input and output channels, respectively.}
		\label{fig:encoder_decoder}
	    \vspace{-2ex}
\end{figure}

The encoder of IQ-VED takes in the I/Q signal matrix $\boldsymbol{r}(n)$ ($\hat{\boldsymbol{r}}(n)$ in Section~\ref{ssec:loc} for brevity) and encodes it to a latent representation $\boldsymbol{z}$. Specifically, IQ-VED adopts a two-stream design, where the I/Q components $\boldsymbol{r}_{\mathrm{I}}(n)$ and $\boldsymbol{r}_{\mathrm{Q}}(n)$ are fed into two separate encoders, as shown in Figure~\ref{fig:iqved}.
Each encoder consists of i) multiple layers of One-Dimensional Convolutional Neural Network (1D-CNN)~\cite{lecun1999object} for feature extraction, ii) batch norm layers~\cite{ioffe2015batch} for normalization, and iii) leaky ReLU~\cite{glorot2011deep} layers for adding non-linearity, as illustrated in Figure~\ref{subfig:encoder}. Both $\boldsymbol{r}_{\mathrm{I}}(n)$ and $\boldsymbol{r}_{\mathrm{Q}}(n)$ are treated as multi-channel 1-D sequences, with each channel corresponding to one fast-time index in $\boldsymbol{r}(n)$. 

\rev{Essentially, the IQ-VED encoder decomposes the input I/Q signals and filters out motion interference. The resulting respiration-induced signal is compressed and mapped to the latent distribution, which will then be sampled to drive the decoder so as to recover the desired respiratory waveform. The overall convolutional filter aims to extract useful features, so it can be deemed as a demixing function~\cite{khemakhem2020variational} to reverse the entanglement between respiration signal and non-linear motion interference.} 
It is well known that processing the I/Q components of complex signals separately, though substantially lowering the training complexity, may cause misalignment. To overcome this problem, we specifically align their respective latent spaces in Section~\ref{sssec:alignment}.

\subsubsection{Latent Space Alignment}\label{sssec:alignment}
The outputs of the encoder are two Gaussian distributions $\boldsymbol{z}_{\mathrm{I}} \sim \mathcal{N}(\boldsymbol{\mu}_\mathrm{I},\boldsymbol{\Sigma}^2_\mathrm{I}) $ and $\boldsymbol{z}_{\mathrm{Q}} \sim\mathcal{N}(\boldsymbol{\mu}_\mathrm{Q},\boldsymbol{\Sigma}^2_\mathrm{Q}) $ parameterized by respective means and variances, according to Equation~\eqref{eq:elbo}.
Since both latent distributions are integral parts of the complex signal representation, IQ-VED should guarantee that their processing (via individual encoders) has been conducted in a coordinated manner. Fortunately, since both in-phase and quadrature signals are 1-D perspectives of the same complex radar signal, they share common structures sufficient to align their corresponding latent representations. 
To this end, we choose to minimize the 2-Wasserstein distance~\cite{olkin1982distance} between them:
\begin{equation}\label{eq:was_inf}
\mathbf{W}_{\mathrm{IQ}}=\left(\inf _{J \in \mathcal{J}(\mathcal{N}(\boldsymbol{\mu}_\mathrm{I},\boldsymbol{\Sigma}^2_\mathrm{I}), \mathcal{N}(\boldsymbol{\mu}_\mathrm{Q},\boldsymbol{\Sigma}^2_\mathrm{Q}))} \int\|\boldsymbol{z}_{\mathrm{I}}-\boldsymbol{z}_{\mathrm{Q}}\|^{2} d J(\boldsymbol{z}_{\mathrm{I}}, \boldsymbol{z}_{\mathrm{Q}})\right)^{\frac{1}{2}},
\end{equation}
where $\mathcal{J}$ denotes the set of all joint distributions $J$ that has $\mathcal{N}(\boldsymbol{\mu}_\mathrm{I},\boldsymbol{\Sigma}^2_\mathrm{I})$ and $\mathcal{N}(\boldsymbol{\mu}_\mathrm{Q},\boldsymbol{\Sigma}^2_\mathrm{Q})$ as respective marginals. The reason for employing the Wasserstein distance is twofold. On one hand, minimizing the distance shifts the two distributions ``close'' to each other, enforcing them to encode the same respiration features.
On the other hand, unlike KL divergence, Wasserstein distance is able to provide a useful gradient when the distributions are not overlapping~\cite{kolouri2018sliced}. As a result, while most parts of the two distributions are meant to be aligned, some discrepancies inherent to the I/Q components, e.g., amplitude and phase of BBR offset are allowed to be maintained. 

In the case of multivariate Gaussian distributions, a closed-form solution of Equation~\eqref{eq:was_inf} can be obtained according to~\cite{givens1984class}:
\begin{equation}\label{eq:wasserstein}
    \mathbf{W}_{\mathrm{IQ}} = \left[\left\|\boldsymbol{\mu}_{\mathrm{I}}-\boldsymbol{\mu}_{\mathrm{Q}}\right\|_{2}^{2}\right. \left.+  \operatorname{Tr}\left(\boldsymbol{\Sigma}_{\mathrm{I}}\right)+\operatorname{Tr}\left(\boldsymbol{\Sigma}_{\mathrm{Q}}\right)-2\left(\boldsymbol{\Sigma}_{\mathrm{I}}^{\frac{1}{2}} \boldsymbol{\Sigma}_{\mathrm{I}} \boldsymbol{\Sigma}_{\mathrm{Q}}^{\frac{1}{2}}\right)^{\frac{1}{2}}\right]^{\frac{1}{2}}.
\end{equation}
Since the covariance matrices obtained by IQ-VED are of diagonal form, Equation~\eqref{eq:wasserstein} can be simplified as follows:
\begin{equation}
    \mathbf{W}_{\mathrm{IQ}} = \left\|\boldsymbol{\mu}_{\mathrm{I}}-\boldsymbol{\mu}_{\mathrm{Q}}\right\|_{2}^{2}+\left\|\boldsymbol{\Sigma}_{\mathrm{I}}^{\frac{1}{2}}-\boldsymbol{\Sigma}_{\mathrm{Q}}^{\frac{1}{2}}\right\|_{\text {Frob }}^{2}, 
    \label{eq:was_final}
\end{equation}
where $\left\|\cdot \right\|_{\text {Frob }}$ is the Frobenius norm, defined as the square root of the sum of the absolute squares of the matrix elements.

\subsubsection{IQ-VED Decoder and Loss Function}
\label{sssec:loss}
As shown in Figure~\ref{subfig:decoder}, the decoder can be deemed as the reverse of the encoder. %
To this end, we replace 1D-CNN in the decoder with 1-D transposed convolutional layers~\cite{zeiler2010deconvolutional} to upsample the latent representation and map them to a longer sequence, so as to finally derive respiratory waveform $\boldsymbol{r}^{\prime}(n)$. Note that the encoder and decoder are not exactly symmetric: at the last stage of the decoder, a single-channel signal is recovered, instead of a multi-channel one as the input to the encoder. %
To train IQ-VED, we employ three loss functions, namely the reconstruction loss, the I/Q regularizing loss, and the distribution alignment loss. %

\paragraph{Reconstruction Loss}
To correctly recover respiratory waveform, this loss function compares the output of IQ-VED decoder with the ground truth obtained by a wearable sensor, and tries to make them similar. $L^2$ loss is used to define the reconstruction loss $\mathcal{L}_{R C}$, which measures the sum of all the squared differences between the two waveforms. This loss practically implements the term $\mathbb{E}_{q_{\phi}(\boldsymbol{z} | \boldsymbol{r})}\left[\log p_{\psi}(\boldsymbol{r'} | \boldsymbol{z})\right]$ in Equation~\eqref{eq:elbo}:
\begin{equation}
    \mathcal{L}_{\mathrm{R C}} = \quad \left\|r_{\mathrm{gt}}-r'\right\|.
\end{equation}

\paragraph{I/Q Regularizing Loss}
In Section~\ref{sssec:background}, it is pointed out that VED regularizes the latent distribution according to a standard Gaussian prior. For IQ-VED, two distributions from the I/Q encoders should be regularized. The I/Q regularizing losses are defined as:
\begin{align}
    \mathcal{L}_{\mathrm{IR}} &= \mathbf{D}_\mathrm{KL} \left(\mathcal{N}\left(\boldsymbol{\mu}_{\mathrm{I}}, \boldsymbol{\Sigma}_{\mathrm{I}}^{2}\right), \mathcal{N}(\mathbf{0}, \mathbf{I})\right),\\
    \mathcal{L}_{\mathrm{QR}} &= \mathbf{D}_\mathrm{KL} \left(\mathcal{N}\left(\boldsymbol{\mu}_{\mathrm{Q}}, \boldsymbol{\Sigma}_{\mathrm{Q}}^{2}\right), \mathcal{N}(\mathbf{0}, \mathbf{I})\right).
\end{align}

\paragraph{Distribution Alignment Loss}
As described in Section~\ref{sssec:alignment}, the misalignment between the two distributions from the I/Q encoders can be measured by a Wasserstein distance. Therefore, the distribution alignment loss is defined according to Equation~\eqref{eq:was_final}:
\begin{equation}
    \mathcal{L}_{\mathrm{DA}} = \mathbf{W}_{\mathrm{IQ}} =  \left\|\boldsymbol{\mu}_{\mathrm{I}}-\boldsymbol{\mu}_{\mathrm{Q}}\right\|_{2}^{2}+\left\|\boldsymbol{\Sigma}_{\mathrm{I}}^{\frac{1}{2}}-\boldsymbol{\Sigma}_{\mathrm{Q}}^{\frac{1}{2}}\right\|_{\text {Frob }}^{2}.
\end{equation}
Combining these loss functions, the overall loss function for training IQ-VED can be obtained as follows:
\begin{equation}\label{eq:loss}
    \mathcal{L}_{\mathrm{IQ-VED}}=\mathcal{L}_{\mathrm{RC}} + \gamma (\mathcal{L}_{\mathrm{IR}} +  \mathcal{L}_{\mathrm{QR}}) + \eta \mathcal{L}_{\mathrm{DA}},
\end{equation}
where $\gamma$ and $\eta$ are the respective weights for regularizing losses and distribution alignment loss. $\mathcal{L}_{\mathrm{IR}}$ and $\mathcal{L}_{\mathrm{QR}}$ share the same weight $\gamma$ because the I/Q data and encoders are symmetric and of equal importance. In addition, $\gamma$ is supposed to be greater than 1, so as to emphasize the representation capacity of the latent variables $\boldsymbol{z}$ and encourage disentanglement of the representations~\cite{higgins2016beta}, as will be confirmed in Section~\ref{sssec:impact_weight}.

\subsubsection{Waveform Recovery and Biomarker Recognition} \label{sssec:waveinf}
With a well-trained IQ-VED, respiratory waveform can be recovered from radar signal even under motion interference. One immediate way of applying IQ-VED to waveform recovery is to re-sample the latent vectors $\boldsymbol{z}_{\mathrm{I}}$ and $\boldsymbol{z}_{\mathrm{Q}}$ from $\mathcal{N}\left(\boldsymbol{\mu}_{\mathrm{I}}, \boldsymbol{\Sigma}_{\mathrm{I}}^{2}\right)$ and $\mathcal{N}\left(\boldsymbol{\mu}_{\mathrm{Q}}, \boldsymbol{\Sigma}_{\mathrm{Q}}^{2}\right)$. However, this leads to non-deterministic outputs that may cause problem in practice. 
To tackle this problem, we perform a deterministic inference without sampling $\boldsymbol{z}_{\mathrm{I}}$ and $\boldsymbol{z}_{\mathrm{Q}}$ as follows:
\begin{eqnarray}
    \boldsymbol{r'}^{*} 
    &=& \arg\max_{\boldsymbol{r'}} p_{\psi}\left(\boldsymbol{r'} | \boldsymbol{r}_\mathrm{I}, \boldsymbol{r}_\mathrm{Q}, \boldsymbol{z}_{\mathrm{I}}^{*}+\boldsymbol{z}_{\mathrm{Q}}^{*}\right),
\end{eqnarray}
where the deterministic latent vectors $\boldsymbol{z}_{\mathrm{I}}^{*}$ and $\boldsymbol{z}_{\mathrm{Q}}^{*}$ are obtained as $\boldsymbol{z}_{\mathrm{I}}^{*}=\mathbb{E}[\boldsymbol{z}_{\mathrm{I}}|\boldsymbol{r}_\mathrm{I}]$ and $\boldsymbol{z}_{\mathrm{Q}}^{*}=\mathbb{E}[\boldsymbol{z}_{\mathrm{Q}}|\boldsymbol{r}_\mathrm{Q}]$.

\begin{figure}[t]
    \setlength\abovecaptionskip{8pt}
    \vspace{-1ex}
	   \captionsetup[subfigure]{justification=centering}
		\centering
		\subfloat[Time domain.]{
		    \begin{minipage}[b]{0.47\linewidth}
		        \centering
			    \includegraphics[width = 0.96\textwidth]{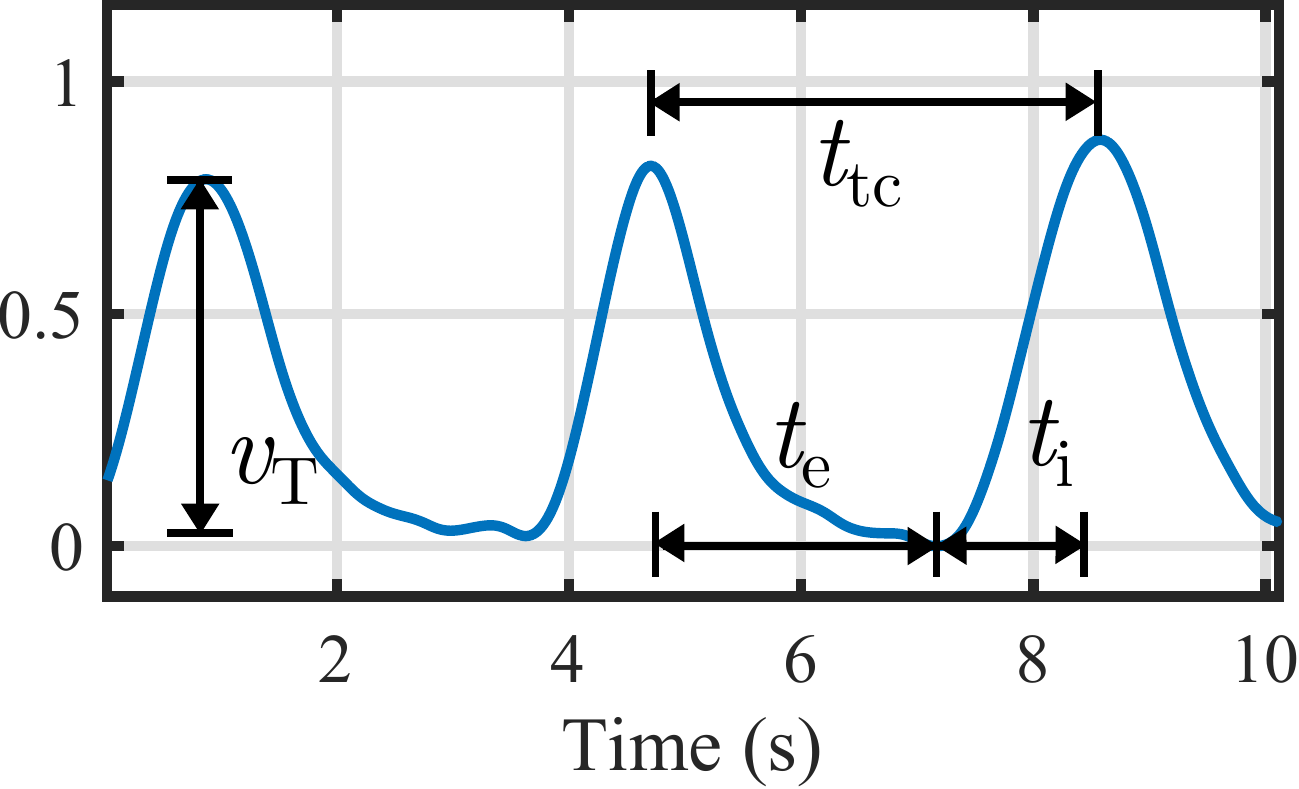}
			    \label{subfig:breath_t}
			\end{minipage}
		}
		\subfloat[Frequency domain.]{
		  \begin{minipage}[b]{0.47\linewidth}
		        \centering
			    \includegraphics[width = 0.96\textwidth]{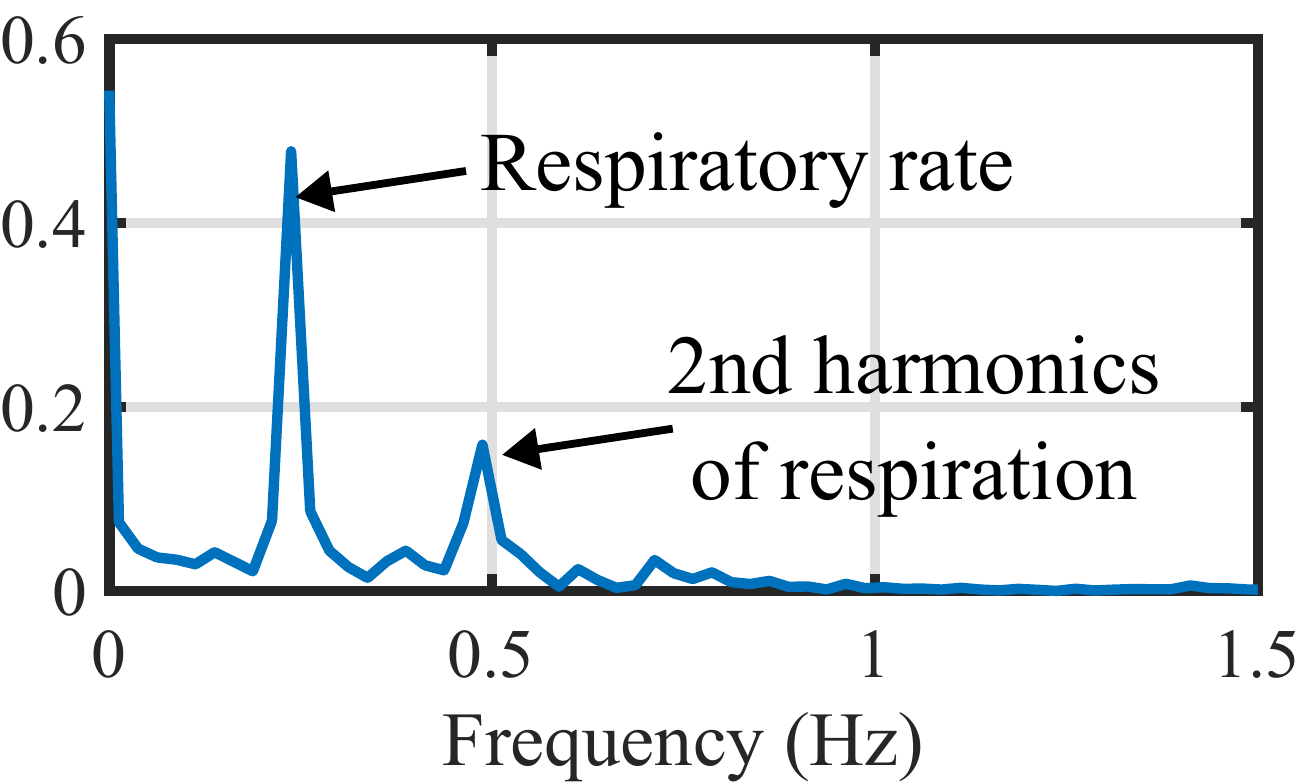}
			    \label{subfig:breath_f}
			\end{minipage}
		}
		\caption{The recovered respiration signal while exercising on the spot outputted by IQ-VED in different domains.}
		\label{fig:rate}
	    \vspace{-2ex}
\end{figure}

To determine the respiratory rate based on the recovered waveform, Fast Fourier Transform (FFT) can be applied.
Since the frequency of respiration ranges from 0.16~$\!$Hz to 0.6~$\!$Hz, the search space can be narrowed down, and the peak frequency in the range can be identified as the respiratory rate. As an example, a sample respiration given subject exercising on the spot is shown in Figure~\ref{fig:rate}, with both time and frequency representations. In particular, Figure~\ref{subfig:breath_f} shows that the respiratory rate can be estimated to be 0.24~\!Hz, or approximately 14.4 beats per minute (bpm). In addition, the instantaneous frequency can also be obtained by taking the reciprocal of the total cycle time.

Apart from respiratory rate, other biomarkers can also be inferred, such as tidal volume ($v_\mathrm{T}$) denoting the amount of air that moves in or out of the lungs with each respiratory cycle, total cycle time $t_{\mathrm{tc}}$ measuring the total time for a respiration cycle (peak to peak),  inspiratory time $t_{\mathrm{i}}$ indicating the time of inhaling in a cycle (valley to peak), expiratory time $t_{\mathrm{e}}$ indicating the time of exhaling in a cycle (peak to valley), and inhalation/exhalation ratio (I/E ratio) representing a compromise between ventilation and oxygenation. By finding the peaks and valleys, and their corresponding timestamp in the recovered waveform, these biomarkers can be calculated. We illustrate all the aforementioned time-related biomarkers in Figure~\ref{subfig:breath_t}.

\section{Implementation}\label{sec:implementation}
\paragraph{Hardware Implementations}
Our \name\ prototype leverages IR-UWB signals for monitoring human respiration.  The core component of \name\ is a compact and low-cost Novelda X4M05~\cite{xethru} IR-UWB radar transceiver. The radar operates at a center frequency of 7.29 or 8.7~\!GHz with a bandwidth of 1.5~\!GHz. The sampling rate of the radar is 23.328~\!GHz, and the frame rate is set to 50~\!fps. The radar has a pair of tx-rx antennas with an FoV of $65^{\circ}$ in both azimuth and elevation angles. A Raspberry Pi single-board computer~\cite{rpi} is used to control the transceiver and to interface with a desktop computer; this computer has an Intel Xeon W-2133 CPU, 16~\!GB RAM, and a GeForce RTX 2080 Ti graphics card. NeuLog respiration monitor belt logger sensor NUL-236~\cite{neulog} is used to collect ground truth respiratory waveform, the sampling rate of the NeuLog sensor is also set to 50~\!fps, the same as the radar.

\paragraph{Software Implementations}
We implement \name\ based on Python~3.7 and C/C++, with the neural network components built upon PyTorch 1.7.1~\cite{pytorch}. To align the ground truth respiration signal from the respiration monitor belt logger and radar signals, the Precision Time Protocol~\cite{ptp} relying on message exchanges over Ethernet is used to synchronize the clocks between hardware components. 
In the data augmentation process, each signal $\boldsymbol{r}(n)$ is rotated from $0$ to $2\pi$ with an interval of $\pi/30$ for 60 times. The parameters of IQ-VED are set as follows: $\gamma$ and $\eta$ in Equation~\eqref{eq:loss} are set to 3 and 2e-4, respectively. For the encoder, 5 consecutive 1-D convolutional layers are used, whose kernel size is set to 3, stride to 1, padding size to 1; and the number of output channels of these convolutional layers is set to 32, 64, 128, 256, and 512. %
As for the decoder, 5 consecutive transposed convolutional layers are used, their kernel size is set to 3, stride to 1, scale factor to 2, and the number of output channels of these layers is set to 512, 256, 128, 64, and 32. %
All weights are initialized by the Xavier uniform initializer~\cite{kumar2017weight}. 
\rev{%
Consequently, IQ-VED involves $2.36\times 10^7$ parameters; it incurs $1.21\times10^8$ multiply-accumulate operations for each signal instance during inference.} 
The collected dataset is divided into training and test sets. The training set contains 8,000 pairs of radar signal matrices and respiration ground truths obtained from NeuLog sensor, and the test set contains 4,000 pairs. The size of the raw training radar signal matrix is $1000 \times 138$. For the training process, the batch size is set to 64, the IQ-VED loss in Equation~\eqref{eq:loss} is adopted, and the learning rate and momentum of the Stochastic Gradient Descent optimizer~\cite{bottou2012stochastic} are respectively set to 0.01 and 0.9.

\section{Evaluation}\label{sec:evaluation}
In this section, we perform intensive evaluations on the performance of \name\ given several real-life scenarios and under various parameter settings.

\subsection{Experiment Setup}\label{ssec:setup}
In order to conduct the evaluations, we recruit 12 volunteers~(6 females and 6 males), aged from 15 to 64, and weighing from 50 to 80~\!kg. All volunteers are healthy, and we measure their respiration in the natural state
without the volunteers consciously controlling breathing or undergoing external forceful intervention. The volunteers are asked to carry out 7 common activities with different degrees of body movements: playing on phone, typewriting, exercising on the spot, shaking legs, walking on a treadmill, standing up/sitting down, and turning over in bed, %
all in real-life environments such as office, gym, and bedroom. For brevity, the names of these movements are abbreviated as PP, TW, ES, SL, WT, SS, and TO, respectively. All experiments have strictly followed the standard procedures of IRB of our institute.

The IR-UWB radar is placed to face a human subject, within a range of 0.5 to 2\!~m, and on the same height as the chest of the subject. We collect a larger number of data entries for shorter activities (e.g. walking on treadmill that costs several minutes) and a smaller number of data entries for longer activities (overnight sleeping with turning over in bed) so as to guarantee a balanced dataset. Our data collection leads to a 66-hour dataset of RF and ground truth recordings, including approximately 72,000 respiration cycles and roughly the same amount of data from every subject. After collection, both the raw and ground truth data
are sliced into 20~\!s samples. Two-thirds of the collected data are used for training IQ-VED, and the remaining one-third is used for testing the performance of \name\ in recovering respiratory waveform.

\subsection{Metric and Baseline Selection}

\subsubsection{Cosine Similarity}
The cosine similarity $\mathcal{S}(\boldsymbol{r}^{\prime}, \boldsymbol{r}_{\mathrm{gt}})$ between the IQ-VED recovered waveform $\boldsymbol{r}^{\prime}(n)$ and the ground truth $\boldsymbol{r}_{\mathrm{gt}}(n)$ is used to measure the recovering performance of IQ-VED. Specifically, the cosine similarity is measured by the cosine of the angle between two vectors $\boldsymbol{r}^{\prime}(n)$ and $\boldsymbol{r}_{\mathrm{gt}}(n)$, and then determines to what extent the two vectors point to the same ``direction'' in a high dimensional space. It is defined as follows:
\begin{equation}
\mathcal{S}(\boldsymbol{r}^{\prime}, \boldsymbol{r}_{\mathrm{gt}})=\frac{\boldsymbol{r}^{\prime} \cdot \boldsymbol{r}_{\mathrm{gt}}}{\|\boldsymbol{r}^{\prime}\|\|\boldsymbol{r}_{\mathrm{gt}}\|}=\frac{\sum_{i=1}^{N} \boldsymbol{r}^{\prime} (i) \boldsymbol{r}_{\mathrm{gt}}(i)}{\sqrt{\sum_{i=1}^{N} \boldsymbol{r}^{\prime 2} (i)} \sqrt{\sum_{i=1}^{N} \boldsymbol{r}_{\mathrm{gt}}^{2}(i)}},
\end{equation}
whose value lies in the range of $[0, 1]$.

\subsubsection{Time Estimation Error}
In Section~\ref{sssec:waveinf}, we have discussed several time-related biomarkers that can be deduced from the waveform, including $t_{\mathrm{tc}}$, $t_{\mathrm{i}}$, and $t_{\mathrm{e}}$. All of them can be calculated from the timestamps of the peaks and valleys in respiratory waveforms, so we study estimation errors in terms of the peak and valley times. Suppose the timestamps of the peak and valley are $t_{\mathrm{p}}$ and $t_{\mathrm{v}}$, then the errors are
defined as absolute differences between the estimated and actual values, i.e., $|t_{\mathrm{p}}^\mathrm{e}-t_{\mathrm{p}}^\mathrm{a}|$ and $|t_{\mathrm{v}}^\mathrm{e}-t_{\mathrm{v}}^\mathrm{a}|$.

\subsubsection{Respiratory Rate Estimation Error} \label{sssec:rrerr}
After recovering peaks and valleys in respiratory waveforms, the instantaneous respiratory rate $\rho$ can be estimated by taking the reciprocal of $t_{\mathrm{tc}}$. 
The error of respiratory rate is defined as the absolute difference between the estimated respiratory rate $\rho_\mathrm{R}^\mathrm{e}$ and the actual respiratory rate $\rho_\mathrm{R}^\mathrm{a}$, namely, $|\rho_\mathrm{R}^\mathrm{e}-\rho_\mathrm{R}^\mathrm{a}|$.

\subsubsection{Volume Estimation Error}
Besides the event times in the waveform, the amplitude of the waveform is also worthy of exploring, as it can be used to represent the tidal volume $v_{\mathrm{T}}$ of respiration given their proportional relation.
Since we are not interested in the absolute value of the waveform amplitude, we define the relative error of volume as the absolute difference between the estimated value and actual value divided by the actual value $|v_{\mathrm{T}}^\mathrm{e}-v_{\mathrm{T}}^\mathrm{a}|/v_{\mathrm{T}}^\mathrm{a}$.

\subsubsection{Baseline Selection}
BreathListener~\cite{xu2019breathlisener}, a respiration monitoring system for driving environments, is picked as a comparison baseline, because 
BreathListener also claims to take into account motion interference (albeit only small-scale ones) caused by running vehicles. However, we have to port BreathListener to radar as it was designed for acoustic sensing. Essentially, Breathlistener adopts a two-stage processing pipeline. It first employs EEMD~\cite{wu2009ensemble} to separate respiration from interference. As discussed in Section~\ref{ssec:conventional}, EEMD can only recover coarse-grained waveform (if not incorrect one). Therefore, it further applies a Generative Adversarial Network (GAN)~\cite{goodfellow2014generative}
for adding details to the recovered waveform.

\subsection{Performance Results}%
We start by performing an overall evaluation, showing intuitive waveform results and cosine similarities under different body movements. We then compare \name\ with a baseline method. Finally, we measure various estimation errors quantitatively.

\subsubsection{Overall Performance} 
\begin{figure}[t]
    \setlength\abovecaptionskip{8pt}
    \vspace{-2ex}
	   \captionsetup[subfigure]{justification=centering}
		\centering
		\subfloat[Typewriting (TW).]{
		    \begin{minipage}[b]{0.95\linewidth}
		        \centering
			    \includegraphics[width=.95\linewidth]{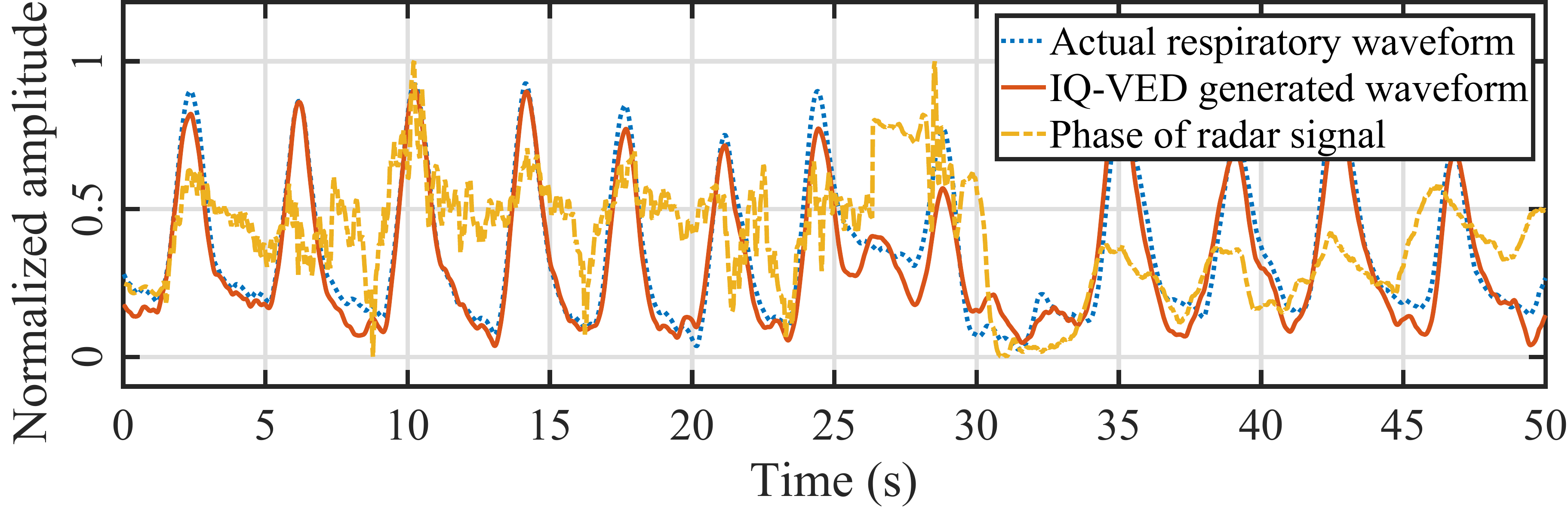}
			    \label{subfig:q_tw}
			    \vspace{-.5ex}
			\end{minipage}
		}
		\\\vspace{-.5ex}
		\subfloat[Exercising on the spot (ES).]{
		  \begin{minipage}[b]{0.95\linewidth}
		        \centering
			    \includegraphics[width=.95\linewidth]{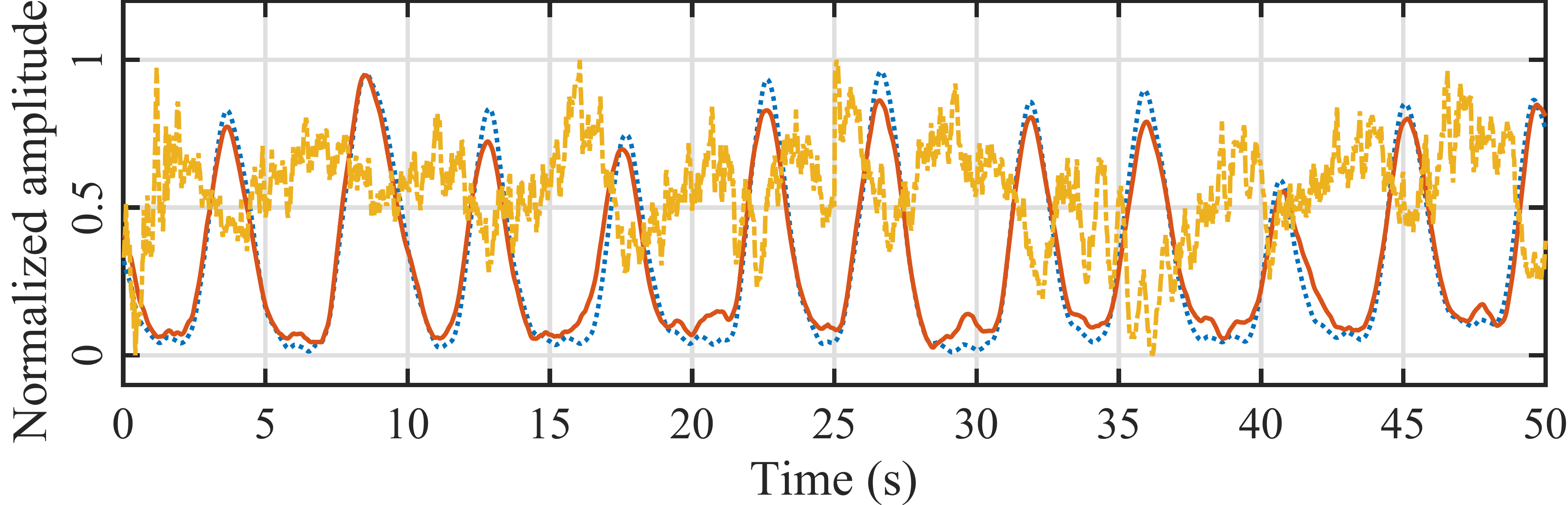}
			    \label{subfig:q_sw}
			    \vspace{-.5ex}
			\end{minipage}
		}
		\\\vspace{-.5ex}
		\subfloat[Walking on a treadmill (WT).]{
		  \begin{minipage}[b]{0.95\linewidth}
		        \centering
			    \includegraphics[width=.95\linewidth]{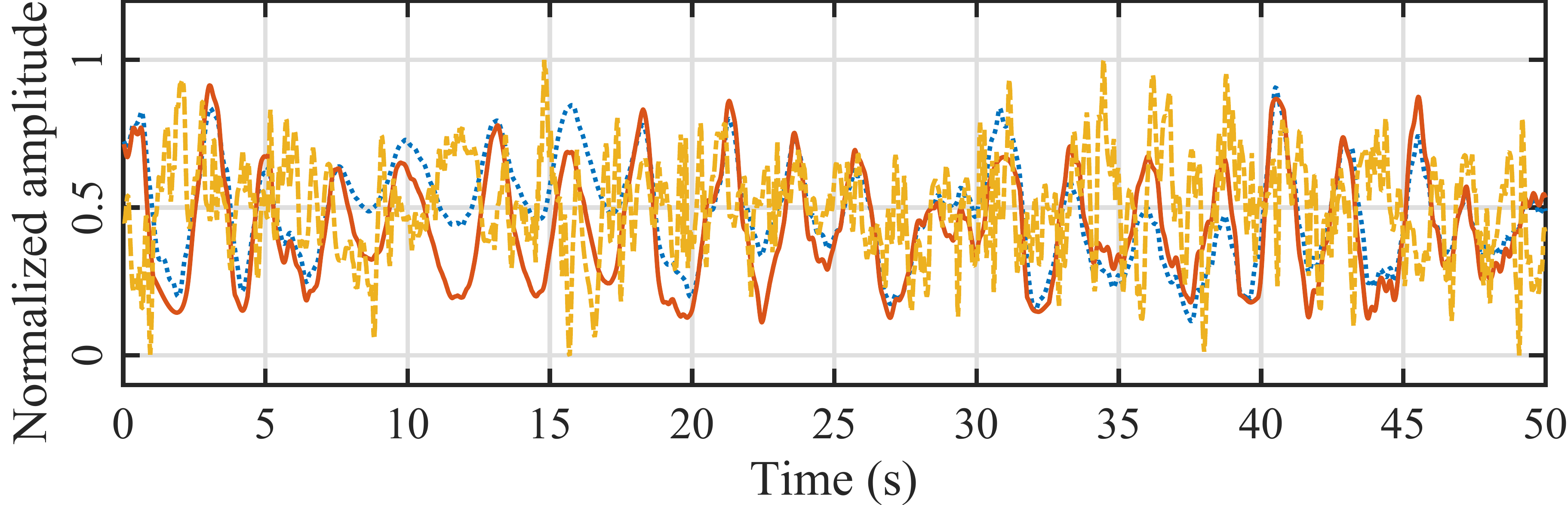}
			    \label{subfig:q_wt}
			    \vspace{-.5ex}
			\end{minipage}
		}
		\\\vspace{-.5ex}
		\subfloat[Turning over in bed (TO).]{
		  \begin{minipage}[b]{0.95\linewidth}
		        \centering
			    \includegraphics[width=.95\linewidth]{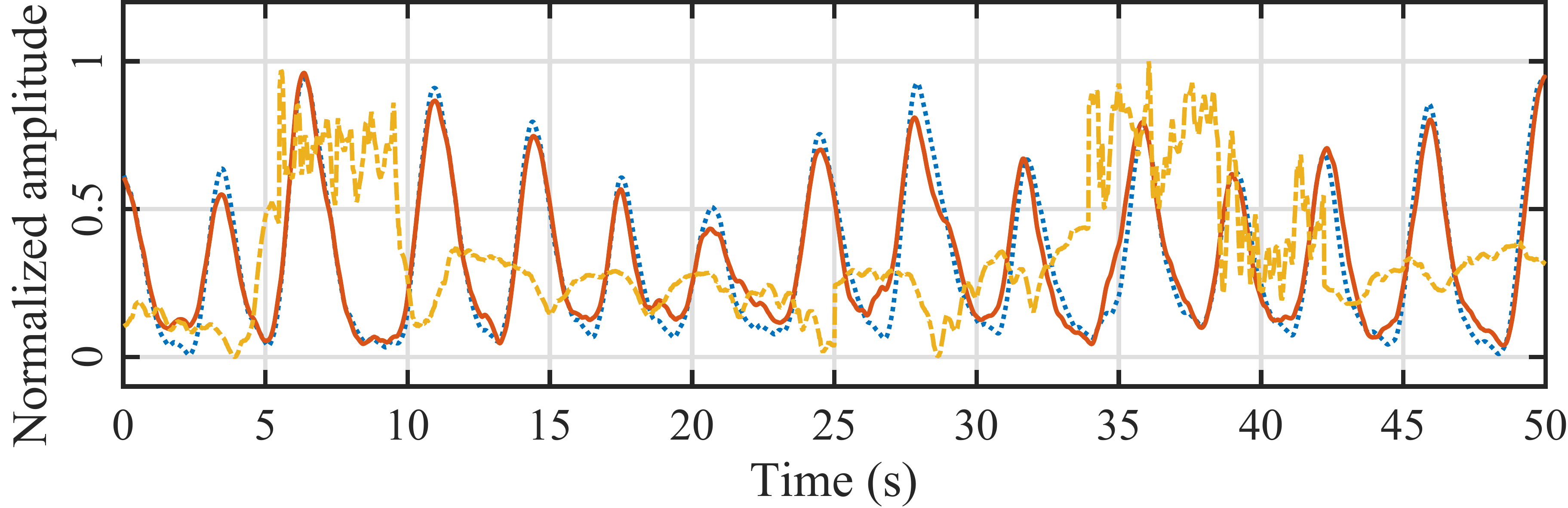}
			    \label{subfig:q_to}\vspace{-.5ex}
			\end{minipage}
		}
		\vspace{-.5ex}
		\caption{Qualitative result showing that \name\ recovers the respiratory waveform from the noisy radar signal under different body movements.}
		\label{fig:qualitative}
	    \vspace{-3ex}
\end{figure}

Figure~\ref{fig:qualitative} shows the respiratory waveform generated by IQ-VED compared to its corresponding ground truth versions during several 50~\!s activities.\footnote{Although our IQ-VED is trained with 20~\!s samples, its CNN-based encoder is flexible enough to accommodate an arbitrary sample length in practice.} In Figure~\ref{subfig:q_tw}, it can be observed that the motion interference caused by typewriting is sporadic, but the intensive interference period can affect the non-intensive one by shifting the phase of respiration significantly (the respiratory waveform can be roughly observed during the latter period). Fortunately, \name\ not
only recovers respiratory waveform during the intensive period, but also corrects the phase throughout the whole period, at a minor cost of waveform deforming during the transitional phase. Compared with Figure~\ref{subfig:q_tw}, both ES (Figure~\ref{subfig:q_sw}) and WT (Figure~\ref{subfig:q_wt}) incur more stationary motion interference with WT being much more intensive, clearly affecting respiration to a greater extent and causing deterioration in both respiratory waveforms $\boldsymbol{r}^{\prime}(n)$ and $\boldsymbol{r}_{\mathrm{gt}}(n)$. As the last example, Figure~\ref{subfig:q_to} shows that, though TO (during the time span of $[5, 10]$~\!s and $[34, 43]$~\!s) affects the signal phase significantly, IQ-VED successfully recovers respiration both when the human subject is lying quasi-statically and turning over.

\begin{figure}[b]
    \setlength\abovecaptionskip{8pt}
    \vspace{-2ex}
    \centering
	\includegraphics[width=.8\linewidth]{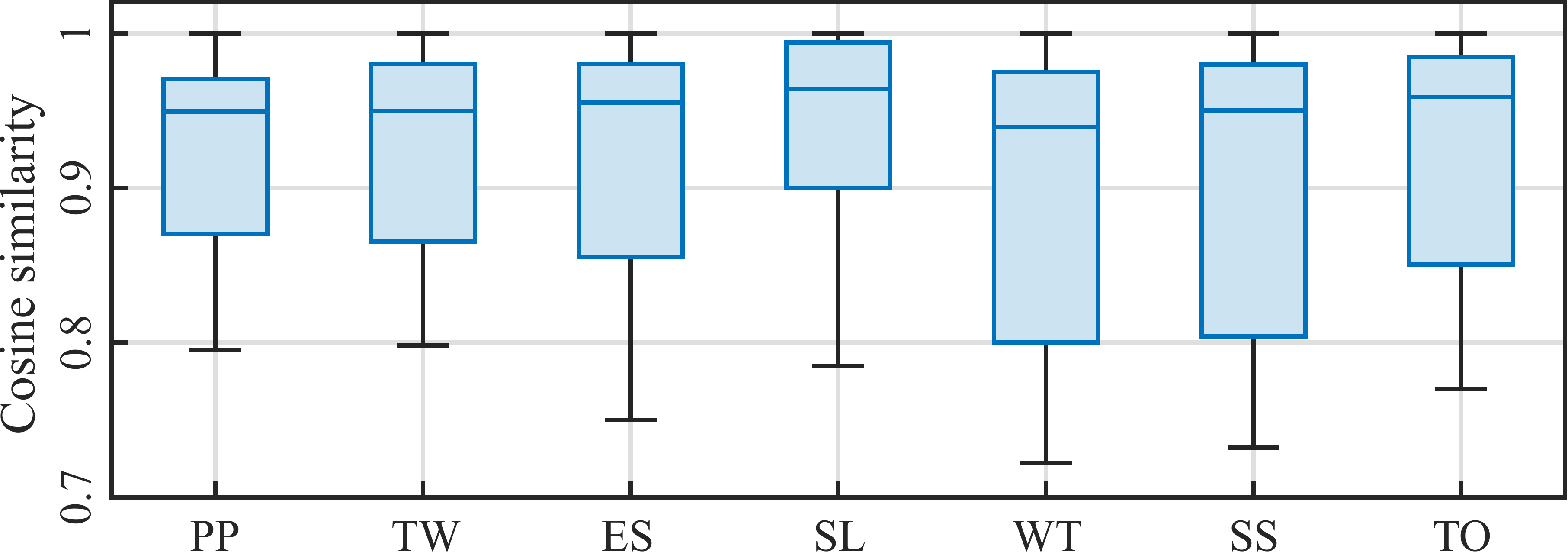}
	\caption{Cosine similarity between recovered waveform and ground truth under different body movements.}
	\label{fig:similarity}
\end{figure}

To further explore how individual body movement types affect the performance of \name, the cosine similarity under each body movement is studied in Figure~\ref{fig:similarity}. 
PP and SL can be observed as having the least impact on the \name\ performance, as the body parts involved in the movements are far from the subject's chest. 
As expected, WT and SS cause the worst performance of \name\, since they both induce large body movements that severely interfere with the respiration signal. Overall, the average cosine similarity between the recovered and ground truth respiratory waveform is 0.9162, indicating a very successful recovery, as a similarity greater than 0.8 suggests a strong positive correlation.

\subsubsection{Comparison with Baseline Method} \label{ssec:comp_bs}
\begin{figure}[b]
    \setlength\abovecaptionskip{8pt}
    \vspace{-2ex}
	   \captionsetup[subfigure]{justification=centering}
		\centering
	    \subfloat[Turning over in bed (TO).]{
		  \begin{minipage}[b]{\linewidth}
		        \centering
			    \includegraphics[width = 0.95\textwidth]{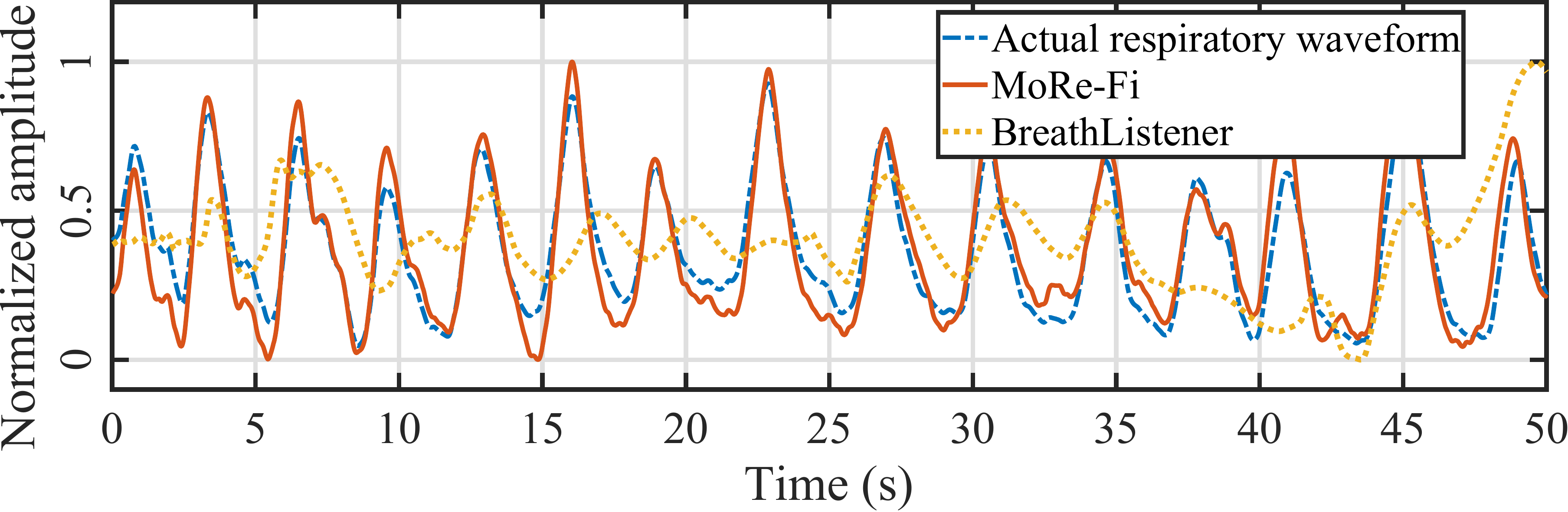}
			    \label{subfig:bs_waveform_to}
			\end{minipage}
		}
		\\%\vspace{-.5ex}%
	    \subfloat[Walking on a treadmill (WT).]{
		  \begin{minipage}[b]{\linewidth}
		        \centering
			    \includegraphics[width = 0.95\textwidth]{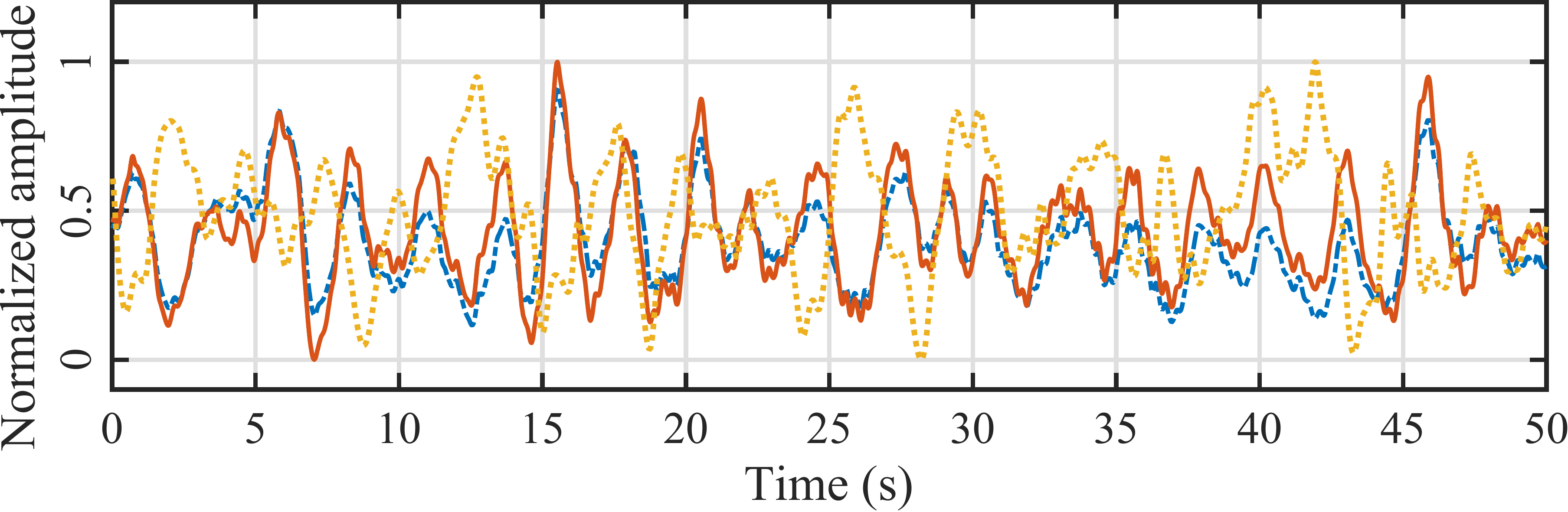}
			    \label{subfig:bs_waveform_wt}
			\end{minipage}
		}
		\\%\vspace{-.5ex}%
		\subfloat[Cosine similarity.]{
		    \begin{minipage}[b]{0.32\linewidth}
		        \centering
			    \includegraphics[width = 0.96\textwidth]{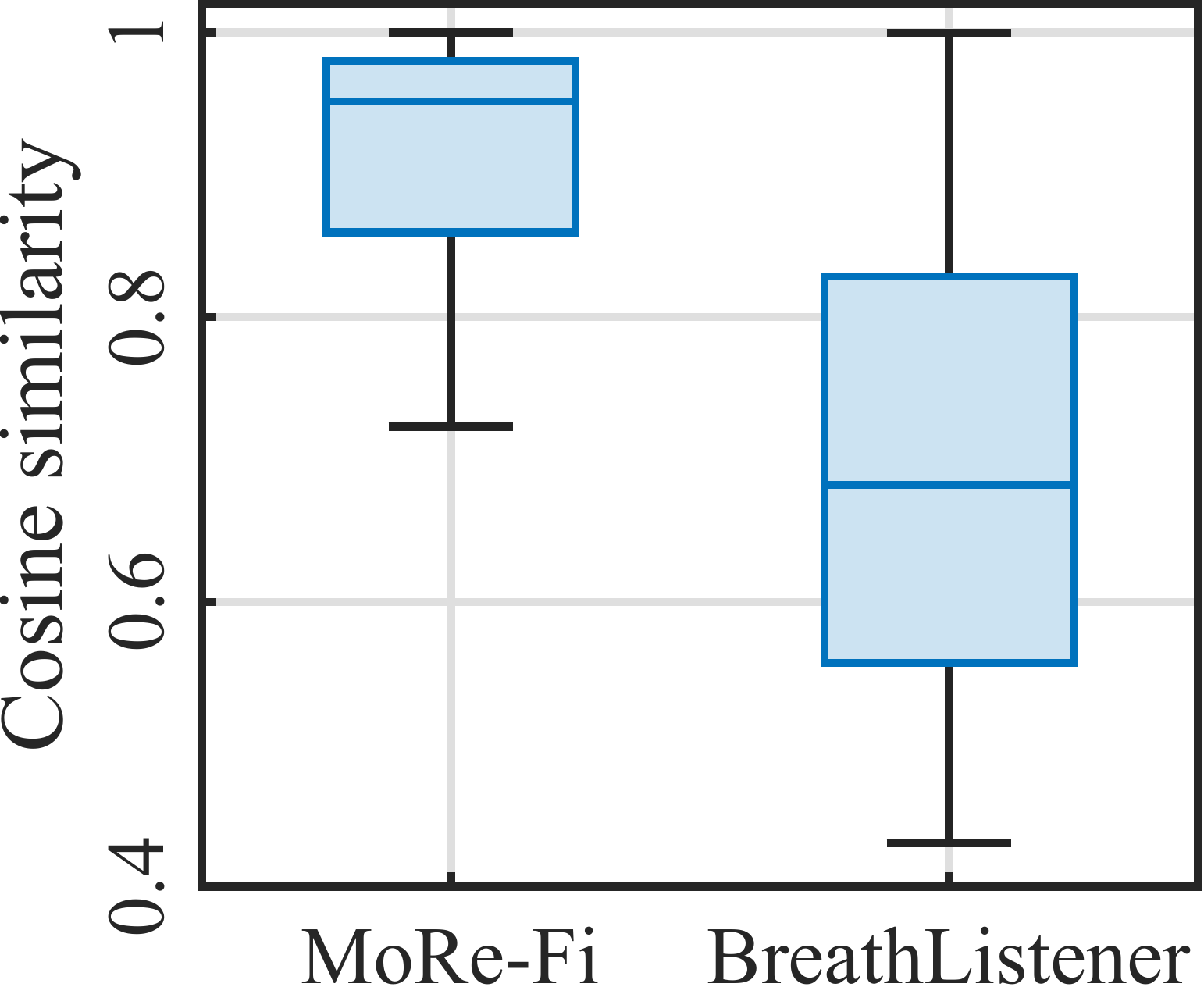}
			    \label{subfig:bs_cosine}
			\end{minipage}
		}
		\subfloat[Respiratory rate.]{
		  \begin{minipage}[b]{0.32\linewidth}
		        \centering
			    \includegraphics[width = 0.96\textwidth]{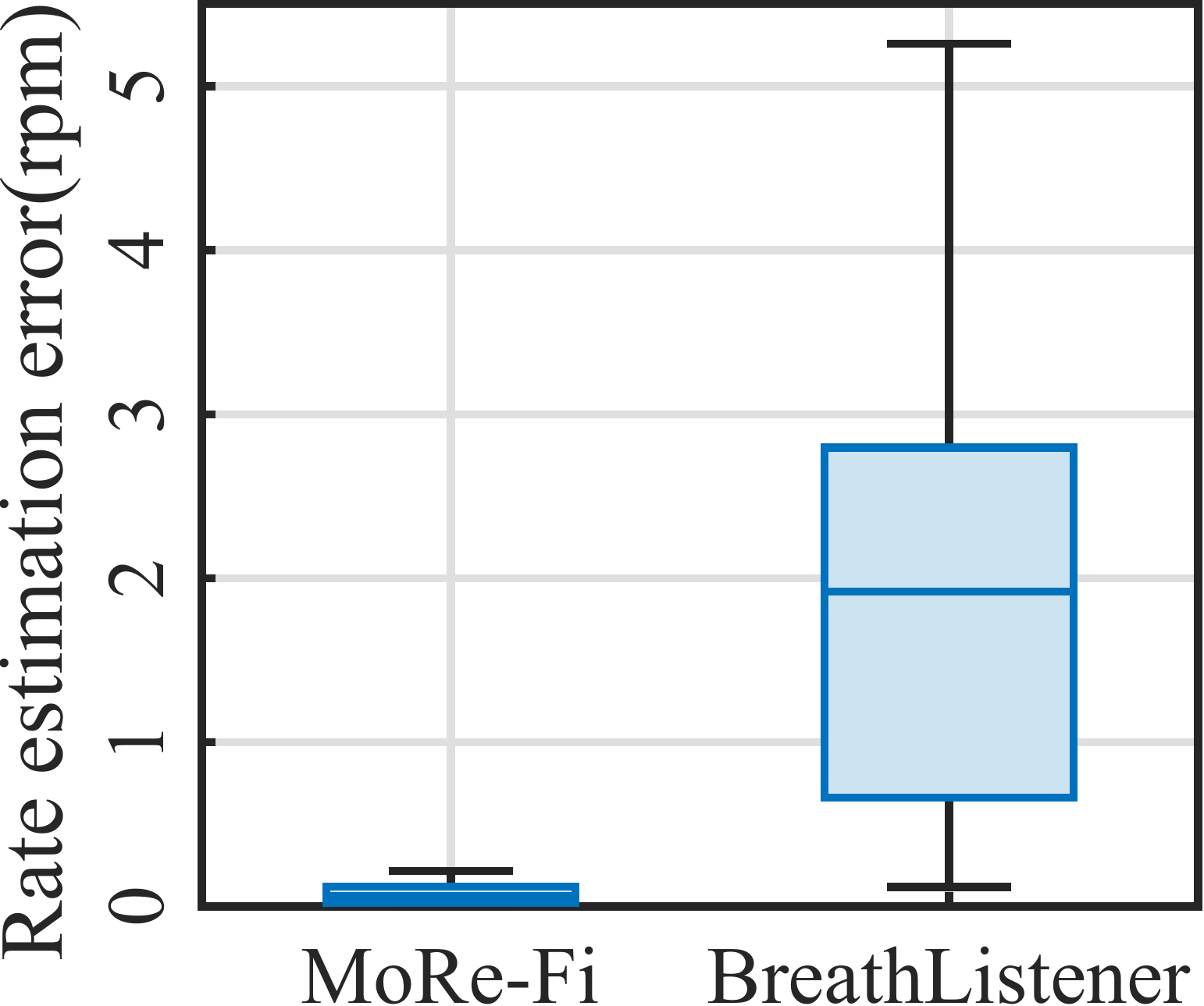}
			    \label{subfig:bs_rate}
			\end{minipage}
		}
		\subfloat[Peak and valley time.]{
		  \begin{minipage}[b]{0.32\linewidth}
		        \centering
			    \includegraphics[width = 0.96\textwidth]{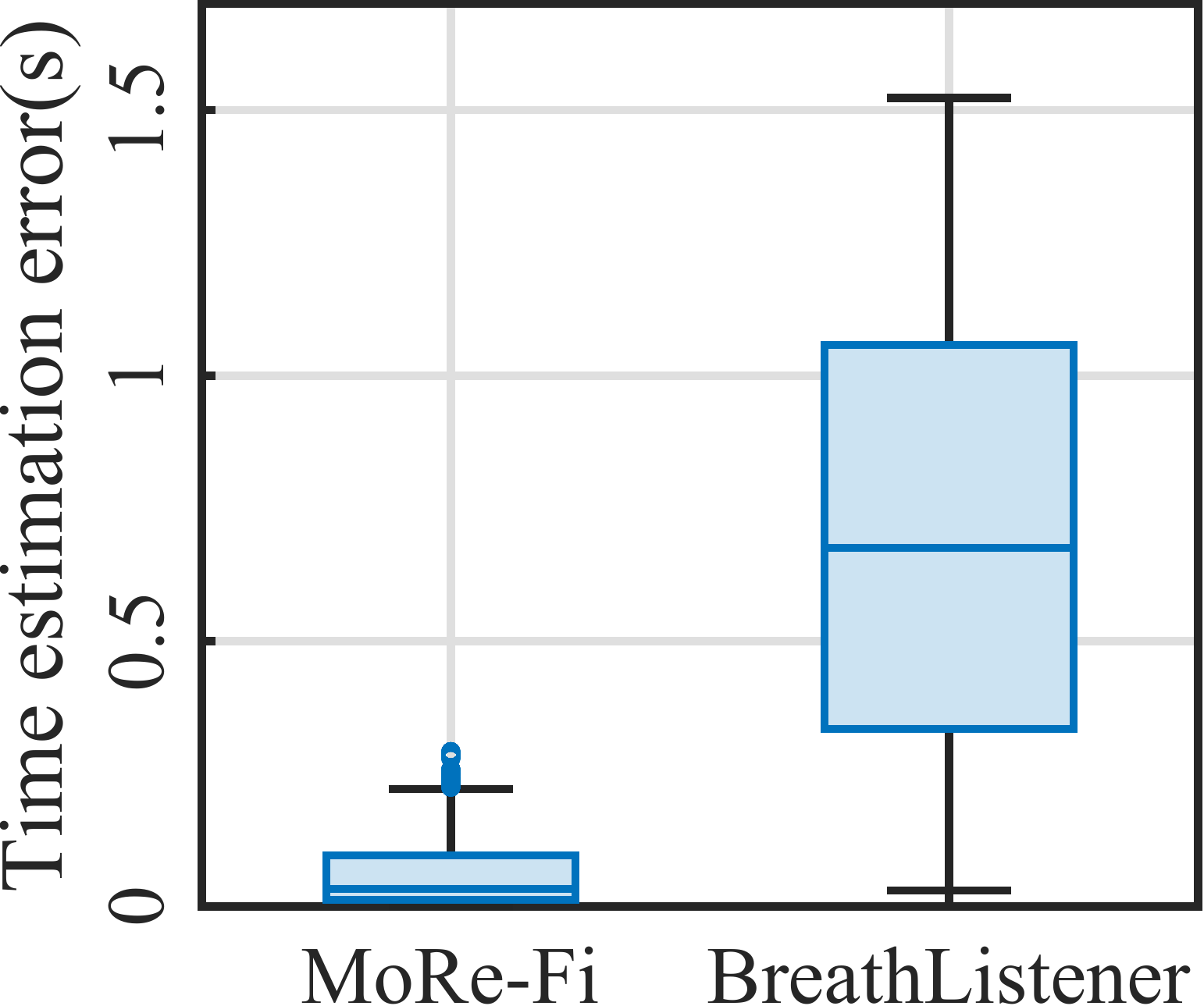}
			    \label{subfig:bs_time}
			\end{minipage}
		}
		\caption{Comparison with baseline method.}
		\label{fig:baseline}
\end{figure}

We compare \name\ with BreathListener~\cite{xu2019breathlisener} in terms of recovered waveform quality in Figure~\ref{fig:baseline}. Two examples are shown in Figure~\ref{subfig:bs_waveform_to} and~\ref{subfig:bs_waveform_wt} to directly contrast the waveforms, then comparisons in terms of three metrics are provided in the remaining subfigures. In general, \name\ recovers respiratory waveform accurately, whereas BreathListener tends to generate distorted (sometimes even erroneous) waveform when the body movements become more intensive. 
In Figures~\ref{subfig:bs_cosine},~\ref{subfig:bs_rate}, and~\ref{subfig:bs_time}, \name\ exhibits a much better performance in cosine similarity, respiratory rate estimation, and peak/valley time estimation than the baseline, all thanks to its motion-robust design. 

The inferior performance of BreathListener can be attributed to the mismatch between EEMD and GAN adopted by it. As discussed in Section~\ref{ssec:conventional}, the EEMD algorithm is incapable of handling complex I/Q signals, so one has to first project the I/Q signals to a 1-D sequence in an information-lossy manner. A consequence of this drawback is that motion interference cannot be correctly separated, as illustrated in Figure~\ref{fig:conventional}. Given the potentially erroneous decomposition of EEMD, GAN that already suffers from instability during training~\cite{shmelkov2018good} becomes even harder to converge. 
For those converged cases, the EEMD decomposed waveform is already close to ground truth, though possibly with wrong features (e.g., phase) that GAN barely helps to correct. Consequently, the biomarkers inferred from the BreathListener recovered waveform can have very large errors, as shown in In Figures~\ref{subfig:bs_rate} and~\ref{subfig:bs_time}.
On the contrary, our IQ-VED is trained and operates in an integrated manner: it uses the encoder to decompose signal and the decoder to reconstruct respiratory waveform. As a result, \name\ is far more effective than the baseline, as demonstrated by these comparisons.

\subsubsection{Estimation Errors of Indicators} \label{sssec:eein}
\begin{figure}[h]
\vspace{-1ex}
    \setlength\abovecaptionskip{8pt}
    \centering
	\includegraphics[width=.83\linewidth]{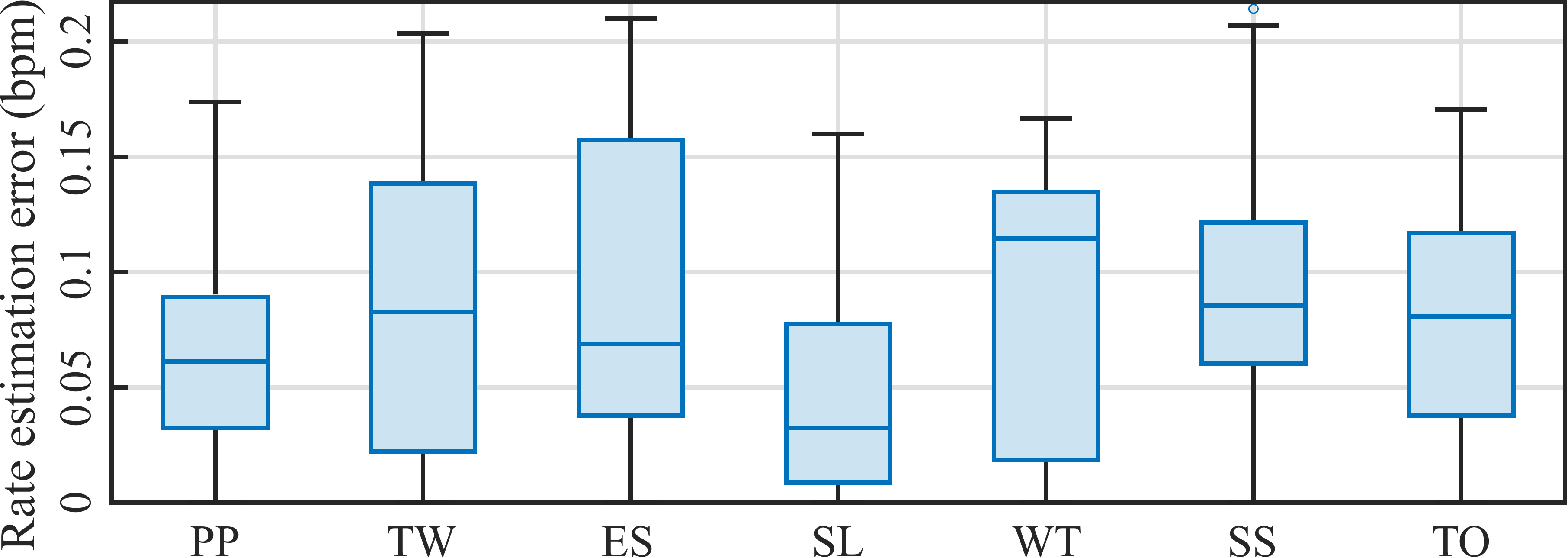}
	\caption{Estimation error of respiratory rate.}
	\label{fig:rate_err}
	\vspace{-1ex}
\end{figure}

Given the overall performance of \name\ reported in the previous sub-sections, we hereby pay special attention to the estimation errors of several indicators. Naturally, the performance of \name\ in estimating instantaneous respiratory rate is first evaluated in Figure~\ref{fig:rate_err}, showing a very consistent accuracy with the majority of errors being under 0.1~\!bpm. Similar to the cosine similarity, both PP and SL have the least impact on rate estimation, while WT and SS cause the worst performance because they entail large body movements.    

\begin{figure}[h]
\vspace{-1ex}
    \setlength\abovecaptionskip{8pt}
    \centering
	\includegraphics[width=.86\linewidth]{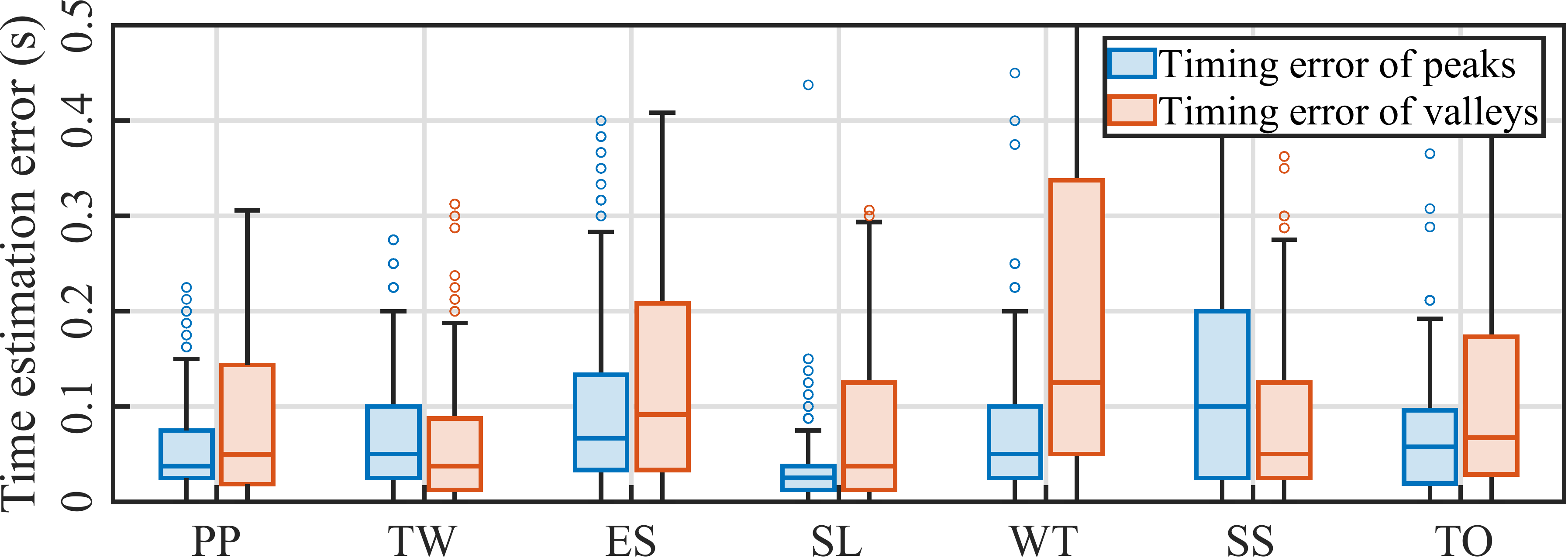}
	\caption{Estimation errors of peak and valley times on respiratory waveform.}
	\label{fig:timing}
	\vspace{-1ex}
\end{figure}

To evaluate the accuracy of time-related biomarkers such as $t_{\mathrm{tc}}$, $t_{\mathrm{i}}$, and $t_{\mathrm{e}}$, we inspect the estimation errors of the peak and valley times on respiratory waveform, and the results are shown in Figure~\ref{fig:timing}. It can be observed that most of the mean errors are below 0.1~\!s, indicating high accuracy of \name's event time estimation. An interesting phenomenon is that the errors of valley time are noticeably larger than those of the peak time; this can be attributed to the fact that the valleys in the waveform, as shown in Figure~\ref{fig:qualitative}, are relatively ``flatter'' than the peaks, thus making it harder for IQ-VED to capture and recover the exact times of the valleys.

\begin{figure}[t]
    \setlength\abovecaptionskip{8pt}
    \centering
	\includegraphics[width=.83\linewidth]{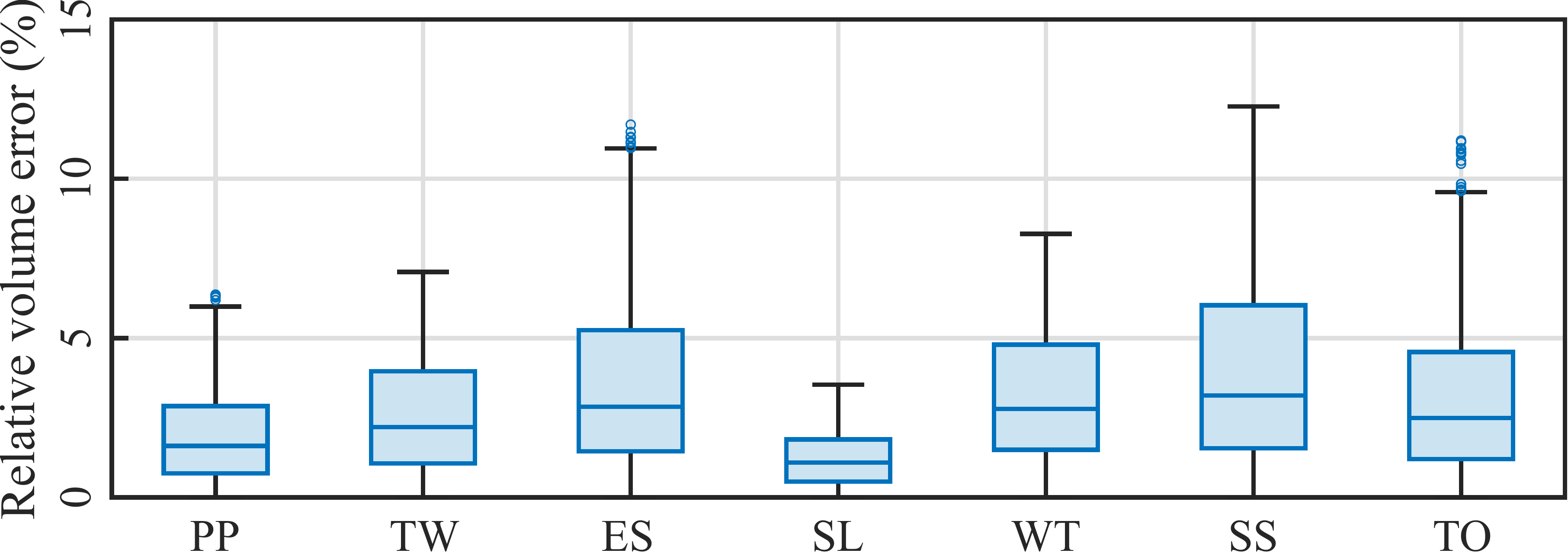}
	\caption{Estimation error of tidal volume.}
	\label{fig:volume}
	\vspace{-2ex}
\end{figure}

Finally, the performance of tidal volume estimation is reported in Figure~\ref{fig:volume}, which shows that the mean errors for all seven body movements are below 3\%, and more than 75\% of the errors are below 5\%. Similar to previous sections, we find that PP and SL have the least impact on rate estimation, while WT and SS cause the worst performance. Particularly, it appears that SS has the most adverse effects on the result, which can be explained by the fact that the chest displacement caused by a human subject standing up or sitting down varies the most (comparing with other body movements) along the propagation direction of radar signals. Overall, these promising results suggest that \name\ has the potential to further enable lung volume monitoring, as will be discussed in Section~\ref{sec:discussion}.

\subsection{Impact of Practical Factors}
Because biomarkers can all be inferred from respiratory waveform, we focus on evaluating the cosine similarity of the waveforms in this section.

\subsubsection{Human Subjects}
\begin{figure}[h]
    \setlength\abovecaptionskip{8pt}
    \centering
	\includegraphics[width=.86\linewidth]{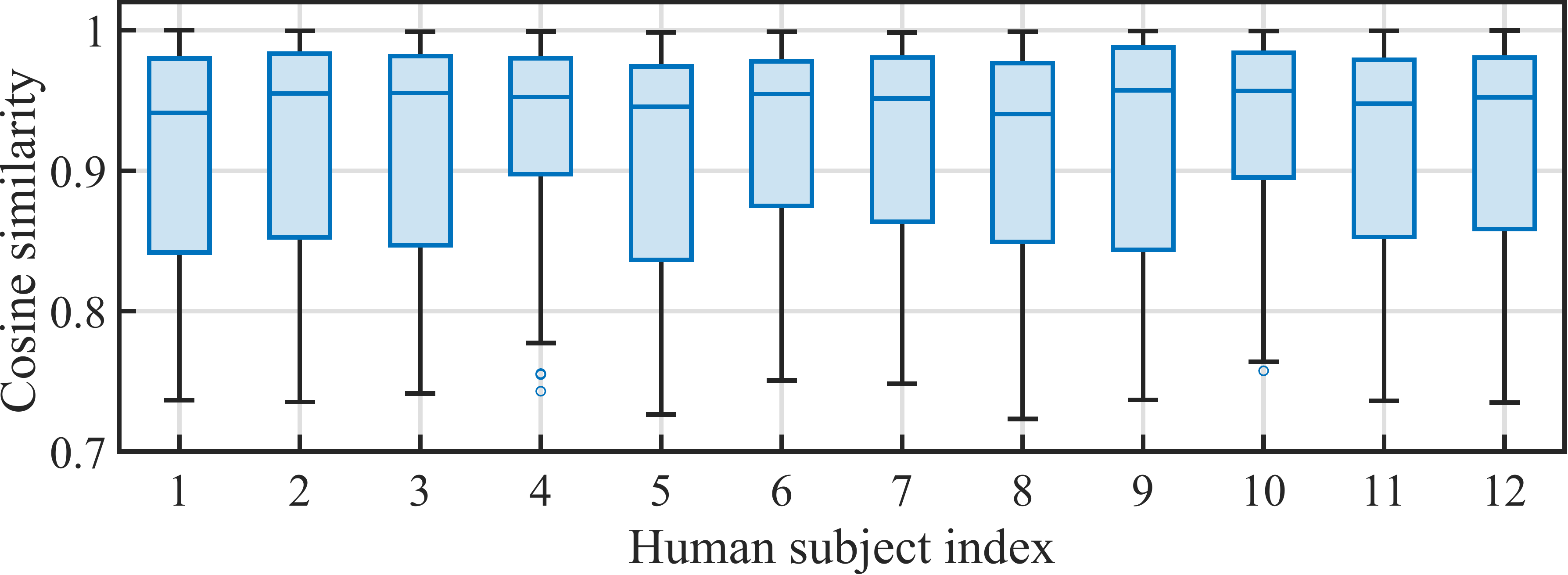}
	\caption{Impact of different human subjects on the cosine similarity.}
	\label{fig:diff_user}
	\vspace{-2ex}
\end{figure}

We show the cosine similarities of \name\ recovered respiratory waveform for all 12 subjects in Figure~\ref{fig:diff_user}. Based on the figure, one may readily conclude that the mean cosine similarities are always greater than 0.95, and more than 75\% of all similarities are above 0.85. These results show that the respiratory waveform recovery of \name\ remains accurate across all involved subjects, largely insensitive to physical discrepancies among them. 

\subsubsection{Training Set Size}
\begin{figure}[b]
    \setlength\abovecaptionskip{6pt}
    \vspace{-1ex}
	\centering
    {
		\begin{minipage}[t]{0.48\linewidth}
			\centering
			\includegraphics[width = .96\textwidth]{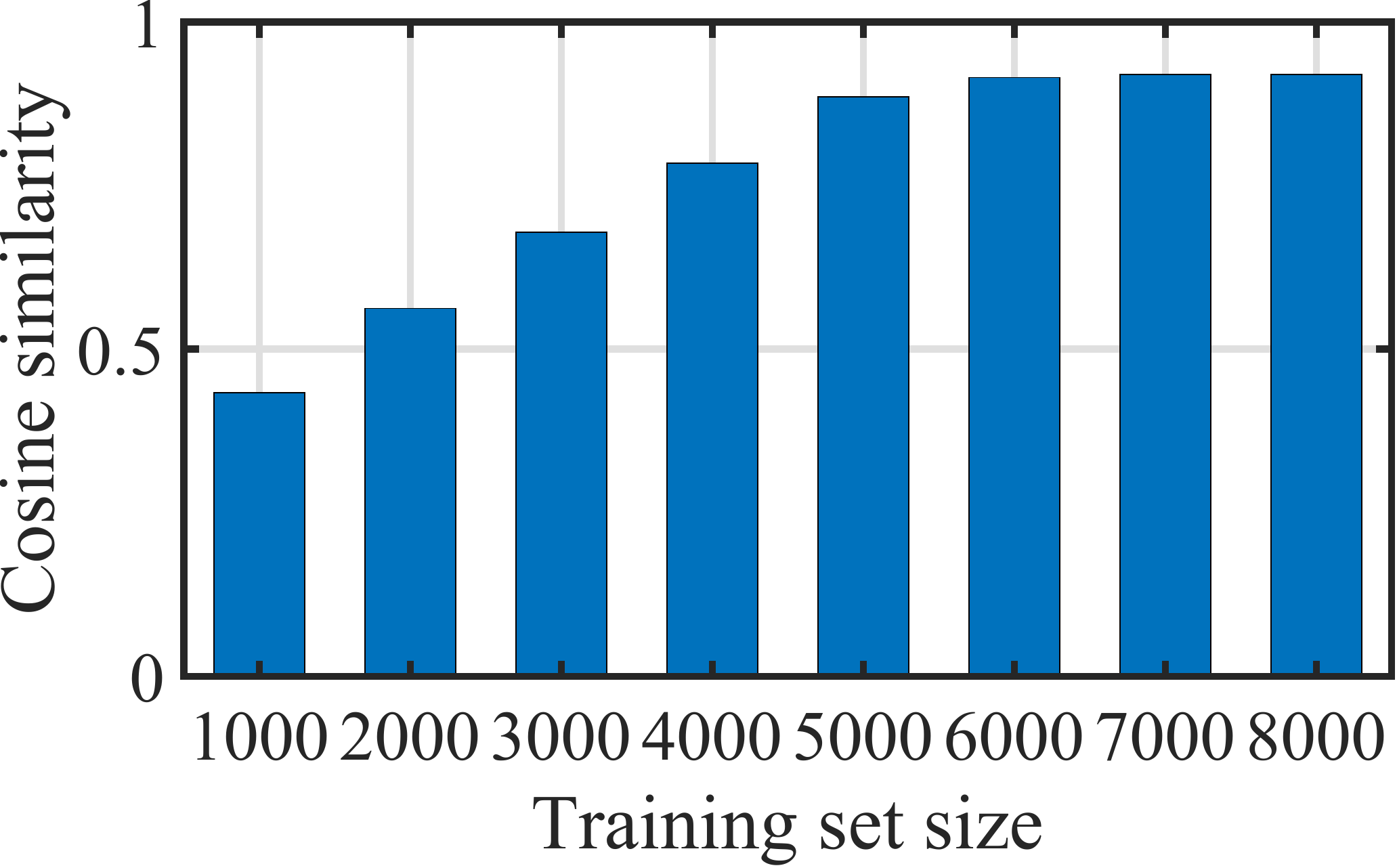}
				\caption{Impact of training set size.}
					\label{subfig:train_size}
		\end{minipage}
	}
	\hfill
	{
		\begin{minipage}[t]{0.48\linewidth}
			\centering
			\includegraphics[width = 0.98\textwidth]{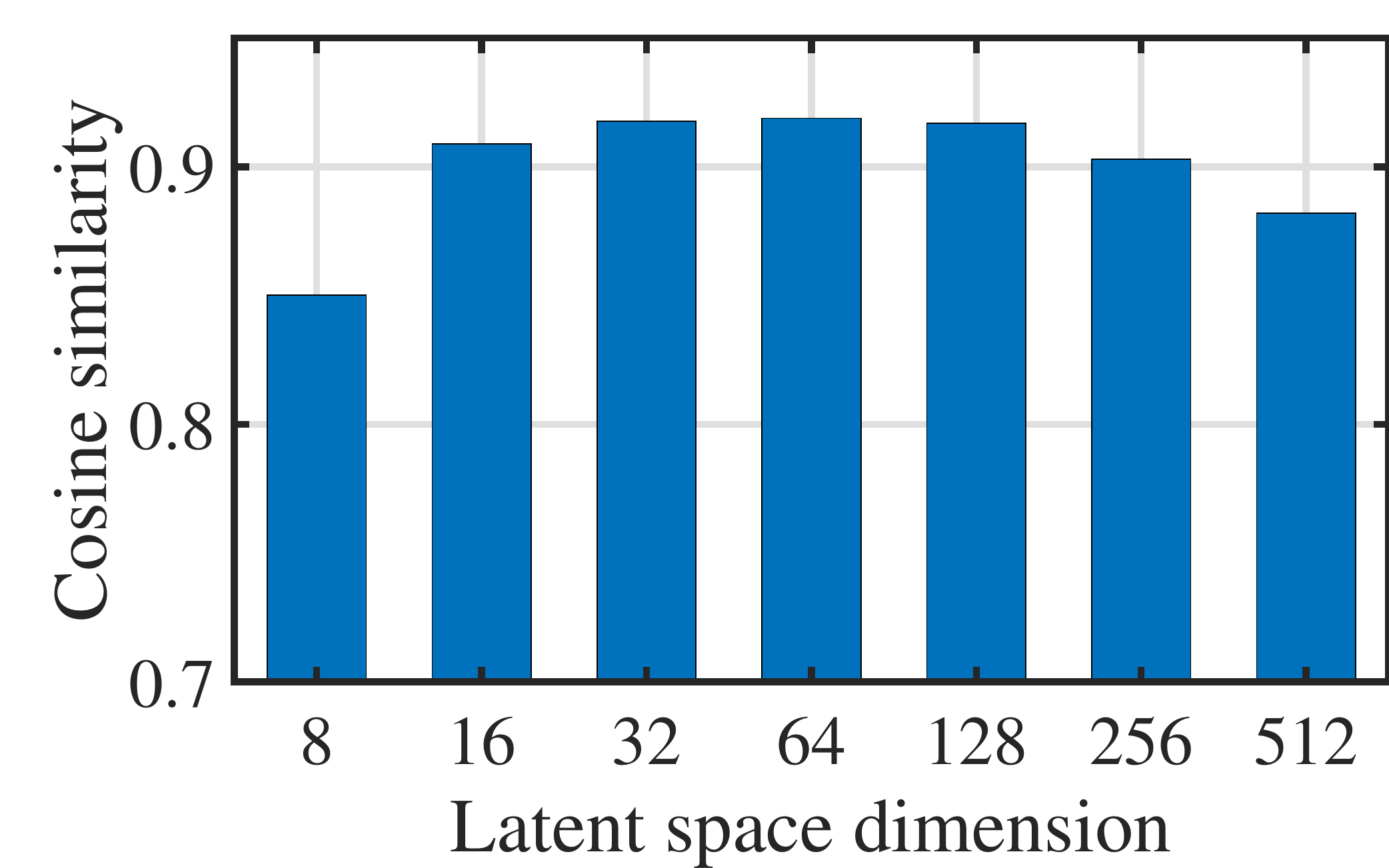}
			\caption{Impact of latent space dimension.}
				\label{subfig:latent_dim}
		\end{minipage}
	}
\end{figure}

As stated in Section~\ref{sec:implementation}, we collect 8,000 data samples of subjects performing different activities for training the IQ-VED network of \name. Figure~\ref{subfig:train_size} shows the impact of training set size on the cosine similarity between the recovered and ground truth waveforms. One may observe that, as the training set size increases, the cosine similarity first increases and then comes to saturation. Specifically, \name\ achieves a cosine similarity greater than 0.9 with 6,000 training samples, which corresponds to 
33 hours of activity data. Because more training data improve the waveform recovery performance only by a negligible margin, our selection of 8,000 training samples is sufficient.

\subsubsection{Latent Space Dimension}
Another property of IQ-VED that affects the recovery performance is the number of latent space dimensions. On one hand, a small latent space dimension may limit the capacity of the latent representation and potentially prevent the loss function from converging to a sufficiently small value. On the other hand, as most practical signals are sparse, overly increasing the dimension of the latent space can be unnecessary while causing slow convergence in training. Consequently, a competent system should strike a balance between expressiveness and compactness of the latent space. According to Figure~\ref{subfig:latent_dim} that shows the impact of latent space dimension on the cosine similarity between the recovered and ground truth waveforms, the performance first improves with the dimension thanks to a better expressiveness but degrades after the dimension reaching 64 due to the increased hardness in training. Therefore, 64 is chosen as the dimension of the latent space for IQ-VED.

\subsubsection{Weights of the Loss Function} \label{sssec:impact_weight}
\begin{figure}[ht]
    \setlength\abovecaptionskip{6pt}
    \vspace{-3ex}
	   \captionsetup[subfigure]{justification=centering}
		\centering
		\subfloat[$\gamma$ (when $\eta=2\mathrm{e}\!-\!4$).]{
		    \begin{minipage}[b]{0.48\linewidth}
		        \centering
			    \includegraphics[width = 0.96\textwidth]{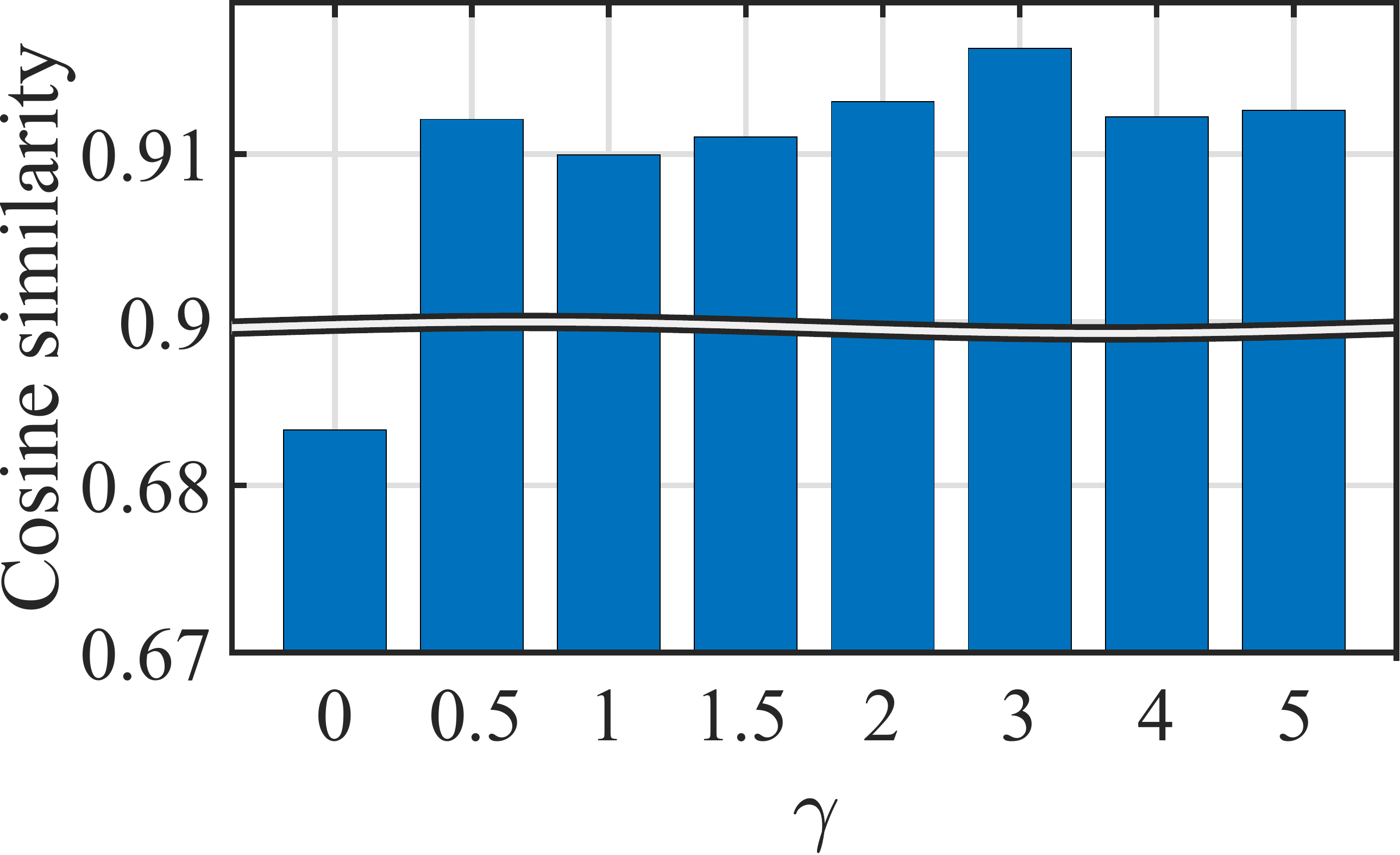}
			    \label{subfig:eta}
			\end{minipage}
		}
		\subfloat[$\eta$ (when $\gamma = 3$).]{
		  \begin{minipage}[b]{0.48\linewidth}
		        \centering
			    \includegraphics[width = 0.96\textwidth]{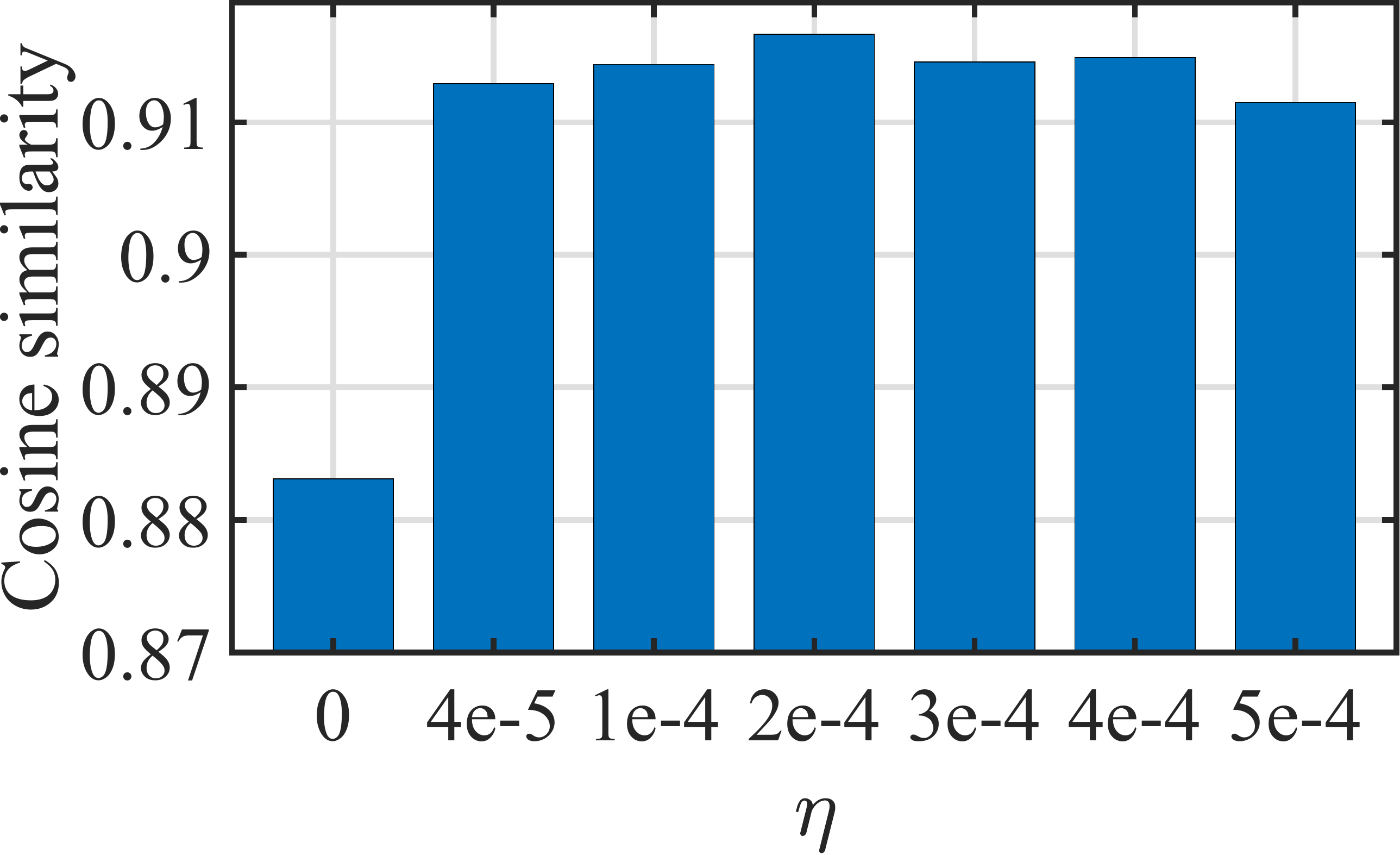}
			    \label{subfig:gamma}
			\end{minipage}
		}
		\caption{\rev{Impact of different weights of the loss function.}}
		\label{fig:weight}
	    \vspace{-2ex}
\end{figure}

The weights in Equation~\eqref{eq:loss} are crucial parameters to be tuned for IQ-VED. In theory, a larger $\gamma$ encourages continuity and disentanglement of the latent space, potentially improving the generalization capability of IQ-VED. A larger $\eta$ improves the alignment of the I/Q representations but may restrict their expressiveness of the underlying I/Q signal. To determine the optimal weights, Figure~\ref{fig:weight} shows the cosine similarity between the recovered and ground truth waveforms as the functions of individual weights; one can clearly observe that
$\gamma=3$ and $\eta=2\mathrm{e}\!-\!4$ allow IQ-VED to achieve the best performance in waveform recovery. 
\subsubsection{Subject Clothing}
\begin{figure}[b]
    \setlength\abovecaptionskip{6pt}
    \vspace{-2ex}
	\centering
    {
		\begin{minipage}[t]{0.48\linewidth}
			\centering
			\includegraphics[width = .98\textwidth]{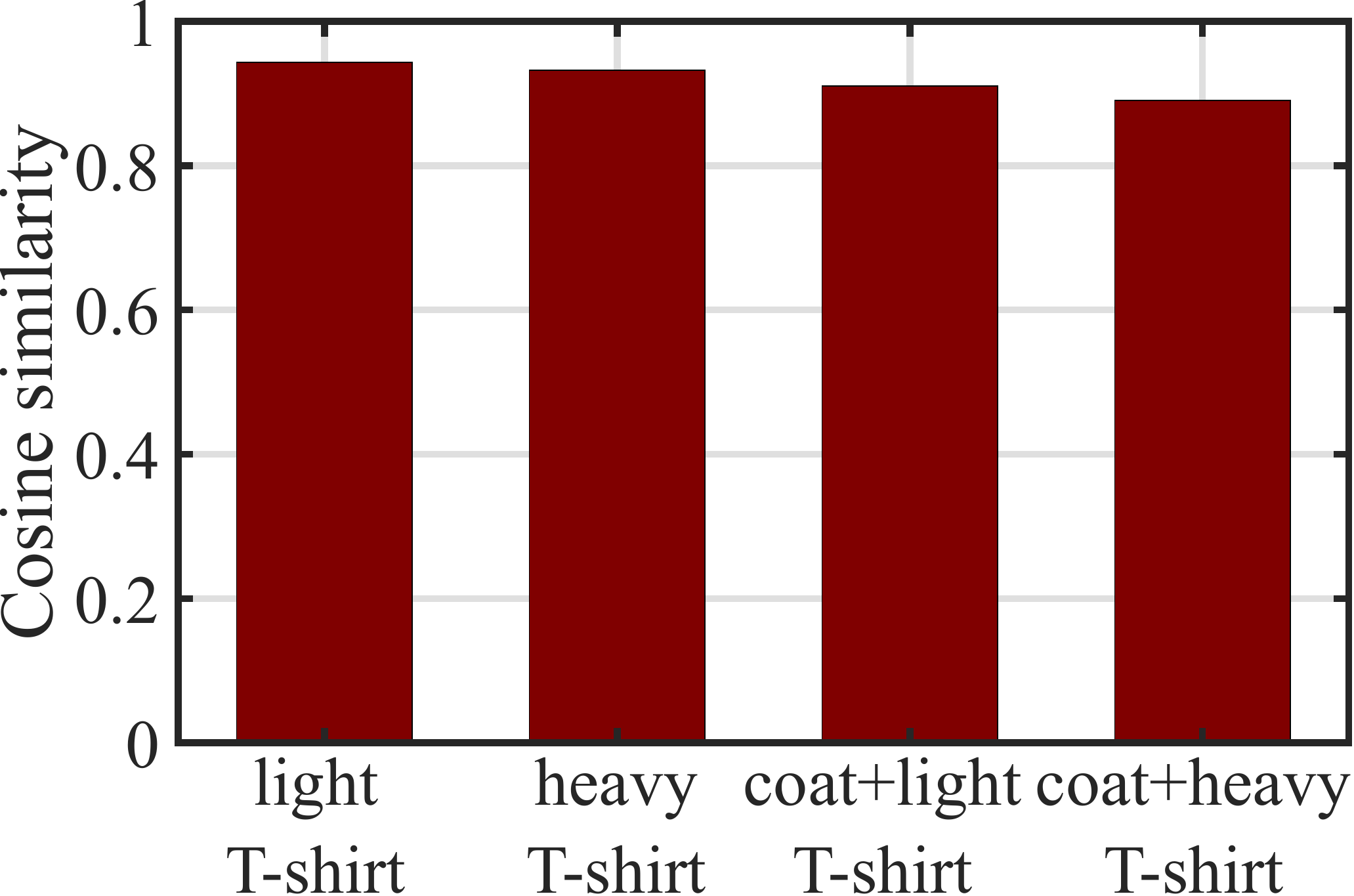}
			\caption{Impact of subject clothing.}
			\label{subfig:clothing}
		\end{minipage}
	}
	\hfill
	{
		\begin{minipage}[t]{0.48\linewidth}
			\centering
			\includegraphics[width = 0.95\textwidth]{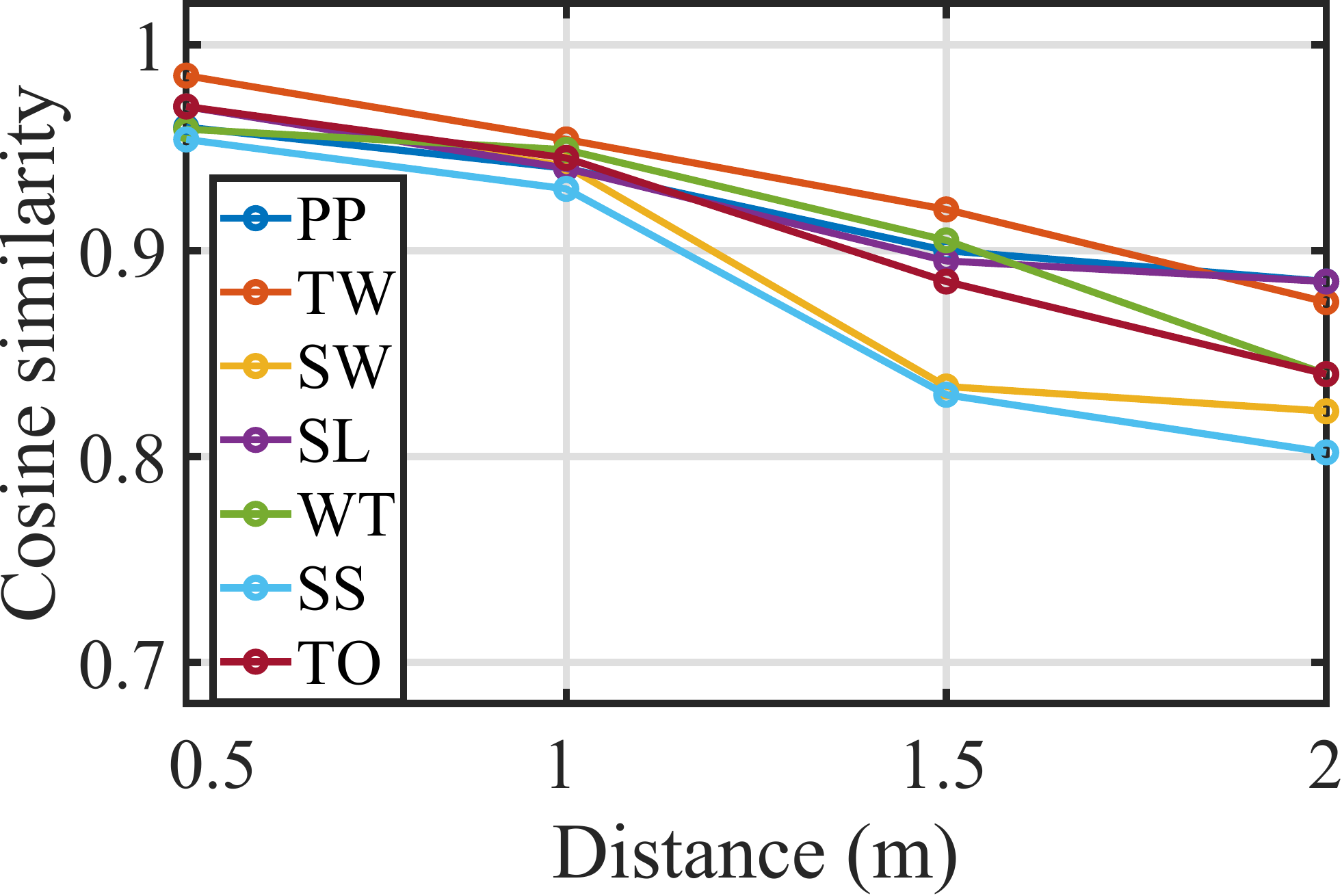}
			\caption{Impact of sensing distance.}
			\label{subfig:distance}
		\end{minipage}
	}
\end{figure}

In the experiments, we ask the subjects to wear different clothes and record respective measurements accordingly. The results shown in Figure~\ref{subfig:clothing} involve four types of clothes: i) lightweight T-shirt, ii) heavyweight T-shirt, iii) coat+lightweight T-shirt, and iv) coat + heavyweight T-shirt. According to Figure~\ref{subfig:clothing}, \name\ reaches an overall average cosine similarity 
above 0.9 across all types of clothes. Intuitively, \name\ does performs slightly better when a subject wears less, because heavier clothes attenuate more severely the reflected signals from the subject's chest. Fortunately, even in the worst case (with coat and heavy T-shirt), the average cosine similarity remains well above 0.8. 

\subsubsection{Sensing Distance}
Sensing distance is a major limiting factor of RF respiration monitoring system. We ask the human subjects to be away from the radar at 0.5~\!m, 1~\!m, 1.5~\!m, and 2~\!m to study the impact of sensing distance. Not surprisingly, the cosine similarities reported in Figure~\ref{subfig:distance} clearly demonstrate a negative effect of sensing distance on performance. %
Nonetheless, the average cosine similarity remains above 0.8 even under the most intensive body movements, firmly proving the effectiveness of \name. 

\section{Potential Medical Adoptions}\label{sec:discussion}
\name\ is expected to not only continuously extract respiratory waveform, but also enable earlier intervention for people with potential pulmonary disease. 
Unfortunately, we cannot fully evaluate the second one due to the lack of subjects with related diseases and of the support from medical professionals.
Therefore, instead of evaluating the performance in disease diagnosis, we hereby give a brief discussion on potential medical adoptions of \name.

The biomarkers evaluated in Sections~\ref{sssec:waveinf} and~\ref{sssec:eein} are directly obtainable from the timestamps and amplitude of respiratory waveform; they certainly reflect changes in respiratory patterns and hence can serve as indicators to a series of health conditions, including apnea (cessation of breathing)~\cite{white2006sleep}, tachypnea (abnormally rapid breathing)~\cite{avery1966transient}, hyperpnea (abnormally slow breathing)~\cite{aaron1992oxygen}, dyspnea~\cite{manning1995pathophysiology} (shortness of breath), Cheyne-Stokes respiration (progressively deeper breathing followed by a gradual decrease that results in an apnea)~\cite{lanfranchi1999prognostic}, and Biot respiration (regular deep inhalations followed by periods of apnea)~\cite{farney2008adaptive}.
\begin{figure}[b]
    \setlength\abovecaptionskip{8pt}
    \vspace{-3ex}
	   \captionsetup[subfigure]{justification=centering}
		\centering
		\subfloat[Respiration volume and flow.]{
		    \begin{minipage}[b]{0.47\linewidth}
		        \centering
			    \includegraphics[width = 0.96\textwidth]{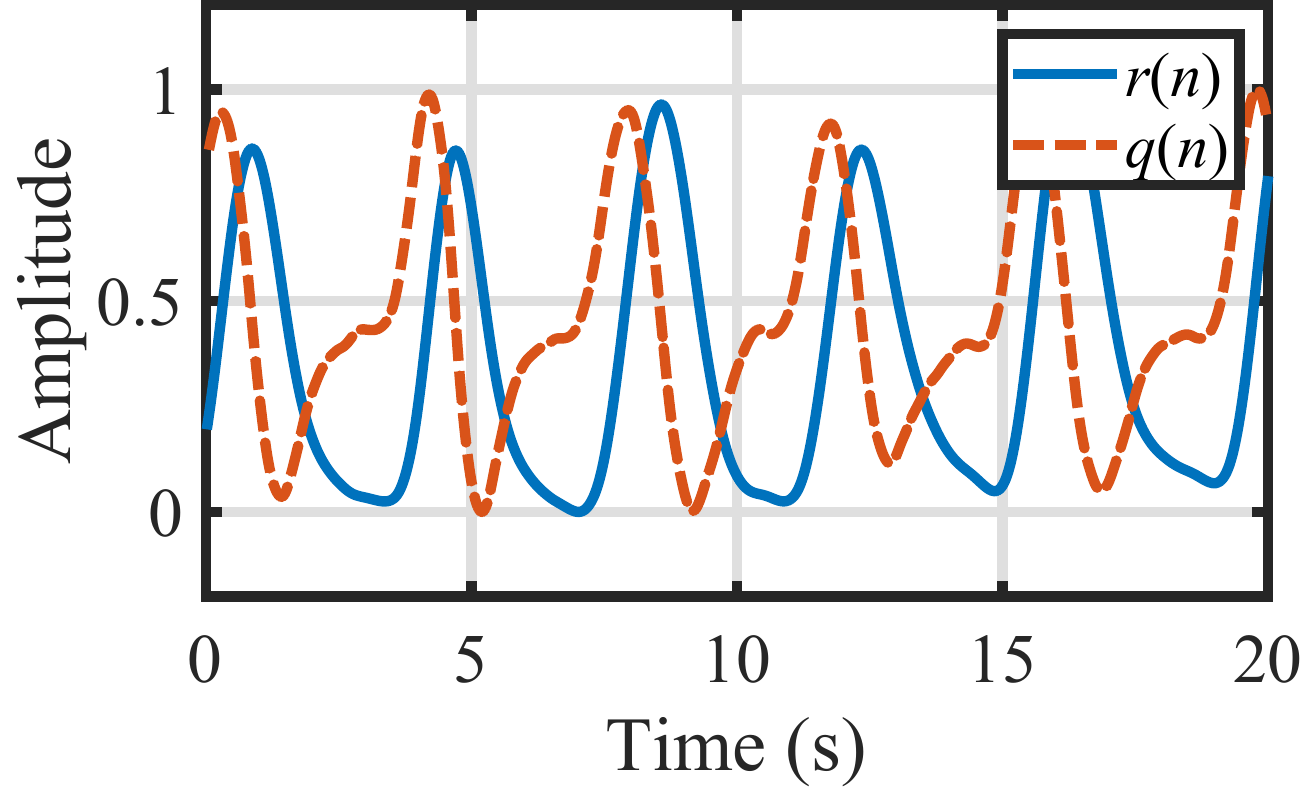}
			    \label{subfig:rq}
			\end{minipage}
		}
		\subfloat[Flow-volume loop.]{
		  \begin{minipage}[b]{0.47\linewidth}
		        \centering
			    \includegraphics[width = 0.96\textwidth]{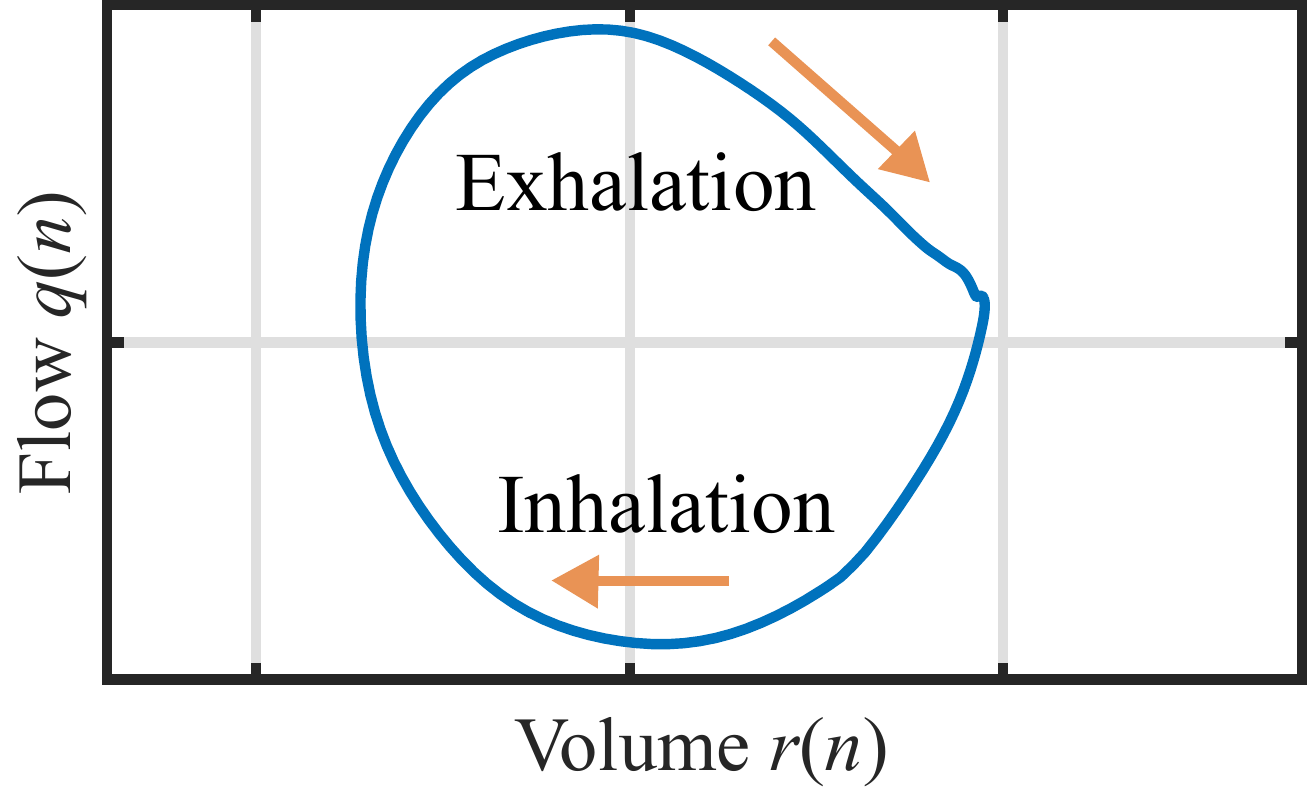}
			    \label{subfig:loop}
			\end{minipage}
		}
		\caption{Visualization of respiration volume and flow.}
		\label{fig:volume_flow}
\end{figure}
In addition to the directly observable biomarkers, respiration flow can also be derived from the fine-grained waveform. Because respiration flow $q(n)$ can be seen as the derivative of respiration volume, and respiration volume is proportional to the amplitude of the recovered waveform $r^{\prime}(n)$, we have $q(n) = c \frac{dr^{\prime}(n)}{dn}$, where $n$ denotes time, and $c$ is a scaling factor. 
We further apply the noise-robust differentiator~\cite{snrd} to get the respiration flow $q(n)$:
\begin{align}
    q(n) &\approx \frac{1}{h} \textstyle{\sum_{k=1}^{M} b_{k}} \cdot\left(r(n+k)-r(n-k)\right),
\end{align}
where $b_{k}=\frac{1}{2^{2 m+1}}\left[\left(\begin{array}{c}
    2 m \\
    m-k+1
    \end{array}\right)-\left(\begin{array}{c}
    2 m \\
    m-k-1
    \end{array}\right)\right]$,
    $h$ is the time interval between two consecutive points, $m=\frac{N-3}{2}$, $M=\frac{N-1}{2}$, and $N$ is the number of points used to estimate derivative. An example is provide in Figure~\ref{subfig:rq}. By tracing the change of flow and volume together, the flow-volume loop graph is obtained in Figure~\ref{subfig:loop}, which clearly visualizes both inhalation and exhalation processes.

\begin{figure}[t]
    \setlength\abovecaptionskip{8pt}
    \vspace{-1.5ex}
	   \captionsetup[subfigure]{justification=centering}
		\centering
		\subfloat[Restrictive lung disease.]{
		    \begin{minipage}[b]{0.46\linewidth}
		        \centering
			    \includegraphics[width=0.96\textwidth]{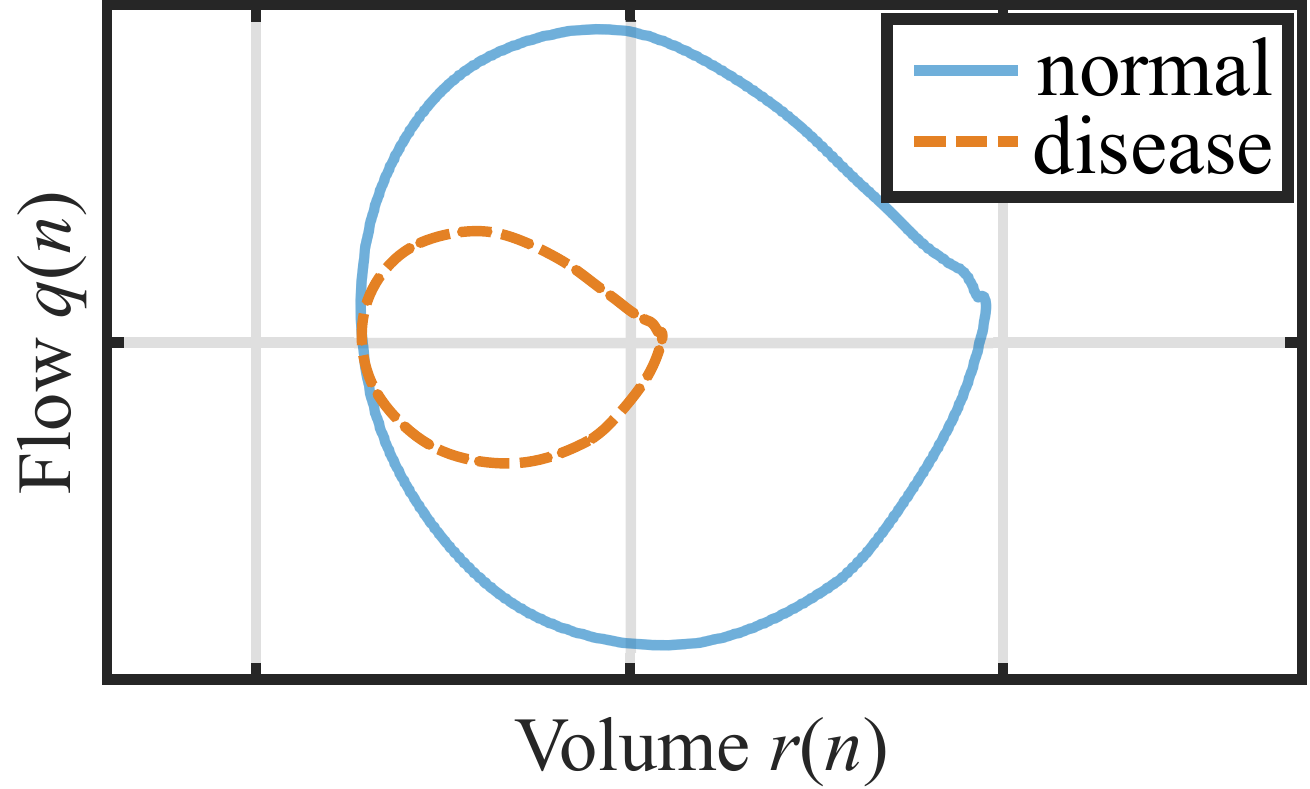}
			    \label{subfig:restrictive}
			\end{minipage}
		}
		\subfloat[Obstructive lung disease.]{
		  \begin{minipage}[b]{0.46\linewidth}
		        \centering
			    \includegraphics[width=0.96\textwidth]{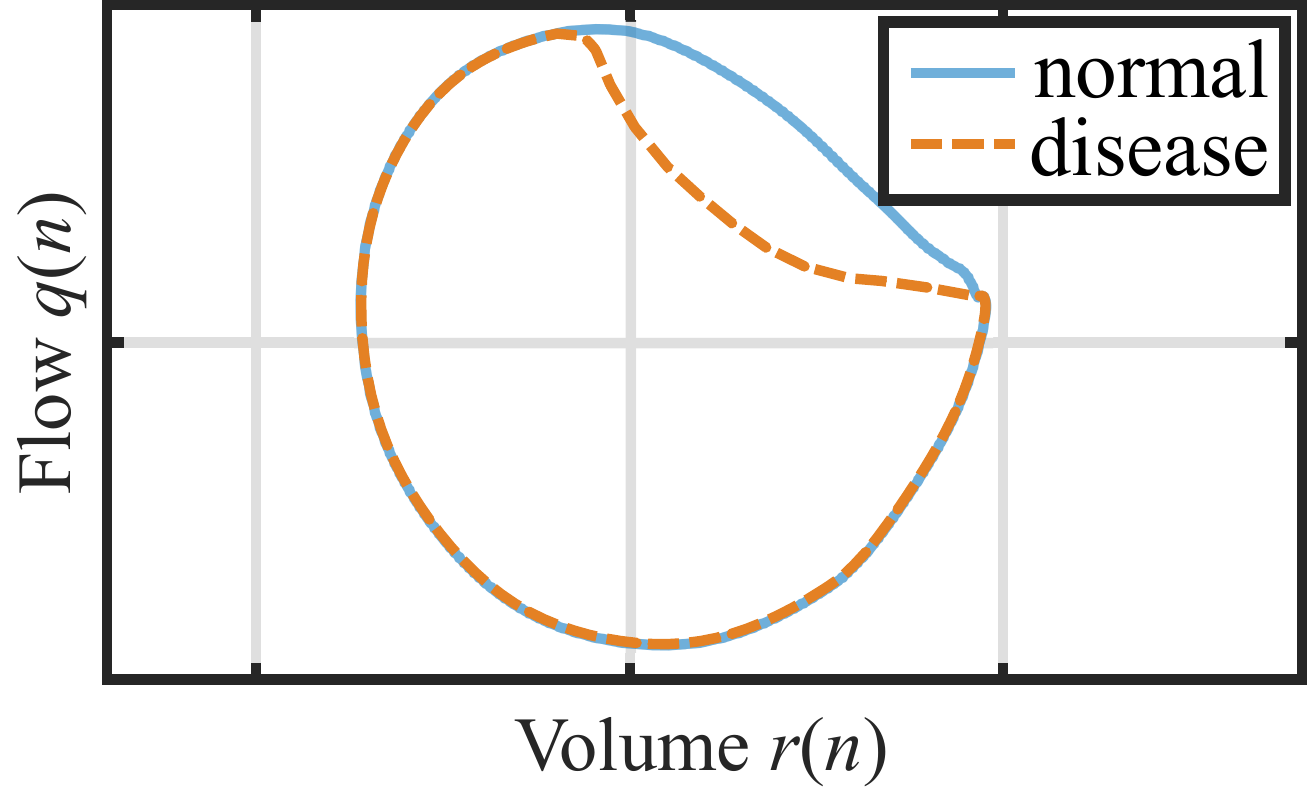}
			    \label{subfig:obstructive}
			\end{minipage}
		}
		\\
		\subfloat[Fixed upper airway.]{
		    \begin{minipage}[b]{0.46\linewidth}
		        \centering
			    \includegraphics[width = 0.96\textwidth]{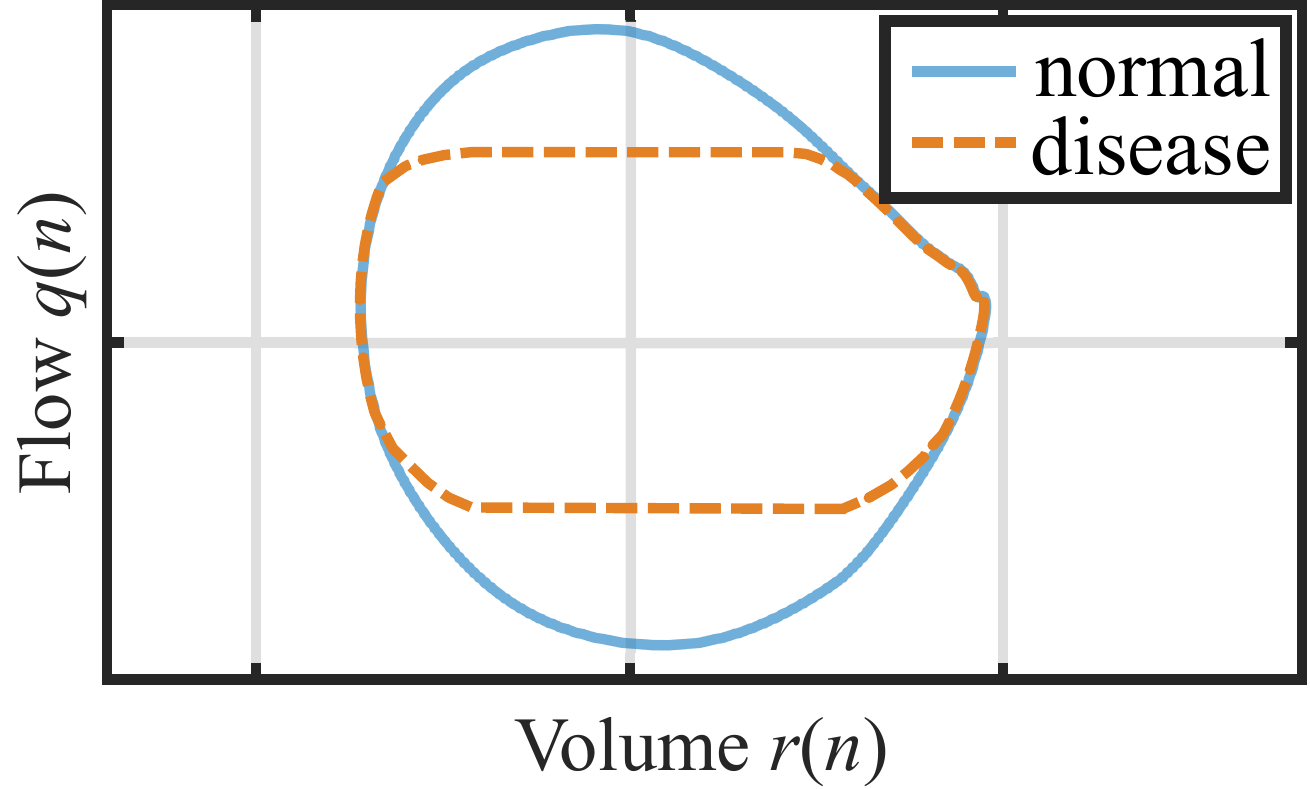}
			    \label{subfig:fixed}
			\end{minipage}
		}
		\subfloat[Variable extrathoracic.]{
		  \begin{minipage}[b]{0.46\linewidth}
		        \centering
			    \includegraphics[width = 0.96\textwidth]{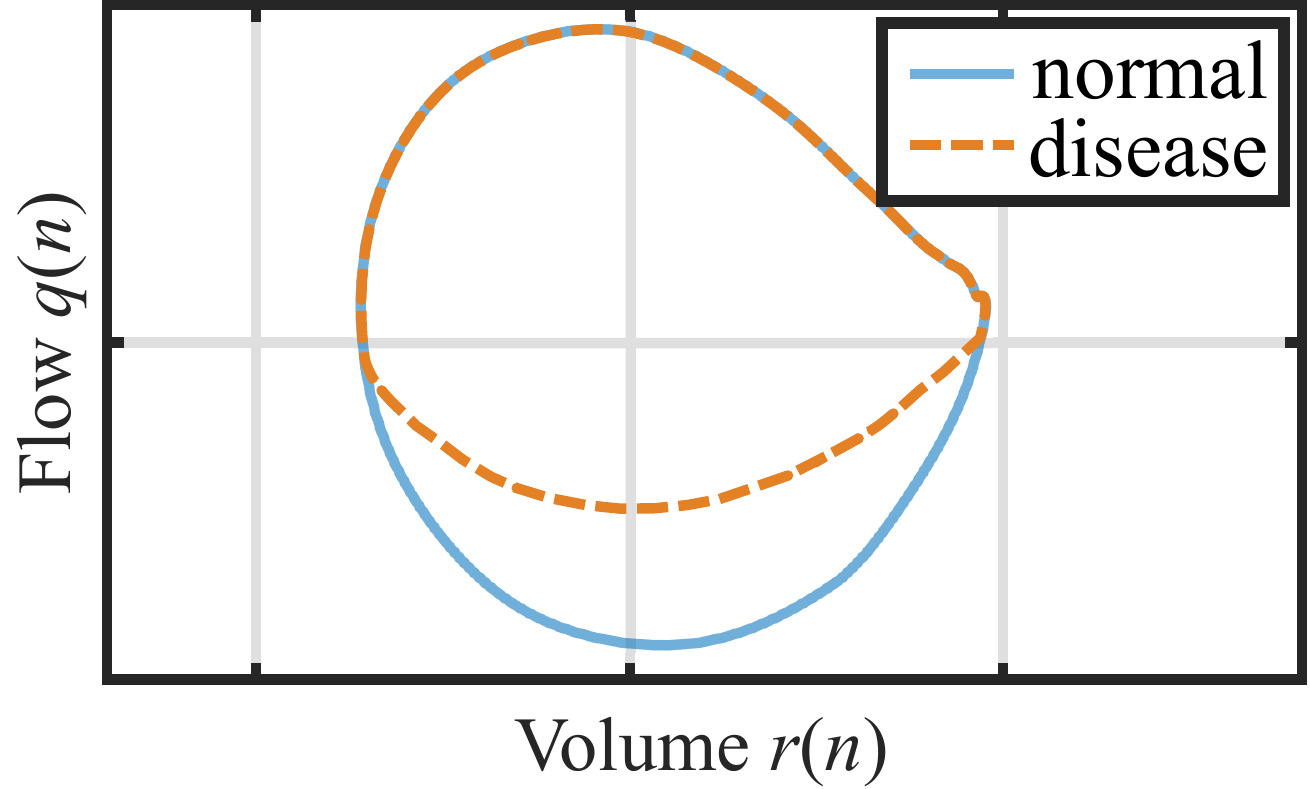}
			    \label{subfig:variable}
			\end{minipage}
		}
		\caption{Flow-volume loop patterns of different pulmonary diseases related to abnormal airflow.}
		\label{fig:obstructive_disease}
	    \vspace{-3ex}
\end{figure}

The shape of the flow-volume loop can provide diagnostic information for many chronic pulmonary diseases related to abnormal airflow, as illustrated in Figure~\ref{fig:obstructive_disease}. Figures~\ref{subfig:restrictive} to~\ref{subfig:variable} respectively demonstrate i) restrictive lung disease as a result of a decreased lung volume~\cite{naji2006effectiveness}, %
ii) obstructive lung disease caused by obstruction to airflow when exhaling~\cite{flenley1985sleep}, %
iii) fixed upper airway obstruction occurring when infections spread along the planes formed by the deep cervical fascia~\cite{miller1985stenosis},
and iv) variable extrathoracic obstruction resulting from tumors of the lower trachea or main bronchus~\cite{haponik1981abnormal}.

The capabilities of continuous, motion-robust monitoring potential lung diseases make \name\ useful in many scenarios. To take a few examples, \name\ can be deployed in hospitals, nursing facilities, as well at homes. In hospitals, the long-term respiration data provided by \name\ allow effective triage and ongoing care management. In nursing facilities and at home, health professionals and normal users will become more aware of the respiratory status of the care recipients, and any sudden changes in the recovered respiratory waveform can be addressed immediately to prevent further exacerbation. Moreover, in the recent COVID-19 pandemic, \name\ can be deployed to enable earlier intervention, and prevent the spread of the deadly infectious disease.

\section{Related Work}\label{sec:related}
Existing works on respiration monitoring can be roughly categorized into wearable sensor-based methods ~\cite{neulog, guder2016paper, min2014simplified, chu2019respiration, fang2016bodyscan} and contact-free methods~\cite{kaltiokallio2014non,xu2019breathlisener,song2020spirosonic, wang2019contactless, min2010noncontact,ravichandran2015wibreathe,adib2015smart,yang2017vital, nguyen2016continuous}. 
The contact nature of wearable sensors causes time-consuming adoptions, degrades user experience, and may even change users' respiration habits ~\cite{gu2015assessment}, so our paper focuses on innovating 
contact-free respiration monitoring. 
Contact-free respiration monitoring exploits either RF-sensing~\cite{kaltiokallio2014non,ravichandran2015wibreathe,adib2015smart, nguyen2016continuous,yang2017vital,zeng2019farsense,zheng2020v2ifi} or acoustic sensing~\cite{wang2018c,xu2019breathlisener,song2020spirosonic,wang2019contactless, min2010noncontact}.\footnote{\rev{We deliberately omit the light sensing methods (e.g.,~\cite{van2016robust}) as they rely on a camera to perform motion tracking, which is both complicated and ineffective: it certainly cannot handle large-scale body movements faced by \name.}} 
Being a typical RF-sensing system, \name\
aims to achieve both \textit{motion-robust} and \textit{fine-grained} respiratory waveform recovery, hence our brief literature survey emphasizes these two aspects when discussing both sensing media.

RF-based respiration monitoring started with estimating respiratory rate and recovering coarse-grained waveform of human subjects in static conditions~\cite{kaltiokallio2014non, adib2015smart, ravichandran2015wibreathe, yang2017vital, zhang2017wicare, zeng2019farsense, yue2018extract}, readily achievable by basic spectrum analysis and filtering. Because these methods do not take motion-robustness into account, they simply suspend respiration monitoring upon encountering sudden motion interference. To achieve motion robustness, early proposals~\cite{changzhili2008Random,cli2016random} rely on tricky placements of multiple radars, so as to cancel out motion interference at the cost of cumbersome synchronization. 
Later researchers have also tried to apply linear filtering for mitigating the effects of motion interference~\cite{Tu20161D, lv2018large}, but their assumptions of unrealistic 1-D body movements~\cite{Tu20161D} or the existence of quasi-static periods during movements~\cite{lv2018large} are too strong to be realistic.
The latest proposal V\textsuperscript{2}iFi~\cite{zheng2020v2ifi} employs an adapted VMD algorithm for removing motion interference from turning steering wheel and the running vehicle, but it achieves coarse-grained respiration monitoring by leveraging only the amplitude of RF signal.

Besides RF-sensing, acoustic sensing is an alternative candidate for respiration monitoring due to its readily deployable consumer-grade hardware. C-FMCW~\cite{wang2018c} and BreathJunior~\cite{wang2019contactless} adopt either continuous-wave or white-noise for respiration monitoring. 
They both target respiratory rate by assuming a subject to remain static. 
BreathListener~\cite{xu2019breathlisener} targets respiratory waveform recovery in driving environments by tackling only small-scale body movements. As explained in Section~\ref{ssec:comp_bs}, the two-stage algorithm adopted there is both complex and incompatible between its two stages.
SpiroSonic~\cite{song2020spirosonic} monitors human lung function by turning the speaker and microphone of a smartphone into a spirometer;
it leverages deep learning to extract several vital indicators such as peak expiratory flow, forced expiratory volume, and forced vital capacity. Though SpiroSonic is not a continuous respiration monitoring system as \name\ does, it does tolerate very small-scale hand drifts when holding a phone, by selectively fitting to relatively ``clean'' data segments free of strong motion interference. In general, acoustic sensing has limited applicability because they are prone to ambient and motion-induced acoustic interference.
\section{Conclusion}\label{sec:conclusion}
Taking an important step toward continuous and ubiquitous health care, we have proposed \name\ in this paper for motion-robust and contact-free respiration monitoring, aiming to recover fine-grained waveform rather than respiratory rate only. Built upon a radar platform~\cite{octopus} and up-to-date deep learning technologies, \name\ expands the scope of contact-free vital sign monitoring by tackling full-scale body movements gracefully. Essentially, though existing signal processing methods unanimously fail to handle the composition between body movements and respiration-induced chest motion, our IQ-VED (the core of \name) succeeds by exerting its non-linear decomposition ability. Via extensive experiments on healthy subjects, we have demonstrated the promising performance of \name\ in fine-grained waveform recovery and long-term respiration monitoring. As this line of work progresses, we are planning to evaluate \name's performance in real-life clinical scenarios, as we believe this work has significant implications to various medical applications including pulmonary disease diagnosis. \rev{Moreover, we are planning to exploit the spatial diversity of offered by large-scale antenna arrays~\cite{iLocScan,siwa} to approach the issue of monitoring vital signs of a walking subject.}

\section*{Acknowledgments}
We are grateful to the anonymous reviewers for their valuable and constructive comments. We would also like to thank WiRUSH~\cite{wirush} for providing fund to develop \name.

\newpage
\balance
\bibliographystyle{ACM-Reference-Format}
\bibliography{main}


\begin{thebibliography}{92}

%%% the Web address.

\ifx \showCODEN    \undefined \def \showCODEN     #1{\unskip}     \fi
\ifx \showDOI      \undefined \def \showDOI       #1{#1}\fi
\ifx \showISBNx    \undefined \def \showISBNx     #1{\unskip}     \fi
\ifx \showISBNxiii \undefined \def \showISBNxiii  #1{\unskip}     \fi
\ifx \showISSN     \undefined \def \showISSN      #1{\unskip}     \fi
\ifx \showLCCN     \undefined \def \showLCCN      #1{\unskip}     \fi
\ifx \shownote     \undefined \def \shownote      #1{#1}          \fi
\ifx \showarticletitle \undefined \def \showarticletitle #1{#1}   \fi
\ifx \showURL      \undefined \def \showURL       {\relax}        \fi
% invisible to TeX
\providecommand\bibfield[2]{#2}
\providecommand\bibinfo[2]{#2}
\providecommand\natexlab[1]{#1}
\providecommand\showeprint[2][]{arXiv:#2}

\bibitem[\protect\citeauthoryear{Aaron, Seow, Johnson, and Dempsey}{Aaron
  et~al\mbox{.}}{1992}]%
        {aaron1992oxygen}
\bibfield{author}{\bibinfo{person}{E.A. Aaron}, \bibinfo{person}{K.C. Seow},
  \bibinfo{person}{B.D. Johnson}, {and} \bibinfo{person}{J.A. Dempsey}.}
  \bibinfo{year}{1992}\natexlab{}.
\newblock \showarticletitle{{Oxygen Cost of Exercise Hyperpnea: Implications
  for Performance}}.
\newblock \bibinfo{journal}{\emph{Journal of Applied Physiology}}
  \bibinfo{volume}{72}, \bibinfo{number}{5} (\bibinfo{year}{1992}),
  \bibinfo{pages}{1818--1825}.
\newblock


\bibitem[\protect\citeauthoryear{Adib, Kabelac, Katabi, and Miller}{Adib
  et~al\mbox{.}}{2014}]%
        {3Dtracking-NSDI}
\bibfield{author}{\bibinfo{person}{Fadel Adib}, \bibinfo{person}{Zach Kabelac},
  \bibinfo{person}{Dina Katabi}, {and} \bibinfo{person}{Robert~C. Miller}.}
  \bibinfo{year}{2014}\natexlab{}.
\newblock \showarticletitle{{3D Tracking via Body Radio Reflections}}. In
  \bibinfo{booktitle}{\emph{Proc. of the 10th USENIX NSDI}}.
  \bibinfo{pages}{317--329}.
\newblock


\bibitem[\protect\citeauthoryear{Adib, Mao, Kabelac, Katabi, and Miller}{Adib
  et~al\mbox{.}}{2015}]%
        {adib2015smart}
\bibfield{author}{\bibinfo{person}{Fadel Adib}, \bibinfo{person}{Hongzi Mao},
  \bibinfo{person}{Zachary Kabelac}, \bibinfo{person}{Dina Katabi}, {and}
  \bibinfo{person}{Robert~C. Miller}.} \bibinfo{year}{2015}\natexlab{}.
\newblock \showarticletitle{Smart Homes that Monitor Breathing and Heart Rate}.
  In \bibinfo{booktitle}{\emph{Proc. of the 33rd ACM CHI}}.
  \bibinfo{pages}{837--846}.
\newblock


\bibitem[\protect\citeauthoryear{Avery, Gatewood, and Brumley}{Avery
  et~al\mbox{.}}{1966}]%
        {avery1966transient}
\bibfield{author}{\bibinfo{person}{Mary~Ellen Avery},
  \bibinfo{person}{Olga~Baghdassarian Gatewood}, {and} \bibinfo{person}{George
  Brumley}.} \bibinfo{year}{1966}\natexlab{}.
\newblock \showarticletitle{{Transient Tachypnea of Newborn: Possible Delayed
  Resorption of Fluid at Birth}}.
\newblock \bibinfo{journal}{\emph{American Journal of Diseases of Children}}
  \bibinfo{volume}{111}, \bibinfo{number}{4} (\bibinfo{year}{1966}),
  \bibinfo{pages}{380--385}.
\newblock


\bibitem[\protect\citeauthoryear{Bhatt, Kim, Wells, Bailey, Ramsdell, Foreman,
  Jensen, Stinson, Wilson, Lynch, et~al\mbox{.}}{Bhatt et~al\mbox{.}}{2014}]%
        {bhatt2014fev1}
\bibfield{author}{\bibinfo{person}{Surya~P. Bhatt}, \bibinfo{person}{Young-il
  Kim}, \bibinfo{person}{James~M. Wells}, \bibinfo{person}{William~C. Bailey},
  \bibinfo{person}{Joe~W. Ramsdell}, \bibinfo{person}{Marilyn~G. Foreman},
  \bibinfo{person}{Robert~L. Jensen}, \bibinfo{person}{Douglas~S. Stinson},
  \bibinfo{person}{Carla~G. Wilson}, \bibinfo{person}{David~A. Lynch},
  {et~al\mbox{.}}} \bibinfo{year}{2014}\natexlab{}.
\newblock \showarticletitle{{FEV1/FEV6 to Diagnose Airflow Obstruction.
  Comparisons with Computed Tomography and Morbidity Indices}}.
\newblock \bibinfo{journal}{\emph{Annals of the American Thoracic Society}}
  \bibinfo{volume}{11}, \bibinfo{number}{3} (\bibinfo{year}{2014}),
  \bibinfo{pages}{335--341}.
\newblock


\bibitem[\protect\citeauthoryear{Blei, Kucukelbir, and McAuliffe}{Blei
  et~al\mbox{.}}{2017}]%
        {blei2017variational}
\bibfield{author}{\bibinfo{person}{David~M. Blei}, \bibinfo{person}{Alp
  Kucukelbir}, {and} \bibinfo{person}{Jon~D. McAuliffe}.}
  \bibinfo{year}{2017}\natexlab{}.
\newblock \showarticletitle{{Variational Inference: A Review for
  Statisticians}}.
\newblock \bibinfo{journal}{\emph{Journal of the American statistical
  Association}} \bibinfo{volume}{112}, \bibinfo{number}{518}
  (\bibinfo{year}{2017}), \bibinfo{pages}{859--877}.
\newblock


\bibitem[\protect\citeauthoryear{Bottou}{Bottou}{2012}]%
        {bottou2012stochastic}
\bibfield{author}{\bibinfo{person}{L{\'e}on Bottou}.}
  \bibinfo{year}{2012}\natexlab{}.
\newblock \showarticletitle{{Stochastic Gradient Descent Tricks}}.
\newblock In \bibinfo{booktitle}{\emph{Neural Networks: Tricks of the Trade}}.
  \bibinfo{publisher}{Springer}, \bibinfo{pages}{421--436}.
\newblock


\bibitem[\protect\citeauthoryear{Brochard, Martin, Blanch, Pelosi, Belda,
  Jubran, Gattinoni, Mancebo, Ranieri, Richard, et~al\mbox{.}}{Brochard
  et~al\mbox{.}}{2012}]%
        {brochard2012clinical}
\bibfield{author}{\bibinfo{person}{Laurent Brochard}, \bibinfo{person}{Greg~S.
  Martin}, \bibinfo{person}{Lluis Blanch}, \bibinfo{person}{Paolo Pelosi},
  \bibinfo{person}{F.~Javier Belda}, \bibinfo{person}{Amal Jubran},
  \bibinfo{person}{Luciano Gattinoni}, \bibinfo{person}{Jordi Mancebo},
  \bibinfo{person}{V.~Marco Ranieri}, \bibinfo{person}{Jean-Christophe~M.
  Richard}, {et~al\mbox{.}}} \bibinfo{year}{2012}\natexlab{}.
\newblock \showarticletitle{{Clinical Review: Respiratory Monitoring in the
  ICU-A Consensus of 16}}.
\newblock \bibinfo{journal}{\emph{Critical Care}} \bibinfo{volume}{16},
  \bibinfo{number}{2} (\bibinfo{year}{2012}), \bibinfo{pages}{1--14}.
\newblock


\bibitem[\protect\citeauthoryear{Chen, Zheng, Cai, and Luo}{Chen
  et~al\mbox{.}}{2021b}]%
        {movifi}
\bibfield{author}{\bibinfo{person}{Zhe Chen}, \bibinfo{person}{Tianyue Zheng},
  \bibinfo{person}{Chao Cai}, {and} \bibinfo{person}{Jun Luo}.}
  \bibinfo{year}{2021}\natexlab{b}.
\newblock \showarticletitle{{MoVi-Fi: Motion-robust Vital Signs Waveform
  Recovery via Deep Interpreted RF Sensing}}. In
  \bibinfo{booktitle}{\emph{Proc. of the 27th ACM MobiCom}}.
  \bibinfo{pages}{1--14}.
\newblock


\bibitem[\protect\citeauthoryear{Chen, Zheng, and Luo}{Chen
  et~al\mbox{.}}{2021a}]%
        {octopus}
\bibfield{author}{\bibinfo{person}{Zhe Chen}, \bibinfo{person}{Tianyue Zheng},
  {and} \bibinfo{person}{Jun Luo}.} \bibinfo{year}{2021}\natexlab{a}.
\newblock \showarticletitle{{Octopus: A Practical and Versatile Wideband MIMO
  Sensing Platform}}. In \bibinfo{booktitle}{\emph{Proc. of the 27th ACM
  MobiCom}}. \bibinfo{pages}{1--14}.
\newblock


\bibitem[\protect\citeauthoryear{Chu, Nguyen, Pandey, Zhou, Pham, Bar-Yoseph,
  Radom-Aizik, Jain, Cooper, and Khine}{Chu et~al\mbox{.}}{2019}]%
        {chu2019respiration}
\bibfield{author}{\bibinfo{person}{Michael Chu}, \bibinfo{person}{Thao Nguyen},
  \bibinfo{person}{Vaibhav Pandey}, \bibinfo{person}{Yongxiao Zhou},
  \bibinfo{person}{Hoang~N. Pham}, \bibinfo{person}{Ronen Bar-Yoseph},
  \bibinfo{person}{Shlomit Radom-Aizik}, \bibinfo{person}{Ramesh Jain},
  \bibinfo{person}{Dan~M. Cooper}, {and} \bibinfo{person}{Michelle Khine}.}
  \bibinfo{year}{2019}\natexlab{}.
\newblock \showarticletitle{{Respiration Rate and Volume Measurements Using
  Wearable Strain Sensors}}.
\newblock \bibinfo{journal}{\emph{NPJ digital medicine}} \bibinfo{volume}{2},
  \bibinfo{number}{1} (\bibinfo{year}{2019}), \bibinfo{pages}{1--9}.
\newblock


\bibitem[\protect\citeauthoryear{Corren, Togias, and Bousquet}{Corren
  et~al\mbox{.}}{2003}]%
        {corren2003upper}
\bibfield{author}{\bibinfo{person}{Jonathan Corren}, \bibinfo{person}{Alkis
  Togias}, {and} \bibinfo{person}{Jean Bousquet}.}
  \bibinfo{year}{2003}\natexlab{}.
\newblock \bibinfo{booktitle}{\emph{{Upper and Lower Respiratory Disease}}}.
\newblock \bibinfo{publisher}{CRC Press}.
\newblock


\bibitem[\protect\citeauthoryear{Ding, Chen, Zheng, and Luo}{Ding
  et~al\mbox{.}}{2020}]%
        {ding2020rfnet}
\bibfield{author}{\bibinfo{person}{Shuya Ding}, \bibinfo{person}{Zhe Chen},
  \bibinfo{person}{Tianyue Zheng}, {and} \bibinfo{person}{Jun Luo}.}
  \bibinfo{year}{2020}\natexlab{}.
\newblock \showarticletitle{RF-Net: A Unified Meta-Learning Framework for
  RF-Enabled One-Shot Human Activity Recognition}. In
  \bibinfo{booktitle}{\emph{Proc. of the 18th ACM SenSys}}.
  \bibinfo{pages}{517–530}.
\newblock


\bibitem[\protect\citeauthoryear{Fang, Lane, Zhang, Boran, and Kawsar}{Fang
  et~al\mbox{.}}{2016}]%
        {fang2016bodyscan}
\bibfield{author}{\bibinfo{person}{Biyi Fang}, \bibinfo{person}{Nicholas~D.
  Lane}, \bibinfo{person}{Mi Zhang}, \bibinfo{person}{Aidan Boran}, {and}
  \bibinfo{person}{Fahim Kawsar}.} \bibinfo{year}{2016}\natexlab{}.
\newblock \showarticletitle{{BodyScan: Enabling Radio-based Sensing on Wearable
  Devices for Contactless Activity and Vital Sign Monitoring}}. In
  \bibinfo{booktitle}{\emph{Proc. of the 14th ACM MobiSys}}.
  \bibinfo{pages}{97--110}.
\newblock


\bibitem[\protect\citeauthoryear{Farney, Walker, Boyle, Cloward, and
  Shilling}{Farney et~al\mbox{.}}{2008}]%
        {farney2008adaptive}
\bibfield{author}{\bibinfo{person}{Robert~J. Farney}, \bibinfo{person}{James~M.
  Walker}, \bibinfo{person}{Kathleen~M. Boyle}, \bibinfo{person}{Tom~V.
  Cloward}, {and} \bibinfo{person}{Kevin~C. Shilling}.}
  \bibinfo{year}{2008}\natexlab{}.
\newblock \showarticletitle{{Adaptive Servoventilation (ASV) in Patients with
  Sleep Disordered Breathing Associated with Chronic Opioid Medications for
  Non-Malignant Pain}}.
\newblock \bibinfo{journal}{\emph{Journal of Clinical Sleep Medicine}}
  \bibinfo{volume}{4}, \bibinfo{number}{4} (\bibinfo{year}{2008}),
  \bibinfo{pages}{311--319}.
\newblock


\bibitem[\protect\citeauthoryear{Flenley}{Flenley}{1985}]%
        {flenley1985sleep}
\bibfield{author}{\bibinfo{person}{D.C. Flenley}.}
  \bibinfo{year}{1985}\natexlab{}.
\newblock \showarticletitle{{Sleep in Chronic Obstructive Lung Disease}}.
\newblock \bibinfo{journal}{\emph{Clinics in Chest Medicine}}
  \bibinfo{volume}{6}, \bibinfo{number}{4} (\bibinfo{year}{1985}),
  \bibinfo{pages}{651--661}.
\newblock


\bibitem[\protect\citeauthoryear{Fletcher and Peto}{Fletcher and Peto}{1977}]%
        {fletcher1977natural}
\bibfield{author}{\bibinfo{person}{Charles Fletcher} {and}
  \bibinfo{person}{Richard Peto}.} \bibinfo{year}{1977}\natexlab{}.
\newblock \showarticletitle{{The Natural History of Chronic Airflow
  Obstruction}}.
\newblock \bibinfo{journal}{\emph{Br Med J}} \bibinfo{volume}{1},
  \bibinfo{number}{6077} (\bibinfo{year}{1977}), \bibinfo{pages}{1645--1648}.
\newblock


\bibitem[\protect\citeauthoryear{Folke, Cernerud, Ekstr{\"o}m, and
  H{\"o}k}{Folke et~al\mbox{.}}{2003}]%
        {folke2003critical}
\bibfield{author}{\bibinfo{person}{Mia Folke}, \bibinfo{person}{Lars Cernerud},
  \bibinfo{person}{Martin Ekstr{\"o}m}, {and} \bibinfo{person}{Bertil
  H{\"o}k}.} \bibinfo{year}{2003}\natexlab{}.
\newblock \showarticletitle{{Critical Review of Non-Invasive Respiratory
  Monitoring in Medical Care}}.
\newblock \bibinfo{journal}{\emph{Medical and Biological Engineering and
  Computing}} \bibinfo{volume}{41}, \bibinfo{number}{4} (\bibinfo{year}{2003}),
  \bibinfo{pages}{377--383}.
\newblock


\bibitem[\protect\citeauthoryear{Gift, Moore, and Soeken}{Gift
  et~al\mbox{.}}{1992}]%
        {gift1992relaxation}
\bibfield{author}{\bibinfo{person}{Audrey~G. Gift}, \bibinfo{person}{Trellis
  Moore}, {and} \bibinfo{person}{Karen Soeken}.}
  \bibinfo{year}{1992}\natexlab{}.
\newblock \showarticletitle{{Relaxation to Reduce Dyspnea and Anxiety in COPD
  Patients}}.
\newblock \bibinfo{journal}{\emph{Nursing Research}} (\bibinfo{year}{1992}).
\newblock


\bibitem[\protect\citeauthoryear{Givens, Shortt, et~al\mbox{.}}{Givens
  et~al\mbox{.}}{1984}]%
        {givens1984class}
\bibfield{author}{\bibinfo{person}{Clark~R. Givens},
  \bibinfo{person}{Rae~Michael Shortt}, {et~al\mbox{.}}}
  \bibinfo{year}{1984}\natexlab{}.
\newblock \showarticletitle{{A Class of Wasserstein Metrics for Probability
  Distributions}}.
\newblock \bibinfo{journal}{\emph{The Michigan Mathematical Journal}}
  \bibinfo{volume}{31}, \bibinfo{number}{2} (\bibinfo{year}{1984}),
  \bibinfo{pages}{231--240}.
\newblock


\bibitem[\protect\citeauthoryear{Glorot, Bordes, and Bengio}{Glorot
  et~al\mbox{.}}{2011}]%
        {glorot2011deep}
\bibfield{author}{\bibinfo{person}{Xavier Glorot}, \bibinfo{person}{Antoine
  Bordes}, {and} \bibinfo{person}{Yoshua Bengio}.}
  \bibinfo{year}{2011}\natexlab{}.
\newblock \showarticletitle{{Deep Sparse Rectifier Neural Networks}}. In
  \bibinfo{booktitle}{\emph{Proc. of the 14th AISTATS}}.
  \bibinfo{pages}{315--323}.
\newblock


\bibitem[\protect\citeauthoryear{Goodfellow, Pouget-Abadie, Mirza, Xu,
  Warde-Farley, Ozair, Courville, and Bengio}{Goodfellow et~al\mbox{.}}{2014}]%
        {goodfellow2014generative}
\bibfield{author}{\bibinfo{person}{Ian~J. Goodfellow}, \bibinfo{person}{Jean
  Pouget-Abadie}, \bibinfo{person}{Mehdi Mirza}, \bibinfo{person}{Bing Xu},
  \bibinfo{person}{David Warde-Farley}, \bibinfo{person}{Sherjil Ozair},
  \bibinfo{person}{Aaron Courville}, {and} \bibinfo{person}{Yoshua Bengio}.}
  \bibinfo{year}{2014}\natexlab{}.
\newblock \showarticletitle{{Generative Adversarial Networks}}. In
  \bibinfo{booktitle}{\emph{Proc. of The 27th NIPS}}. \bibinfo{pages}{1--9}.
\newblock


\bibitem[\protect\citeauthoryear{Gu and Li}{Gu and Li}{2015}]%
        {gu2015assessment}
\bibfield{author}{\bibinfo{person}{Changzhan Gu} {and}
  \bibinfo{person}{Changzhi Li}.} \bibinfo{year}{2015}\natexlab{}.
\newblock \showarticletitle{{Assessment of Human Respiration Patterns via
  Noncontact Sensing using Doppler Multi-Radar System}}.
\newblock \bibinfo{journal}{\emph{Sensors}} \bibinfo{volume}{15},
  \bibinfo{number}{3} (\bibinfo{year}{2015}), \bibinfo{pages}{6383--6398}.
\newblock


\bibitem[\protect\citeauthoryear{G{\"u}der, Ainla, Redston, Mosadegh, Glavan,
  Martin, and Whitesides}{G{\"u}der et~al\mbox{.}}{2016}]%
        {guder2016paper}
\bibfield{author}{\bibinfo{person}{Firat G{\"u}der}, \bibinfo{person}{Alar
  Ainla}, \bibinfo{person}{Julia Redston}, \bibinfo{person}{Bobak Mosadegh},
  \bibinfo{person}{Ana Glavan}, \bibinfo{person}{T.J. Martin}, {and}
  \bibinfo{person}{George~M. Whitesides}.} \bibinfo{year}{2016}\natexlab{}.
\newblock \showarticletitle{{Paper-based Electrical Respiration Sensor}}.
\newblock \bibinfo{journal}{\emph{Angewandte Chemie International Edition}}
  \bibinfo{volume}{55}, \bibinfo{number}{19} (\bibinfo{year}{2016}),
  \bibinfo{pages}{5727--5732}.
\newblock


\bibitem[\protect\citeauthoryear{Hao, Bi, Xing, Chan, and Tu}{Hao
  et~al\mbox{.}}{2017}]%
        {hao2017mindfulwatch}
\bibfield{author}{\bibinfo{person}{Tian Hao}, \bibinfo{person}{Chongguang Bi},
  \bibinfo{person}{Guoliang Xing}, \bibinfo{person}{Roxane Chan}, {and}
  \bibinfo{person}{Linlin Tu}.} \bibinfo{year}{2017}\natexlab{}.
\newblock \showarticletitle{MindfulWatch: A Smartwatch-based System for
  Real-Time Respiration Monitoring during Meditation}. In
  \bibinfo{booktitle}{\emph{Proc. of the 19th ACM UbiComp}}.
  \bibinfo{pages}{1--19}.
\newblock


\bibitem[\protect\citeauthoryear{Haponik, Bleecker, Allen, Smith, and
  Kaplan}{Haponik et~al\mbox{.}}{1981}]%
        {haponik1981abnormal}
\bibfield{author}{\bibinfo{person}{Edward~F. Haponik},
  \bibinfo{person}{Eugene~R. Bleecker}, \bibinfo{person}{Richard~P. Allen},
  \bibinfo{person}{Philip~L. Smith}, {and} \bibinfo{person}{Joseph Kaplan}.}
  \bibinfo{year}{1981}\natexlab{}.
\newblock \showarticletitle{{Abnormal Inspiratory Flow-Volume Curves in
  Patients with Sleep-Disordered Breathing}}.
\newblock \bibinfo{journal}{\emph{American Review of Respiratory Disease}}
  \bibinfo{volume}{124}, \bibinfo{number}{5} (\bibinfo{year}{1981}),
  \bibinfo{pages}{571--574}.
\newblock


\bibitem[\protect\citeauthoryear{Higgins, Matthey, Pal, Burgess, Glorot,
  Botvinick, Mohamed, and Lerchner}{Higgins et~al\mbox{.}}{2016}]%
        {higgins2016beta}
\bibfield{author}{\bibinfo{person}{Irina Higgins}, \bibinfo{person}{Loic
  Matthey}, \bibinfo{person}{Arka Pal}, \bibinfo{person}{Christopher Burgess},
  \bibinfo{person}{Xavier Glorot}, \bibinfo{person}{Matthew Botvinick},
  \bibinfo{person}{Shakir Mohamed}, {and} \bibinfo{person}{Alexander
  Lerchner}.} \bibinfo{year}{2016}\natexlab{}.
\newblock \showarticletitle{{beta-VAE: Learning Basic Visual Concepts with a
  Constrained Variational Framework}}.
\newblock  (\bibinfo{year}{2016}).
\newblock


\bibitem[\protect\citeauthoryear{Hoffman, Blei, Wang, and Paisley}{Hoffman
  et~al\mbox{.}}{2013}]%
        {hoffman2013stochastic}
\bibfield{author}{\bibinfo{person}{Matthew~D. Hoffman},
  \bibinfo{person}{David~M. Blei}, \bibinfo{person}{Chong Wang}, {and}
  \bibinfo{person}{John Paisley}.} \bibinfo{year}{2013}\natexlab{}.
\newblock \showarticletitle{{Stochastic Variational Inference}}.
\newblock \bibinfo{journal}{\emph{Journal of Machine Learning Research}}
  \bibinfo{volume}{14}, \bibinfo{number}{5} (\bibinfo{year}{2013}).
\newblock


\bibitem[\protect\citeauthoryear{Holland, Hill, Jones, and McDonald}{Holland
  et~al\mbox{.}}{2012}]%
        {holland2012breathing}
\bibfield{author}{\bibinfo{person}{Anne~E. Holland},
  \bibinfo{person}{Catherine~J. Hill}, \bibinfo{person}{Alice~Y. Jones}, {and}
  \bibinfo{person}{Christine~F. McDonald}.} \bibinfo{year}{2012}\natexlab{}.
\newblock \showarticletitle{{Breathing Exercises for Chronic Obstructive
  Pulmonary Disease}}.
\newblock \bibinfo{journal}{\emph{Cochrane Database of Systematic Reviews}}
  \bibinfo{number}{10} (\bibinfo{year}{2012}).
\newblock


\bibitem[\protect\citeauthoryear{Holoborodko}{Holoborodko}{2008}]%
        {snrd}
\bibfield{author}{\bibinfo{person}{Pavel Holoborodko}.}
  \bibinfo{year}{2008}\natexlab{}.
\newblock \bibinfo{title}{{Smooth Noise Robust Differentiators}}.
\newblock
  \bibinfo{howpublished}{http://www.holoborodko.com/pavel/numerical-methods/numerical-derivative/smooth-low-noise-differentiators/}.
\newblock


\bibitem[\protect\citeauthoryear{Hyv\"{a}rinen and Pajunen}{Hyv\"{a}rinen and
  Pajunen}{1999}]%
        {hyv1999Noica}
\bibfield{author}{\bibinfo{person}{Aapo Hyv\"{a}rinen} {and}
  \bibinfo{person}{Petteri Pajunen}.} \bibinfo{year}{1999}\natexlab{}.
\newblock \showarticletitle{Nonlinear Independent Component Analysis: Existence
  and Uniqueness Results}.
\newblock \bibinfo{journal}{\emph{Neural Netw.}} \bibinfo{volume}{12},
  \bibinfo{number}{3} (\bibinfo{year}{1999}), \bibinfo{pages}{429–439}.
\newblock


\bibitem[\protect\citeauthoryear{IEFT}{IEFT}{2017}]%
        {ptp}
\bibfield{author}{\bibinfo{person}{IEFT}.} \bibinfo{year}{2017}\natexlab{}.
\newblock \bibinfo{title}{{Precision Time Protocol Version 2 (PTPv2)}}.
\newblock
\newblock
\newblock
\shownote{Accessed: 2021-04-30.}


\bibitem[\protect\citeauthoryear{Ioffe and Szegedy}{Ioffe and Szegedy}{2015}]%
        {ioffe2015batch}
\bibfield{author}{\bibinfo{person}{Sergey Ioffe} {and}
  \bibinfo{person}{Christian Szegedy}.} \bibinfo{year}{2015}\natexlab{}.
\newblock \showarticletitle{{Batch Normalization: Accelerating Deep Network
  Training by Reducing Internal Covariate Shift}}. In
  \bibinfo{booktitle}{\emph{Proc. of ICML}}. PMLR, \bibinfo{pages}{448--456}.
\newblock


\bibitem[\protect\citeauthoryear{Jia, Bonde, Li, Xu, Wang, Zhang, Howard, and
  Zhang}{Jia et~al\mbox{.}}{2017}]%
        {jia2017monitoring}
\bibfield{author}{\bibinfo{person}{Zhenhua Jia}, \bibinfo{person}{Amelie
  Bonde}, \bibinfo{person}{Sugang Li}, \bibinfo{person}{Chenren Xu},
  \bibinfo{person}{Jingxian Wang}, \bibinfo{person}{Yanyong Zhang},
  \bibinfo{person}{Richard~E. Howard}, {and} \bibinfo{person}{Pei Zhang}.}
  \bibinfo{year}{2017}\natexlab{}.
\newblock \showarticletitle{{Monitoring a Person's Heart Rate and Respiratory
  Rate on a Shared Bed using Geophones}}. In \bibinfo{booktitle}{\emph{Proc. of
  the 15th ACM SenSys}}. \bibinfo{pages}{1--14}.
\newblock


\bibitem[\protect\citeauthoryear{Jutten and Karhunen}{Jutten and
  Karhunen}{2003}]%
        {jutten2003advances}
\bibfield{author}{\bibinfo{person}{Christian Jutten} {and}
  \bibinfo{person}{Juha Karhunen}.} \bibinfo{year}{2003}\natexlab{}.
\newblock \showarticletitle{{Advances in Nonlinear Blind Source Separation}}.
  In \bibinfo{booktitle}{\emph{Proc. of the 4th Int. Symp. on Independent
  Component Analysis and Blind Signal Separation (ICA2003)}}.
  \bibinfo{pages}{245--256}.
\newblock


\bibitem[\protect\citeauthoryear{Kaltiokallio, Yi{\u{g}}itler, J{\"a}ntti, and
  Patwari}{Kaltiokallio et~al\mbox{.}}{2014}]%
        {kaltiokallio2014non}
\bibfield{author}{\bibinfo{person}{Ossi Kaltiokallio},
  \bibinfo{person}{H{\"u}seyin Yi{\u{g}}itler}, \bibinfo{person}{Riku
  J{\"a}ntti}, {and} \bibinfo{person}{Neal Patwari}.}
  \bibinfo{year}{2014}\natexlab{}.
\newblock \showarticletitle{{Non-Invasive Respiration Rate Monitoring using a
  Single COTS TX-RX Pair}}. In \bibinfo{booktitle}{\emph{Proc. of the 13th ACM
  IPSN}}. IEEE, \bibinfo{pages}{59--69}.
\newblock


\bibitem[\protect\citeauthoryear{Khemakhem, Kingma, Monti, and
  Hyvarinen}{Khemakhem et~al\mbox{.}}{2020}]%
        {khemakhem2020variational}
\bibfield{author}{\bibinfo{person}{Ilyes Khemakhem}, \bibinfo{person}{Diederik
  Kingma}, \bibinfo{person}{Ricardo Monti}, {and} \bibinfo{person}{Aapo
  Hyvarinen}.} \bibinfo{year}{2020}\natexlab{}.
\newblock \showarticletitle{{Variational Autoencoders and Nonlinear ICA: A
  Unifying Framework}}. In \bibinfo{booktitle}{\emph{Proc. of the 20th
  AISTATS}}. PMLR, \bibinfo{pages}{2207--2217}.
\newblock


\bibitem[\protect\citeauthoryear{Kingma and Welling}{Kingma and
  Welling}{2013}]%
        {kingma2013auto}
\bibfield{author}{\bibinfo{person}{Diederik~P. Kingma} {and}
  \bibinfo{person}{Max Welling}.} \bibinfo{year}{2013}\natexlab{}.
\newblock \showarticletitle{{Auto-Encoding Variational Bayes}}. In
  \bibinfo{booktitle}{\emph{Proc. of ICLR}}. \bibinfo{pages}{1--14}.
\newblock


\bibitem[\protect\citeauthoryear{Kolouri, Pope, Martin, and Rohde}{Kolouri
  et~al\mbox{.}}{2018}]%
        {kolouri2018sliced}
\bibfield{author}{\bibinfo{person}{Soheil Kolouri}, \bibinfo{person}{Phillip~E.
  Pope}, \bibinfo{person}{Charles~E. Martin}, {and} \bibinfo{person}{Gustavo~K.
  Rohde}.} \bibinfo{year}{2018}\natexlab{}.
\newblock \showarticletitle{{Sliced Wasserstein Auto-Encoders}}. In
  \bibinfo{booktitle}{\emph{Proc. of ICLR}}. \bibinfo{pages}{1--19}.
\newblock


\bibitem[\protect\citeauthoryear{Kullback and Leibler}{Kullback and
  Leibler}{1951}]%
        {kullback1951information}
\bibfield{author}{\bibinfo{person}{Solomon Kullback} {and}
  \bibinfo{person}{Richard~A. Leibler}.} \bibinfo{year}{1951}\natexlab{}.
\newblock \showarticletitle{{On Information and Sufficiency}}.
\newblock \bibinfo{journal}{\emph{The Annals of Mathematical Statistics}}
  \bibinfo{volume}{22}, \bibinfo{number}{1} (\bibinfo{year}{1951}),
  \bibinfo{pages}{79--86}.
\newblock


\bibitem[\protect\citeauthoryear{Kumar}{Kumar}{2017}]%
        {kumar2017weight}
\bibfield{author}{\bibinfo{person}{Siddharth~Krishna Kumar}.}
  \bibinfo{year}{2017}\natexlab{}.
\newblock \showarticletitle{{On Weight Initialization in Deep Neural
  Networks}}.
\newblock \bibinfo{journal}{\emph{arXiv preprint arXiv:1704.08863}}
  (\bibinfo{year}{2017}).
\newblock


\bibitem[\protect\citeauthoryear{Lanfranchi, Braghiroli, Bosimini, Mazzuero,
  Colombo, Donner, and Giannuzzi}{Lanfranchi et~al\mbox{.}}{1999}]%
        {lanfranchi1999prognostic}
\bibfield{author}{\bibinfo{person}{Paola~A. Lanfranchi},
  \bibinfo{person}{Alberto Braghiroli}, \bibinfo{person}{Enzo Bosimini},
  \bibinfo{person}{Giorgio Mazzuero}, \bibinfo{person}{Roberto Colombo},
  \bibinfo{person}{Claudio~F. Donner}, {and} \bibinfo{person}{Pantaleo
  Giannuzzi}.} \bibinfo{year}{1999}\natexlab{}.
\newblock \showarticletitle{{Prognostic Value of Nocturnal Cheyne-Stokes
  Respiration in Chronic Heart Failure}}.
\newblock \bibinfo{journal}{\emph{Circulation}} \bibinfo{volume}{99},
  \bibinfo{number}{11} (\bibinfo{year}{1999}), \bibinfo{pages}{1435--1440}.
\newblock


\bibitem[\protect\citeauthoryear{LeCun, Haffner, Bottou, and Bengio}{LeCun
  et~al\mbox{.}}{1999}]%
        {lecun1999object}
\bibfield{author}{\bibinfo{person}{Yann LeCun}, \bibinfo{person}{Patrick
  Haffner}, \bibinfo{person}{L{\'e}on Bottou}, {and} \bibinfo{person}{Yoshua
  Bengio}.} \bibinfo{year}{1999}\natexlab{}.
\newblock \showarticletitle{{Object Recognition with Gradient-Based Learning}}.
\newblock In \bibinfo{booktitle}{\emph{Shape, Contour and Grouping in Computer
  Vision}}. \bibinfo{publisher}{Springer}, \bibinfo{pages}{319--345}.
\newblock


\bibitem[\protect\citeauthoryear{Lee, Gao, Xu, and Boric-Lubecke}{Lee
  et~al\mbox{.}}{2017}]%
        {lee2017effects}
\bibfield{author}{\bibinfo{person}{Alexander Lee}, \bibinfo{person}{Xiaomeng
  Gao}, \bibinfo{person}{Jia Xu}, {and} \bibinfo{person}{Olga Boric-Lubecke}.}
  \bibinfo{year}{2017}\natexlab{}.
\newblock \showarticletitle{{Effects of Respiration Depth on Human Body Radar
  Cross Section Using 2.4 GHz Continuous Wave Radar}}. In
  \bibinfo{booktitle}{\emph{2017 39th Annual International Conference of the
  IEEE Engineering in Medicine and Biology Society (EMBC)}}. IEEE,
  \bibinfo{pages}{4070--4073}.
\newblock


\bibitem[\protect\citeauthoryear{Levanon}{Levanon}{1988}]%
        {levanon1988radar}
\bibfield{author}{\bibinfo{person}{Nadav Levanon}.}
  \bibinfo{year}{1988}\natexlab{}.
\newblock \bibinfo{booktitle}{\emph{{Radar Principles}}}.
\newblock \bibinfo{publisher}{Wiley}.
\newblock


\bibitem[\protect\citeauthoryear{Li and Lin}{Li and Lin}{2008}]%
        {changzhili2008Random}
\bibfield{author}{\bibinfo{person}{Changzhi Li} {and} \bibinfo{person}{Jenshan
  Lin}.} \bibinfo{year}{2008}\natexlab{}.
\newblock \showarticletitle{Random Body Movement Cancellation in Doppler Radar
  Vital Sign Detection}.
\newblock \bibinfo{journal}{\emph{IEEE Transactions on Microwave Theory and
  Techniques}} \bibinfo{volume}{56}, \bibinfo{number}{12}
  (\bibinfo{year}{2008}), \bibinfo{pages}{3143--3152}.
\newblock


\bibitem[\protect\citeauthoryear{Luo and Mesgarani}{Luo and Mesgarani}{2018}]%
        {luo2018tasnet}
\bibfield{author}{\bibinfo{person}{Yi Luo} {and} \bibinfo{person}{Nima
  Mesgarani}.} \bibinfo{year}{2018}\natexlab{}.
\newblock \showarticletitle{{TasNet: Time-Domain Audio Separation Network for
  Real-Time, Single-Channel Speech Separation}}. In
  \bibinfo{booktitle}{\emph{2018 IEEE International Conference on Acoustics,
  Speech and Signal Processing (ICASSP)}}. IEEE, \bibinfo{pages}{696--700}.
\newblock


\bibitem[\protect\citeauthoryear{Lv, Chen, An, Wang, Li, Ye, Huangfu, Li, and
  Ran}{Lv et~al\mbox{.}}{2018}]%
        {lv2018large}
\bibfield{author}{\bibinfo{person}{Qinyi Lv}, \bibinfo{person}{Lei Chen},
  \bibinfo{person}{Kang An}, \bibinfo{person}{Jun Wang}, \bibinfo{person}{Huan
  Li}, \bibinfo{person}{Dexin Ye}, \bibinfo{person}{Jiangtao Huangfu},
  \bibinfo{person}{Changzhi Li}, {and} \bibinfo{person}{Lixin Ran}.}
  \bibinfo{year}{2018}\natexlab{}.
\newblock \showarticletitle{{Doppler Vital Signs Detection in the Presence of
  Large-Scale Random Body Movements}}.
\newblock \bibinfo{journal}{\emph{IEEE Transactions on Microwave Theory and
  Techniques}} \bibinfo{volume}{66}, \bibinfo{number}{9}
  (\bibinfo{year}{2018}), \bibinfo{pages}{4261--4270}.
\newblock


\bibitem[\protect\citeauthoryear{Manning and Schwartzstein}{Manning and
  Schwartzstein}{1995}]%
        {manning1995pathophysiology}
\bibfield{author}{\bibinfo{person}{Harold~L. Manning} {and}
  \bibinfo{person}{Richard~M. Schwartzstein}.} \bibinfo{year}{1995}\natexlab{}.
\newblock \showarticletitle{{Pathophysiology of Dyspnea}}.
\newblock \bibinfo{journal}{\emph{New England Journal of Medicine}}
  \bibinfo{volume}{333}, \bibinfo{number}{23} (\bibinfo{year}{1995}),
  \bibinfo{pages}{1547--1553}.
\newblock


\bibitem[\protect\citeauthoryear{Miller, Brown, and Teirstein}{Miller
  et~al\mbox{.}}{1985}]%
        {miller1985stenosis}
\bibfield{author}{\bibinfo{person}{Albert Miller}, \bibinfo{person}{Lee~K.
  Brown}, {and} \bibinfo{person}{Alvin~S. Teirstein}.}
  \bibinfo{year}{1985}\natexlab{}.
\newblock \showarticletitle{{Stenosis of Main Bronchi Mimicking Fixed Upper
  Airway Obstruction in Sarcoidosis}}.
\newblock \bibinfo{journal}{\emph{Chest}} \bibinfo{volume}{88},
  \bibinfo{number}{2} (\bibinfo{year}{1985}), \bibinfo{pages}{244--248}.
\newblock


\bibitem[\protect\citeauthoryear{Mimilakis, Drossos, Virtanen, and
  Schuller}{Mimilakis et~al\mbox{.}}{2017}]%
        {mimilakis2017recurrent}
\bibfield{author}{\bibinfo{person}{Stylianos~Ioannis Mimilakis},
  \bibinfo{person}{Konstantinos Drossos}, \bibinfo{person}{Tuomas Virtanen},
  {and} \bibinfo{person}{Gerald Schuller}.} \bibinfo{year}{2017}\natexlab{}.
\newblock \showarticletitle{{A Recurrent Encoder-Decoder Approach with
  Skip-Filtering Connections for Monaural Singing Voice Separation}}. In
  \bibinfo{booktitle}{\emph{2017 IEEE 27th International Workshop on Machine
  Learning for Signal Processing (MLSP)}}. IEEE, \bibinfo{pages}{1--6}.
\newblock


\bibitem[\protect\citeauthoryear{Min, Kim, Shin, Yun, Lee, and Lee}{Min
  et~al\mbox{.}}{2010}]%
        {min2010noncontact}
\bibfield{author}{\bibinfo{person}{Se~Dong Min}, \bibinfo{person}{Jin~Kwon
  Kim}, \bibinfo{person}{Hang~Sik Shin}, \bibinfo{person}{Yong~Hyeon Yun},
  \bibinfo{person}{Chung~Keun Lee}, {and} \bibinfo{person}{Myoungho Lee}.}
  \bibinfo{year}{2010}\natexlab{}.
\newblock \showarticletitle{{Noncontact Respiration Rate Measurement System
  Using an Ultrasonic Proximity Sensor}}.
\newblock \bibinfo{journal}{\emph{IEEE Sensors Journal}} \bibinfo{volume}{10},
  \bibinfo{number}{11} (\bibinfo{year}{2010}), \bibinfo{pages}{1732--1739}.
\newblock


\bibitem[\protect\citeauthoryear{Min, Yun, and Shin}{Min et~al\mbox{.}}{2014}]%
        {min2014simplified}
\bibfield{author}{\bibinfo{person}{Se~Dong Min}, \bibinfo{person}{Yonghyeon
  Yun}, {and} \bibinfo{person}{Hangsik Shin}.} \bibinfo{year}{2014}\natexlab{}.
\newblock \showarticletitle{{Simplified Structural Textile Respiration Sensor
  based on Capacitive Pressure Sensing Method}}.
\newblock \bibinfo{journal}{\emph{IEEE Sensors Journal}} \bibinfo{volume}{14},
  \bibinfo{number}{9} (\bibinfo{year}{2014}), \bibinfo{pages}{3245--3251}.
\newblock


\bibitem[\protect\citeauthoryear{Muñoz-Ferreras, Peng, Gómez-García, and
  Li}{Muñoz-Ferreras et~al\mbox{.}}{2016}]%
        {cli2016random}
\bibfield{author}{\bibinfo{person}{José-María Muñoz-Ferreras},
  \bibinfo{person}{Zhengyu Peng}, \bibinfo{person}{Roberto Gómez-García},
  {and} \bibinfo{person}{Changzhi Li}.} \bibinfo{year}{2016}\natexlab{}.
\newblock \showarticletitle{Random Body Movement Mitigation for
  FMCW-radar-based Vital-Sign Monitoring}. In \bibinfo{booktitle}{\emph{IEEE
  Topical Conference on Biomedical Wireless Technologies, Networks, and Sensing
  Systems}}. \bibinfo{pages}{22--24}.
\newblock


\bibitem[\protect\citeauthoryear{Naji, Connor, Donnelly, and McDonnell}{Naji
  et~al\mbox{.}}{2006}]%
        {naji2006effectiveness}
\bibfield{author}{\bibinfo{person}{Nizar~A. Naji}, \bibinfo{person}{Marian~C.
  Connor}, \bibinfo{person}{Seamas~C. Donnelly}, {and}
  \bibinfo{person}{Timothy~J. McDonnell}.} \bibinfo{year}{2006}\natexlab{}.
\newblock \showarticletitle{{Effectiveness of Pulmonary Rehabilitation in
  Restrictive Lung Disease}}.
\newblock \bibinfo{journal}{\emph{Journal of Cardiopulmonary Rehabilitation and
  Prevention}} \bibinfo{volume}{26}, \bibinfo{number}{4}
  (\bibinfo{year}{2006}), \bibinfo{pages}{237--243}.
\newblock


\bibitem[\protect\citeauthoryear{NeuLog}{NeuLog}{2017}]%
        {neulog}
\bibfield{author}{\bibinfo{person}{NeuLog}.} \bibinfo{year}{2017}\natexlab{}.
\newblock \bibinfo{title}{Respiration Monitor Belt Logger Sensor NUL-236}.
\newblock
  \bibinfo{howpublished}{\url{https://neulog.com/respiration-monitor-belt/}}.
\newblock
\newblock
\shownote{Accessed: 2021-04-28.}


\bibitem[\protect\citeauthoryear{Nguyen, Zhang, Halbower, and Vu}{Nguyen
  et~al\mbox{.}}{2016}]%
        {nguyen2016continuous}
\bibfield{author}{\bibinfo{person}{Phuc Nguyen}, \bibinfo{person}{Xinyu Zhang},
  \bibinfo{person}{Ann Halbower}, {and} \bibinfo{person}{Tam Vu}.}
  \bibinfo{year}{2016}\natexlab{}.
\newblock \showarticletitle{Continuous and Fine-Grained Breathing Volume
  Monitoring from Afar Using Wireless Signals}. In
  \bibinfo{booktitle}{\emph{Proc. of the 35th IEEE INFOCOM}}.
  \bibinfo{pages}{1--9}.
\newblock


\bibitem[\protect\citeauthoryear{{Novelda AS}}{{Novelda AS}}{2017}]%
        {xethru}
\bibfield{author}{\bibinfo{person}{{Novelda AS}}.}
  \bibinfo{year}{2017}\natexlab{}.
\newblock \bibinfo{title}{{The World Leader in Ultra Wideband (UWB) Sensing}}.
\newblock \bibinfo{howpublished}{\url{https://novelda.com/technology/}}.
\newblock
\newblock
\shownote{Accessed: 2021-04-22.}


\bibitem[\protect\citeauthoryear{Olkin and Pukelsheim}{Olkin and
  Pukelsheim}{1982}]%
        {olkin1982distance}
\bibfield{author}{\bibinfo{person}{Ingram Olkin} {and}
  \bibinfo{person}{Friedrich Pukelsheim}.} \bibinfo{year}{1982}\natexlab{}.
\newblock \showarticletitle{{The Distance between Two Random Vectors with Given
  Dispersion Matrices}}.
\newblock \bibinfo{journal}{\emph{Linear Algebra Appl.}}  \bibinfo{volume}{48}
  (\bibinfo{year}{1982}), \bibinfo{pages}{257--263}.
\newblock


\bibitem[\protect\citeauthoryear{Organization}{Organization}{2020}]%
        {who}
\bibfield{author}{\bibinfo{person}{World~Health Organization}.}
  \bibinfo{year}{2020}\natexlab{}.
\newblock \bibinfo{title}{{The Top 10 Causes of Death}}.
\newblock
  \bibinfo{howpublished}{\url{https://www.who.int/news-room/fact-sheets/detail/the-top-10-causes-of-death}}.
\newblock
\newblock
\shownote{Accessed: 2021-04-14.}


\bibitem[\protect\citeauthoryear{Paszke, Gross, Massa, Lerer, Bradbury, Chanan,
  Killeen, Lin, Gimelshein, Antiga, et~al\mbox{.}}{Paszke
  et~al\mbox{.}}{2019}]%
        {pytorch}
\bibfield{author}{\bibinfo{person}{Adam Paszke}, \bibinfo{person}{Sam Gross},
  \bibinfo{person}{Francisco Massa}, \bibinfo{person}{Adam Lerer},
  \bibinfo{person}{James Bradbury}, \bibinfo{person}{Gregory Chanan},
  \bibinfo{person}{Trevor Killeen}, \bibinfo{person}{Zeming Lin},
  \bibinfo{person}{Natalia Gimelshein}, \bibinfo{person}{Luca Antiga},
  {et~al\mbox{.}}} \bibinfo{year}{2019}\natexlab{}.
\newblock \showarticletitle{{PyTorch: An Imperative Style, High-Performance
  Deep Learning Library}}.
\newblock \bibinfo{journal}{\emph{arXiv preprint arXiv:1912.01703}}
  (\bibinfo{year}{2019}).
\newblock


\bibitem[\protect\citeauthoryear{Pollock, Roa, Benditt, and Celli}{Pollock
  et~al\mbox{.}}{1993}]%
        {pollock1993estimation}
\bibfield{author}{\bibinfo{person}{Mark Pollock}, \bibinfo{person}{Jairo Roa},
  \bibinfo{person}{Joshua Benditt}, {and} \bibinfo{person}{Bartolome Celli}.}
  \bibinfo{year}{1993}\natexlab{}.
\newblock \showarticletitle{{Estimation of Ventilatory Reserve by Stair
  Climbing: A Study in Patients with Chronic Airflow Obstruction}}.
\newblock \bibinfo{journal}{\emph{Chest}} \bibinfo{volume}{104},
  \bibinfo{number}{5} (\bibinfo{year}{1993}), \bibinfo{pages}{1378--1383}.
\newblock


\bibitem[\protect\citeauthoryear{Potok, Schuman, Young, Patton, Spedalieri,
  Liu, Yao, Rose, and Chakma}{Potok et~al\mbox{.}}{2018}]%
        {potok2018study}
\bibfield{author}{\bibinfo{person}{Thomas~E. Potok}, \bibinfo{person}{Catherine
  Schuman}, \bibinfo{person}{Steven Young}, \bibinfo{person}{Robert Patton},
  \bibinfo{person}{Federico Spedalieri}, \bibinfo{person}{Jeremy Liu},
  \bibinfo{person}{Ke-Thia Yao}, \bibinfo{person}{Garrett Rose}, {and}
  \bibinfo{person}{Gangotree Chakma}.} \bibinfo{year}{2018}\natexlab{}.
\newblock \showarticletitle{{A Study of Complex Deep Learning Networks on
  High-performance, Neuromorphic, and Quantum Computers}}.
\newblock \bibinfo{journal}{\emph{ACM Journal on Emerging Technologies in
  Computing Systems (JETC)}} \bibinfo{volume}{14}, \bibinfo{number}{2}
  (\bibinfo{year}{2018}), \bibinfo{pages}{1--21}.
\newblock


\bibitem[\protect\citeauthoryear{Rasouli, Solnik, Furmanek, Piscitelli, Falaki,
  and Latash}{Rasouli et~al\mbox{.}}{2017}]%
        {rasouli2017unintentional}
\bibfield{author}{\bibinfo{person}{Omid Rasouli}, \bibinfo{person}{Stanis{\l}aw
  Solnik}, \bibinfo{person}{Mariusz~P. Furmanek}, \bibinfo{person}{Daniele
  Piscitelli}, \bibinfo{person}{Ali Falaki}, {and} \bibinfo{person}{Mark~L.
  Latash}.} \bibinfo{year}{2017}\natexlab{}.
\newblock \showarticletitle{{Unintentional Drifts During Quiet Stance and
  Voluntary Body Sway}}.
\newblock \bibinfo{journal}{\emph{Experimental brain research}}
  \bibinfo{volume}{235}, \bibinfo{number}{7} (\bibinfo{year}{2017}),
  \bibinfo{pages}{2301--2316}.
\newblock


\bibitem[\protect\citeauthoryear{{Raspberry Pi Foundation}}{{Raspberry Pi
  Foundation}}{2021}]%
        {rpi}
\bibfield{author}{\bibinfo{person}{{Raspberry Pi Foundation}}.}
  \bibinfo{year}{2021}\natexlab{}.
\newblock \bibinfo{title}{{Teach, Learn and Make with Raspberry Pi - Raspberry
  Pi}}.
\newblock \bibinfo{howpublished}{\url{https://https://www.raspberrypi.org/}}.
\newblock
\newblock
\shownote{Accessed: 2021-04-28.}


\bibitem[\protect\citeauthoryear{Ravichandran, Saba, Chen, Goel, Gupta, and
  Patel}{Ravichandran et~al\mbox{.}}{2015}]%
        {ravichandran2015wibreathe}
\bibfield{author}{\bibinfo{person}{Ruth Ravichandran}, \bibinfo{person}{Elliot
  Saba}, \bibinfo{person}{Ke-Yu Chen}, \bibinfo{person}{Mayank Goel},
  \bibinfo{person}{Sidhant Gupta}, {and} \bibinfo{person}{Shwetak~N. Patel}.}
  \bibinfo{year}{2015}\natexlab{}.
\newblock \showarticletitle{{WiBreathe: Estimating Respiration Rate Using
  Wireless Signals in Natural Settings in the Home}}. In
  \bibinfo{booktitle}{\emph{Proc. of the 13rd IEEE PerCom}}.
  \bibinfo{pages}{131--139}.
\newblock


\bibitem[\protect\citeauthoryear{Rezende, Mohamed, and Wierstra}{Rezende
  et~al\mbox{.}}{2014}]%
        {rezende2014stochastic}
\bibfield{author}{\bibinfo{person}{Danilo~Jimenez Rezende},
  \bibinfo{person}{Shakir Mohamed}, {and} \bibinfo{person}{Daan Wierstra}.}
  \bibinfo{year}{2014}\natexlab{}.
\newblock \showarticletitle{{Stochastic Backpropagation and Approximate
  Inference in Deep Generative Models}}. In \bibinfo{booktitle}{\emph{Proc. of
  ICML}}. \bibinfo{pages}{1278--1286}.
\newblock


\bibitem[\protect\citeauthoryear{Shmelkov, Schmid, and Alahari}{Shmelkov
  et~al\mbox{.}}{2018}]%
        {shmelkov2018good}
\bibfield{author}{\bibinfo{person}{Konstantin Shmelkov},
  \bibinfo{person}{Cordelia Schmid}, {and} \bibinfo{person}{Karteek Alahari}.}
  \bibinfo{year}{2018}\natexlab{}.
\newblock \showarticletitle{{How Good is My GAN?}}. In
  \bibinfo{booktitle}{\emph{Proc. of the 15th ECCV}}.
  \bibinfo{pages}{213--229}.
\newblock


\bibitem[\protect\citeauthoryear{Song, Yang, Yang, Chen, Forno, Chen, and
  Gao}{Song et~al\mbox{.}}{2020}]%
        {song2020spirosonic}
\bibfield{author}{\bibinfo{person}{Xingzhe Song}, \bibinfo{person}{Boyuan
  Yang}, \bibinfo{person}{Ge Yang}, \bibinfo{person}{Ruirong Chen},
  \bibinfo{person}{Erick Forno}, \bibinfo{person}{Wei Chen}, {and}
  \bibinfo{person}{Wei Gao}.} \bibinfo{year}{2020}\natexlab{}.
\newblock \showarticletitle{{SpiroSonic: Monitoring Human Lung Function via
  Acoustic Sensing on Commodity Smartphones}}. In
  \bibinfo{booktitle}{\emph{Proc. of The 26th ACM MobiCom}}.
  \bibinfo{pages}{1--14}.
\newblock


\bibitem[\protect\citeauthoryear{Soriano, Kendrick, Paulson, Gupta, Abrams,
  Adedoyin, Adhikari, Advani, Agrawal, Ahmadian, et~al\mbox{.}}{Soriano
  et~al\mbox{.}}{2020}]%
        {soriano2020prevalence}
\bibfield{author}{\bibinfo{person}{Joan~B. Soriano}, \bibinfo{person}{Parkes~J.
  Kendrick}, \bibinfo{person}{Katherine~R. Paulson}, \bibinfo{person}{Vinay
  Gupta}, \bibinfo{person}{Elissa~M. Abrams}, \bibinfo{person}{Rufus~Adesoji
  Adedoyin}, \bibinfo{person}{Tara~Ballav Adhikari},
  \bibinfo{person}{Shailesh~M. Advani}, \bibinfo{person}{Anurag Agrawal},
  \bibinfo{person}{Elham Ahmadian}, {et~al\mbox{.}}}
  \bibinfo{year}{2020}\natexlab{}.
\newblock \showarticletitle{{Prevalence and Attributable Health Burden of
  Chronic Respiratory Diseases, 1990--2017: A Systematic Analysis for the
  Global Burden of Disease Study 2017}}.
\newblock \bibinfo{journal}{\emph{The Lancet Respiratory Medicine}}
  \bibinfo{volume}{8}, \bibinfo{number}{6} (\bibinfo{year}{2020}),
  \bibinfo{pages}{585--596}.
\newblock


\bibitem[\protect\citeauthoryear{Trabelsi, Bilaniuk, Zhang, Serdyuk,
  Subramanian, Santos, Mehri, Rostamzadeh, Bengio, and Pal}{Trabelsi
  et~al\mbox{.}}{2018}]%
        {deepcomplex}
\bibfield{author}{\bibinfo{person}{Chiheb Trabelsi}, \bibinfo{person}{Olexa
  Bilaniuk}, \bibinfo{person}{Ying Zhang}, \bibinfo{person}{Dmitriy Serdyuk},
  \bibinfo{person}{Sandeep Subramanian}, \bibinfo{person}{Joao~Felipe Santos},
  \bibinfo{person}{Soroush Mehri}, \bibinfo{person}{Negar Rostamzadeh},
  \bibinfo{person}{Yoshua Bengio}, {and} \bibinfo{person}{Christopher~J. Pal}.}
  \bibinfo{year}{2018}\natexlab{}.
\newblock \showarticletitle{{Deep Complex Networks}}. In
  \bibinfo{booktitle}{\emph{Proc. of ICLR}}. \bibinfo{pages}{1--19}.
\newblock


\bibitem[\protect\citeauthoryear{Trinh, Dai, Luong, and Le}{Trinh
  et~al\mbox{.}}{2018}]%
        {trinh2018learning}
\bibfield{author}{\bibinfo{person}{Trieu Trinh}, \bibinfo{person}{Andrew Dai},
  \bibinfo{person}{Thang Luong}, {and} \bibinfo{person}{Quoc Le}.}
  \bibinfo{year}{2018}\natexlab{}.
\newblock \showarticletitle{{Learning Longer-Term Dependencies in RNNs with
  Auxiliary Losses}}. In \bibinfo{booktitle}{\emph{Proc. of ICML}}. PMLR,
  \bibinfo{pages}{4965--4974}.
\newblock


\bibitem[\protect\citeauthoryear{Tsuyuki, Midodzi, Villa-Roel, Marciniuk,
  Mayers, Vethanayagam, Chan, and Rowe}{Tsuyuki et~al\mbox{.}}{2020}]%
        {tsuyuki2020diagnostic}
\bibfield{author}{\bibinfo{person}{Ross~T. Tsuyuki}, \bibinfo{person}{William
  Midodzi}, \bibinfo{person}{Cristina Villa-Roel}, \bibinfo{person}{Darcy
  Marciniuk}, \bibinfo{person}{Irvin Mayers}, \bibinfo{person}{Dilini
  Vethanayagam}, \bibinfo{person}{Michael Chan}, {and}
  \bibinfo{person}{Brian~H. Rowe}.} \bibinfo{year}{2020}\natexlab{}.
\newblock \showarticletitle{{Diagnostic Practices for Patients with Shortness
  of Breath and Presumed Obstructive Airway Disorders: A Cross-Sectional
  Analysis}}.
\newblock \bibinfo{journal}{\emph{CMAJ open}} \bibinfo{volume}{8},
  \bibinfo{number}{3} (\bibinfo{year}{2020}), \bibinfo{pages}{E605}.
\newblock


\bibitem[\protect\citeauthoryear{Tu, Hwang, and Lin}{Tu et~al\mbox{.}}{2016}]%
        {Tu20161D}
\bibfield{author}{\bibinfo{person}{Jianxuan Tu}, \bibinfo{person}{Taesong
  Hwang}, {and} \bibinfo{person}{Jenshan Lin}.}
  \bibinfo{year}{2016}\natexlab{}.
\newblock \showarticletitle{{Respiration Rate Measurement Under 1-D Body Motion
  Using Single Continuous-Wave Doppler Radar Vital Sign Detection System}}.
\newblock \bibinfo{journal}{\emph{IEEE Transactions on Microwave Theory and
  Techniques}} \bibinfo{volume}{64}, \bibinfo{number}{6}
  (\bibinfo{year}{2016}), \bibinfo{pages}{1937--1946}.
\newblock


\bibitem[\protect\citeauthoryear{van Gastel, Stuijk, and de~Haan}{van Gastel
  et~al\mbox{.}}{2016}]%
        {van2016robust}
\bibfield{author}{\bibinfo{person}{Mark van Gastel}, \bibinfo{person}{Sander
  Stuijk}, {and} \bibinfo{person}{Gerard de Haan}.}
  \bibinfo{year}{2016}\natexlab{}.
\newblock \showarticletitle{{Robust Respiration Detection from Remote
  Photoplethysmography}}.
\newblock \bibinfo{journal}{\emph{Biomedical Optics Express}}
  \bibinfo{volume}{7}, \bibinfo{number}{12} (\bibinfo{year}{2016}),
  \bibinfo{pages}{4941--4957}.
\newblock


\bibitem[\protect\citeauthoryear{Vos, Allen, Arora, Barber, Bhutta, Brown,
  Carter, Casey, Charlson, Chen, et~al\mbox{.}}{Vos et~al\mbox{.}}{2016}]%
        {vos2016global}
\bibfield{author}{\bibinfo{person}{Theo Vos}, \bibinfo{person}{Christine
  Allen}, \bibinfo{person}{Megha Arora}, \bibinfo{person}{Ryan~M. Barber},
  \bibinfo{person}{Zulfiqar~A. Bhutta}, \bibinfo{person}{Alexandria Brown},
  \bibinfo{person}{Austin Carter}, \bibinfo{person}{Daniel~C. Casey},
  \bibinfo{person}{Fiona~J. Charlson}, \bibinfo{person}{Alan~Z. Chen},
  {et~al\mbox{.}}} \bibinfo{year}{2016}\natexlab{}.
\newblock \showarticletitle{{Global, Regional, and National Incidence,
  Prevalence, and Years Lived with Disability for 310 Diseases and Injuries,
  1990--2015: A Systematic Analysis for the Global Burden of Disease Study
  2015}}.
\newblock \bibinfo{journal}{\emph{The Lancet}} \bibinfo{volume}{388},
  \bibinfo{number}{10053} (\bibinfo{year}{2016}), \bibinfo{pages}{1545--1602}.
\newblock


\bibitem[\protect\citeauthoryear{Wang, Sunshine, and Gollakota}{Wang
  et~al\mbox{.}}{2019}]%
        {wang2019contactless}
\bibfield{author}{\bibinfo{person}{Anran Wang}, \bibinfo{person}{Jacob~E.
  Sunshine}, {and} \bibinfo{person}{Shyamnath Gollakota}.}
  \bibinfo{year}{2019}\natexlab{}.
\newblock \showarticletitle{Contactless Infant Monitoring Using White Noise}.
  In \bibinfo{booktitle}{\emph{Proc. of The 25th ACM MobiCom}}.
  \bibinfo{pages}{52:1--16}.
\newblock


\bibitem[\protect\citeauthoryear{Wang, Zhang, Zheng, Gu, Zhou, and
  Dorizzi}{Wang et~al\mbox{.}}{2018}]%
        {wang2018c}
\bibfield{author}{\bibinfo{person}{Tianben Wang}, \bibinfo{person}{Daqing
  Zhang}, \bibinfo{person}{Yuanqing Zheng}, \bibinfo{person}{Tao Gu},
  \bibinfo{person}{Xingshe Zhou}, {and} \bibinfo{person}{Bernadette Dorizzi}.}
  \bibinfo{year}{2018}\natexlab{}.
\newblock \showarticletitle{{C-FMCW based Contactless Respiration Detection
  using Acoustic Signal}}. In \bibinfo{booktitle}{\emph{Proc. of the 20th ACM
  UbiComp}}. \bibinfo{pages}{170:1--20}.
\newblock


\bibitem[\protect\citeauthoryear{White}{White}{2006}]%
        {white2006sleep}
\bibfield{author}{\bibinfo{person}{David~P. White}.}
  \bibinfo{year}{2006}\natexlab{}.
\newblock \showarticletitle{{Sleep Apnea}}.
\newblock \bibinfo{journal}{\emph{Proceedings of the American Thoracic
  Society}} \bibinfo{volume}{3}, \bibinfo{number}{1} (\bibinfo{year}{2006}),
  \bibinfo{pages}{124--128}.
\newblock


\bibitem[\protect\citeauthoryear{Williams}{Williams}{1993}]%
        {williams1993chronic}
\bibfield{author}{\bibinfo{person}{Simon~Johnson Williams}.}
  \bibinfo{year}{1993}\natexlab{}.
\newblock \bibinfo{booktitle}{\emph{{Chronic Respiratory Illness}}}.
\newblock \bibinfo{publisher}{Psychology Press}.
\newblock


\bibitem[\protect\citeauthoryear{WiRUSH/AIWiSe}{WiRUSH/AIWiSe}{2019}]%
        {wirush}
\bibfield{author}{\bibinfo{person}{WiRUSH/AIWiSe}.}
  \bibinfo{year}{2019}\natexlab{}.
\newblock \bibinfo{title}{{Guangxi Wanyun and Guangzhou AIWiSe Technology Co.,
  Ltd}}.
\newblock
  \bibinfo{howpublished}{\url{https://www.wirush.ai}~~~and~~~\url{https://aiwise.wirush.ai}}.
\newblock


\bibitem[\protect\citeauthoryear{Wu and Huang}{Wu and Huang}{2009}]%
        {wu2009ensemble}
\bibfield{author}{\bibinfo{person}{Zhaohua Wu} {and} \bibinfo{person}{Norden~E.
  Huang}.} \bibinfo{year}{2009}\natexlab{}.
\newblock \showarticletitle{{Ensemble Empirical Mode Decomposition: A
  Noise-Assisted Data Analysis Method}}.
\newblock \bibinfo{journal}{\emph{Advances in Adaptive Data Analysis}}
  \bibinfo{volume}{1}, \bibinfo{number}{01} (\bibinfo{year}{2009}),
  \bibinfo{pages}{1--41}.
\newblock


\bibitem[\protect\citeauthoryear{Xu, Yu, Chen, Zhu, Kong, and Li}{Xu
  et~al\mbox{.}}{2019}]%
        {xu2019breathlisener}
\bibfield{author}{\bibinfo{person}{Xiangyu Xu}, \bibinfo{person}{Jiadi Yu},
  \bibinfo{person}{Yingying Chen}, \bibinfo{person}{Yanmin Zhu},
  \bibinfo{person}{Linghe Kong}, {and} \bibinfo{person}{Minglu Li}.}
  \bibinfo{year}{2019}\natexlab{}.
\newblock \showarticletitle{BreathListener: Fine-Grained Breathing Monitoring
  in Driving Environments Utilizing Acoustic Signals}. In
  \bibinfo{booktitle}{\emph{Proc. of the 17th ACM MobiSys}}.
  \bibinfo{pages}{54–66}.
\newblock


\bibitem[\protect\citeauthoryear{Yang, Pathak, Zeng, Liran, and Mohapatra}{Yang
  et~al\mbox{.}}{2017}]%
        {yang2017vital}
\bibfield{author}{\bibinfo{person}{Zhicheng Yang}, \bibinfo{person}{Parth~H.
  Pathak}, \bibinfo{person}{Yunze Zeng}, \bibinfo{person}{Xixi Liran}, {and}
  \bibinfo{person}{Prasant Mohapatra}.} \bibinfo{year}{2017}\natexlab{}.
\newblock \showarticletitle{Vital Sign and Sleep Monitoring using Millimeter
  Wave}.
\newblock \bibinfo{journal}{\emph{ACM Transactions on Sensor Networks}}
  \bibinfo{volume}{13}, \bibinfo{number}{2} (\bibinfo{year}{2017}),
  \bibinfo{pages}{1--32}.
\newblock


\bibitem[\protect\citeauthoryear{Yue, He, Wang, Rahul, and Katabi}{Yue
  et~al\mbox{.}}{2018}]%
        {yue2018extract}
\bibfield{author}{\bibinfo{person}{Shichao Yue}, \bibinfo{person}{Hao He},
  \bibinfo{person}{Hao Wang}, \bibinfo{person}{Hariharan Rahul}, {and}
  \bibinfo{person}{Dina Katabi}.} \bibinfo{year}{2018}\natexlab{}.
\newblock \showarticletitle{{Extracting Multi-Person Respiration from Entangled
  RF Signals}}. In \bibinfo{booktitle}{\emph{Proc. of the 20th ACM UbiComp}}.
  \bibinfo{pages}{86:1--22}.
\newblock


\bibitem[\protect\citeauthoryear{Zeiler, Krishnan, Taylor, and Fergus}{Zeiler
  et~al\mbox{.}}{2010}]%
        {zeiler2010deconvolutional}
\bibfield{author}{\bibinfo{person}{Matthew~D. Zeiler}, \bibinfo{person}{Dilip
  Krishnan}, \bibinfo{person}{Graham~W. Taylor}, {and} \bibinfo{person}{Rob
  Fergus}.} \bibinfo{year}{2010}\natexlab{}.
\newblock \showarticletitle{{Deconvolutional Networks}}. In
  \bibinfo{booktitle}{\emph{Proc. of the 23rd IEEE CVPR}}. IEEE,
  \bibinfo{pages}{2528--2535}.
\newblock


\bibitem[\protect\citeauthoryear{Zeng, Wu, Xiong, Yi, Gao, and Zhang}{Zeng
  et~al\mbox{.}}{2019}]%
        {zeng2019farsense}
\bibfield{author}{\bibinfo{person}{Youwei Zeng}, \bibinfo{person}{Dan Wu},
  \bibinfo{person}{Jie Xiong}, \bibinfo{person}{Enze Yi},
  \bibinfo{person}{Ruiyang Gao}, {and} \bibinfo{person}{Daqing Zhang}.}
  \bibinfo{year}{2019}\natexlab{}.
\newblock \showarticletitle{{FarSense: Pushing the Range Limit of WiFi-based
  Respiration Sensing with CSI Ratio of Two Antennas}}. In
  \bibinfo{booktitle}{\emph{Proc. of the 21st ACM UbiComp}}.
  \bibinfo{pages}{1--26}.
\newblock


\bibitem[\protect\citeauthoryear{Zhang, Li, Luo, and He}{Zhang
  et~al\mbox{.}}{2014}]%
        {iLocScan}
\bibfield{author}{\bibinfo{person}{Chi Zhang}, \bibinfo{person}{Feng Li},
  \bibinfo{person}{Jun Luo}, {and} \bibinfo{person}{Ying He}.}
  \bibinfo{year}{2014}\natexlab{}.
\newblock \showarticletitle{{iLocScan: Harnessing Multipath for Simultaneous
  Indoor Source Localization and Space Scanning}}. In
  \bibinfo{booktitle}{\emph{Proc. of the 12th ACM SenSys}}.
  \bibinfo{pages}{91—104}.
\newblock


\bibitem[\protect\citeauthoryear{Zhang, Xu, Hu, and Kanhere}{Zhang
  et~al\mbox{.}}{2017}]%
        {zhang2017wicare}
\bibfield{author}{\bibinfo{person}{Jin Zhang}, \bibinfo{person}{Weitao Xu},
  \bibinfo{person}{Wen Hu}, {and} \bibinfo{person}{Salil~S. Kanhere}.}
  \bibinfo{year}{2017}\natexlab{}.
\newblock \showarticletitle{{WiCare: Towards in-Situ Breath Monitoring}}. In
  \bibinfo{booktitle}{\emph{Proc. of the 14th EAI MobiQuitous}}.
  \bibinfo{pages}{126--135}.
\newblock


\bibitem[\protect\citeauthoryear{Zheng, Chen, Cai, Luo, and Zhang}{Zheng
  et~al\mbox{.}}{2020}]%
        {zheng2020v2ifi}
\bibfield{author}{\bibinfo{person}{Tianyue Zheng}, \bibinfo{person}{Zhe Chen},
  \bibinfo{person}{Chao Cai}, \bibinfo{person}{Jun Luo}, {and}
  \bibinfo{person}{Xu Zhang}.} \bibinfo{year}{2020}\natexlab{}.
\newblock \showarticletitle{V\textsuperscript{2}iFi: in-Vehicle Vital Sign
  Monitoring via Compact RF Sensing}. In \bibinfo{booktitle}{\emph{Proc. of the
  22th ACM UbiComp}}. \bibinfo{pages}{70:1--27}.
\newblock


\bibitem[\protect\citeauthoryear{Zheng, Chen, Ding, and Luo}{Zheng
  et~al\mbox{.}}{2021a}]%
        {RFMag}
\bibfield{author}{\bibinfo{person}{Tianyue Zheng}, \bibinfo{person}{Zhe Chen},
  \bibinfo{person}{Shuya Ding}, {and} \bibinfo{person}{Jun Luo}.}
  \bibinfo{year}{2021}\natexlab{a}.
\newblock \showarticletitle{{Enhancing RF Sensing with Deep Learning: A Layered
  Approach}}.
\newblock \bibinfo{journal}{\emph{IEEE Communications Magazine}}
  \bibinfo{volume}{59}, \bibinfo{number}{2} (\bibinfo{year}{2021}),
  \bibinfo{pages}{70--76}.
\newblock


\bibitem[\protect\citeauthoryear{Zheng, Chen, Luo, Ke, Zhao, and Yang}{Zheng
  et~al\mbox{.}}{2021b}]%
        {siwa}
\bibfield{author}{\bibinfo{person}{Tianyue Zheng}, \bibinfo{person}{Zhe Chen},
  \bibinfo{person}{Jun Luo}, \bibinfo{person}{Lin Ke},
  \bibinfo{person}{Chaoyang Zhao}, {and} \bibinfo{person}{Yaowen Yang}.}
  \bibinfo{year}{2021}\natexlab{b}.
\newblock \showarticletitle{{SiWa: See into Walls via Deep UWB Radar}}. In
  \bibinfo{booktitle}{\emph{Proc. of the 27th ACM MobiCom}}.
  \bibinfo{pages}{1--14}.
\newblock


\end{thebibliography}

\end{document}